\documentclass[preprint,12pt]{elsarticle}

\usepackage[T1]{fontenc}
\usepackage[english]{babel}
\usepackage{indentfirst}
\usepackage{caption}
\usepackage{subcaption,array}
\usepackage{amsmath}
\usepackage{siunitx}
\usepackage{amsfonts}
\usepackage{amssymb}
\usepackage{amsthm}
\usepackage{mathtools}
\usepackage{tabularx} 
\newcolumntype{Y}{>{\centering\arraybackslash}X}
\usepackage{enumitem}
\usepackage{algorithm}
\usepackage{algorithmic}
\usepackage[title]{appendix}
\usepackage{hyperref}
\usepackage{graphicx}
\usepackage{stmaryrd}
\usepackage{xspace}
\usepackage{listings}
\usepackage{siunitx}
\usepackage{datetime2}

\lstnewenvironment{rc}[1][]{\lstset{language=R}}{}
\newtheorem{theorem}{Theorem}[section] 
\usepackage{xpatch}
\makeatletter
\AtBeginDocument{\xpatchcmd{\@thm}{\thm@headpunct{.}}{\thm@headpunct{}}{}{}}
\makeatother
\newtheorem{definition}[theorem]{Definition}

\makeatletter
\newcommand\footnoteref[1]{\protected@xdef\@thefnmark{\ref{#1}}\@footnotemark}
\makeatother

\newcommand{\lz}{\ensuremath{\ell_0}\xspace}
\newcommand{\lzc}{\ensuremath{\ell_0^c}\xspace}
\newcommand{\lzw}{\ensuremath{\ell_0(\mathbf{w})}\xspace}
\newcommand{\lzcw}{\ensuremath{\ell_0^c(\mathbf{w})}\xspace}
\newcommand{\dspls}[1]{Dual-sPLS\textsubscript{#1}\xspace }
\newcommand{\Dspls}[1]{Dual-sPLS\textsubscript{#1}\xspace }
\newcommand{\spls}[1]{sPLS\textsubscript{#1}\xspace}
\newcommand{\Ds}{\ensuremath{D_\mathrm{SIM}}\xspace}
\newcommand{\Dsb}{\ensuremath{\overline{D}_\mathrm{SIM}}\xspace}
\newcommand{\DNIR}{\ensuremath{D_\mathrm{NIR}}\xspace}

\DeclareMathOperator*{\argmin}{\arg\min}

\def\R{\mathbb{R}\xspace}
\def\vecz{\textbf{z}\xspace}
\def\vecw{\textbf{w}\xspace}
\def\vecy{\textbf{y}\xspace}
\def\matX{\textbf{X}\xspace}
\def\maty{\textbf{Y}\xspace}
\def\R{\mathbb{R}\xspace}
\def\vecb{\boldsymbol \beta}

\def\vecep{\boldsymbol \epsilon}

\begin{document}

\begin{frontmatter}

\title{Dual-sPLS: a family of Dual Sparse Partial Least Squares regressions for feature selection and prediction with tunable sparsity; evaluation on simulated and near-infrared (NIR) data}
\author[usj,ucbl]{Louna Alsouki}
\author[ifpen]{Laurent Duval}
\author[ucbl]{Cl{\'{e}}ment Marteau}
\author[usj]{Rami El Haddad}
\author[ifpen,ucbl]{Fran{\c{c}}ois Wahl}
%
\affiliation[usj]{organization={Laboratoire de Math\'ematiques et Applications, U. R. Math\'ematiques et mod\'elisation, Facult\'e des sciences, 
Universit\'e Saint-Joseph},
            addressline={B.P. 7-5208}, 
            city={Mar Mikha{\"e}l Beyrouth},
            postcode={1104 2020}, 
            country={Liban}}
\affiliation[ifpen]{organization={IFP Energies nouvelles},
            addressline={1 et 4 avenue de Bois-Préau}, 
           city={Rueil-Malmaison},
            postcode={92852}, 
           country={France}}
\affiliation[ucbl]{organization={Institut Camille Jordan, Universit\'e Claude Bernard Lyon 1},
            addressline={43 boulevard du 11 novembre 1918}, 
            city={Villeurbanne},
            postcode={69100},
            country={France}}

\begin{abstract}
Relating a set of variables \matX to a response \vecy is crucial in chemometrics. A quantitative prediction objective can be enriched by qualitative data interpretation, for instance by locating the most influential features. When high-dimensional problems arise, dimension reduction techniques can be used. Most notable are projections (e.g. Partial Least Squares or PLS ) or variable selections (e.g. lasso). Sparse partial least squares combine both strategies, by blending variable selection into PLS. The variant presented in this paper, Dual-sPLS, generalizes the classical PLS1 algorithm. It provides balance between accurate prediction and efficient interpretation. It is based on penalizations inspired by classical regression methods (lasso, group lasso, least squares, ridge) and uses the dual norm notion. The resulting sparsity is enforced by an intuitive shrinking ratio parameter. Dual-sPLS favorably compares to similar regression methods, on simulated and real chemical data. Code is provided as an open-source package in R: \url{https://CRAN.R-project.org/package=dual.spls}.
\end{abstract}
\pagebreak


\end{frontmatter}


\section{Introduction \label{sec_dspls_introduction}}
Two main feats of chemometrics reside in first, providing reliable inference and second, offering interpretability of chemical data sources. On the one hand, one may expect to estimate, within a given precision, responses \textbf{Y}$\in \R^{N \times Q}$ (e.g. hydrocarbon properties: viscosity, density, cetane number \cite{RamirezVerduzco_L_2012_j-fuel_predicting_cnkvdhhvbfamec}) from spectra or variables represented by quantities \textbf{X}$\in \R^{N \times P}$ (nuclear magnitude resonance or NMR, chromatography, infrared spectroscopy,  etc. \cite{Willard_H_1965_book_methodes_pac}). It aims at relating a target $\textbf{Y}$ to \textbf{X} through a predictive model: for instance, NMR spectra can be linked to viscosity with predictive purposes. On the other hand, one also wishes to interpret how variables in \textbf{X} influence quantities \textbf{Y}: for instance, which spectral bands in NMR affect most viscosity index (see e.g. \cite{Verdier_S_2009_j-fuel_critical_avi})? This can be transcribed by a regression model, often considered linear: 
\begin{equation}\label{eq2}
\vecy=\matX \vecb + \vecep\,,
\end{equation}
where $\vecep$ is expected to be independent of \matX, with zero mean. 
\noindent With the growing size of consolidated analytical chemistry databases, chemometrics still require methodologies to 1) provide accurate predictions 2) extract pertinent knowledge or offer useful insights on measurements 3) combine heterogeneous or high-dimensional data sources.
When the number $P$ of variables (samples) is far  greater than the number of observations (signals) $N$ ($P \gg N$), naive statistical models risk overfitting. This notably happens in standard least squares optimizations. Dimension reduction techniques are generic approaches to deal with high dimensionality. They include  projection methods or variable selection algorithms. Commonly used strategies start with PCA/PCR (principal component analysis/regression), performed only on explanatory variables in \textbf{X}. They however do not incorporate information held by the response \textbf{Y}. Partial least squares (PLS) \cite{Wold_S_1995_j-chem-int-lab-syst_chemometrics_wdwmwiwdwwfi,Wold_S_2001_j-chem-int-lab-syst_pls-regression_btc}, also called projection onto latent structures, is therefore common in chemometrics, with better prediction-prone latent components. However, PLS sometimes lacks appropriate interpretability. \\
As for variable selection, one often resorts to the lasso algorithm (least absolute shrinkage and selection operator \cite{Tibshirani_R_1996_j-r-stat-soc-b-stat-methodol_regression_sslasso}). Shrinkage induces a form of sparsity, which amounts to selecting important variables. It is however known to be sensitive to data types. It does not always yield interpretable coefficients. Blends of the two above -- dimension reduction and variable selection -- have recent avatars called sparse PLS (sPLS). While they enforce lower dimensional decompositions, they do not always provide chemically pertinent feature localization for physico-analytical measurements. Thereby, we propose a dual sparse PLS family dedicated to one dimensional or univariate  responses: $\vecy=\maty$, with $\vecy \in \R^N$. It generalizes the standard PLS1 algorithm by supplementing it with adequate  penalties. This formally provides a unified formulation for regression methods in the spirit of the lasso mentioned above, and also least-squares or ridge, all blended in a PLS formalism. It also allows \emph{variable grouping}: the possibility to gather explanatory variables into more meaningful subsets (contiguous samples around a peak, disjoint spectral bands associated to a compound). This can be used to combine different physico-chemical modalities.
Resolution resorts to the dual norm of the chosen Dual-sPLS penalty. This new method has many advantages:
\begin{enumerate}
\item predictions match or outperform state-of-the-art or comparable methods, 
\item in the different norm options we considered, they additionally yield sparse representations of both simulated and real chemical near infrared data, even singular, a frequent ill-conditioning issue in high dimension,
\item they finally offer a interpretable localization of features.
\end{enumerate}
Those three properties combined offer alternative surrogates to classical approaches (PLS, lasso, least squares, ridge). It permits both accurate inference and pertinent domain-related interpretation. \\
The paper is structured as follows: setting notations, we briefly revise in Section \ref{sec_dspls_background} the background of the PLS, recall classical variable selection methods and evoke their blending in sparse PLS schemes, previously proposed. Then, in Section \ref{sec_dspls_dspls}, we explain principles behind the Dual-sPLS family and detail the list of norm penalties and their algorithms in three main instances: the (group) lasso form ---being the most important--- and least squares and ridge forms. Thereafter Section \ref{sec_dspls_data_model_evaluation} explicits tested data (simulated and real) and the choices of model settings, calibration and validation. Each of the three penalties types are extensively benchmarked in Section \ref{sec_dspls_results}. We finally draw concluding remarks with perspectives in Section \ref{sec_dspls_conclusion} and supplementary material in the appendix.
\subsection*{Notation and definitions}
Matrices, vectors and scalars are denoted by boldface uppercase letters, boldface lowercase and light lowercase letters respectively, e.g. $\textbf{X}$, $\textbf{y}$ and $\lambda$. The transpose of matrix $\textbf{X}$ is $\textbf{X}^T$.  The identity matrix of size $P$ is represented by $I_P$. The $\ell_1$-norm and the $\ell_2$-norm of vector a $\vecw$ of length $P$ are 
\begin{equation}\label{eq_dspls_norm1_norm2}
\| \vecw \|_1=\sum_{p=1}^P |w_p| \qquad \text{and} \qquad \| \vecw \|_2=\sqrt{  \sum_{p=1}^P |w_p|^{ 2} }\,.
\end{equation}
We denote by \lzw the sparsity index or count measure \cite{Cherni_A_2020_j-ieee-tsp_spoq_lpolqrssrams} of the non-zero coordinates of $\vecw$ and \lzcw its complement i.e. \lzcw$=P-$\lzw. To choose the number of latent variables we rely on the mean squared error (MSE) expressed as 
\begin{equation}\label{eq_dspls_MSE}
\text{MSE}=\dfrac{1}{N}\sum_{n=1}^{N}{(y_n -\hat{y}_n)^2}\,,
\end{equation}
for a response vector $\vecy$ of $N$ observations and a given estimate $\hat{\vecy}$.
For performance evaluation, we choose the root mean squares error (RMSE), the mean absolute error (MAE) and the determination coefficient (R$^2$): 
\begin{equation}
\text{RMSE}=\sqrt{\dfrac{1}{N}\sum_{n=1}^{N}{(y_n -\hat{y}_n)^2}}=\dfrac{
1}{\sqrt{N}} \| \vecy - \hat{\vecy}\|_2\,,
\end{equation}
\begin{equation}\label{eq_dspls_errors}
\text{MAE}=\dfrac{1}{N} \sum_{n=1}^{N} | y_n -\hat{y}_n|=\dfrac{
1}{N} \| \vecy - \hat{\vecy}\|_1\,,
\end{equation}
\begin{equation}
\text{R}^2=\dfrac{\sum_{n=1}^{N}(y_n -\bar{y})^2}{\sum_{n=1}^{N}(y_n -\hat{y}_n)^2} \qquad \text{where} \qquad \bar{y}=\dfrac{\sum_{n=1}^{N} y_n}{N} \,.
\end{equation} 
The vector of signs of $\vecw$ is noted sign$(\vecw)$, and $(\vecw)_+$ is the vector composed of $\vecw_p$ if $\vecw_p \geq 0$ and $0$ if $\vecw_p < 0$\footnote{It corresponds to the Rectified Linear Unit (ReLU), a popular activation function for neural networks.}.\\
\noindent In the following, we assume that the matrix $\textbf{X}$ $\in \R^{N \times P}$ of independent variables and the response vector $\vecy$ $\in \R^{N}$ are mean-centered. We use the convention where columns denote variables and rows observations. 


\section{Background \label{sec_dspls_background}}
\subsection{Partial Least Squares (PLS) \label{sec_dspls_pls}} 
PLS originated from econometrics \cite{MateosAparicioMorales_G_2011_j-commun-stat-theory-methods}. It was progressively and succesfully  applied to other fields \cite{Mehmood_T_2016_j-chemometrics_diversity_aplso}: social and behavioral sciences, biosciences from bioinformatics \cite{Boulesteix_A_2007_j-brief-bioinform_partial_lsvtahdgd} to neuroimaging \cite{Krishnan_A_2011_j-neuroimage_partial_lsplsmntr}, and  chemometrics \cite{Wold_S_1995_j-chem-int-lab-syst_chemometrics_wdwmwiwdwwfi,Wold_S_2001_j-chem-int-lab-syst_pls-regression_btc}.
It denotes a class of methods aimed at explaining the relationship between explanatory data and responses with the help of latent variables. They boast the management of both formative and reflective measurements, require low sample sizes and mild distributional assumptions.\\
PLS avatars root on projecting data onto a lower dimensional space using new orthogonal directions constructed as linear combinations of original variables. Its principle consists in compressing the predictor $\textbf{X}$ into a smaller score matrix of $M<P$ variables. When Principal Component Analysis (PCA) \cite{Wright_J_2022_book_high-dimensional_daldmpca} ought to best summarize \matX by taking into account only the correlation between the variables in \textbf{X}, the PLS steps up and also consider the covariance between \textbf{X} and \vecy. The latent space is spanned by $M$ components written as $\textbf{t}_m=\textbf{X}\textbf{w}_m$,  $m \in \{ 1, \dots, M\}$. Weights $\textbf{w}_m$ are constructed in order to obtain an orthogonal basis. Several algorithms have been proposed. NIPALS (nonlinear iterative partial least squares) \cite{Wold_H_1975_incoll_path_mlvnipalsa} and SIMPLS \cite{DeJong_S_1993_j-chemometr-intell-lab-syst_simpls_aaplsr} are most popular. When applied to a one-dimensional reponse, as in our case, both are shown to be equivalent. They solve the following optimization problem for the first component:  
\begin{equation}\label{eq_dspls_PLS_opt}
\max_\textbf{w}(\textbf{y}^T\textbf{X}\vecw) \quad \textrm{s.t.} \quad \|\vecw\|_{2}=1\,.
\end{equation}
\noindent The convex Problem \eqref{eq_dspls_PLS_opt} can be solved with Lagrange multipliers. For $\mu > 0$, it rewrites:
\begin{equation}\label{eq_dspls_Lagrange}
\min_\vecw L(\vecw) \quad \textrm{where} \; L(\vecw)=-\vecz^T\vecw+\mu (\|\vecw\|_{2}-1) \; \textrm{ and } \;  \vecz=\textbf{X}^T\vecy\,.
\end{equation}
\noindent Solving \eqref{eq_dspls_Lagrange} leads to
\begin{equation}
\vecw=\matX^T\vecy\,.
\end{equation}
The PLS algorithm uses the weight vector $\vecw$ to project regressor \matX into score vector $\textbf{t}=\matX\vecw$.
\noindent NIPALS iteratively computes weight vectors by deflation while SIMPLS is more direct. Let $\mathcal{P}_{\textbf{t}_{m-1}} $ denotes the orthogonal projection onto the space spanned by components  $\textbf{t}_1, \dots, \textbf{t}_{m-1}$. The algorithm considers the part of $\textbf{X}$ that is orthogonal to $\textbf{t}_k, k<m$.  For the $m^\textrm{th}$ component, \textbf{X} is replaced by $\textbf{X}_m$ such that: 
\begin{equation} \label{eq_dspls_deflation}
\textbf{X}_m=\textbf{X}-\mathcal{P}_{\textbf{t}_1, \dots, \textbf{t}_{m-1}} \textbf{X}=\textbf{X}_{m-1}-\mathcal{P}_{\textbf{t}_{m-1}} \textbf{X}_{m-1}\,.
\end{equation} 
\noindent The PLS1 algorithm is  described in Algorithm \ref{algo_dspls_PLS1}.
\begin{algorithm} [H]
\caption{ PLS1}
\begin{algorithmic}  \label{algo_dspls_PLS1}
\STATE Input: $\textbf{X},\textbf{y},M$
\STATE $\textbf{X}_1=\textbf{X}$
\FOR{ $ m= 1, \dots, M$ }
\STATE $\textbf{w}_m=\textbf{X}_m^T\textbf{y}$ (weight vector computation)
\STATE $\textbf{t}_m=\textbf{X}_m\textbf{w}_m$ (component construction)
\STATE $\textbf{X}_{m+1}=\textbf{X}_m-\mathcal{P}_{\textbf{t}_m}\textbf{X}_m$ (deflation)
\ENDFOR
\end{algorithmic}
\end{algorithm}
\noindent PLS thus projects \matX onto the space of lower dimension spanned by the loadings $\textbf{w}_1, \dots, \textbf{w}_{m}$, in order to replace \matX by \textbf{T}=\textbf{XW}, where \textbf{T}$ \in \R^{N\times M}$. The PLS regression fitted values for $M$ components is given by:
\begin{equation}
\hat{\vecy}=\textbf{T}\hat{\boldsymbol \beta}=\textbf{T}(\textbf{T}^T\textbf{T})^{-1}\textbf{T}\vecy=\textbf{X}\textbf{W}(\textbf{T}^T\textbf{T})^{-1}\textbf{T}\vecy\,.
\end{equation}
Based on the above, PLS regression coefficients are computed as: 
\begin{equation}
\hat{\boldsymbol \beta}^{PLS}=\textbf{W}(\textbf{T}^T\textbf{T})^{-1}\textbf{T}\vecy\,.
\end{equation}
\subsection{Least absolute shrinkage and selection operator \label{sec_dspls_lasso}} 
By selecting the most important features, variable selection produces a less complicated model. It has the potential advantage of being easier to handle than the complete full set of variables. The optimization problem in standard linear regression is stated as:
\begin{equation}\label{eq_dspls_LS_opt}
\arg\min_{\boldsymbol \beta} \|  \vecy - \textbf{X} \boldsymbol \beta \|^2_2\,.
\end{equation}
\noindent Provided \matX has full column rank, the ordinary least squares (LS) estimation is $\hat{\vecy}_{LS}=\mathcal{P}_{[\matX]}\vecy$, where $[\matX]$ is the space spanned by the columns of $\matX$. In other terms, $\hat{\boldsymbol \beta}^{\text{LS}}= (\textbf{X}^T\textbf{X})^{-1}\textbf{X}^T \vecy$ A popular sparsity-based approach is the lasso developed by Tibshirani in 1996 \cite{Tibshirani_R_1996_j-r-stat-soc-b-stat-methodol_regression_sslasso}. It is reknown for its $\ell_1$ penalty scheme that shrinks less relevant variables to zero. It is obtained by solving:
\begin{equation}\label{eq_dspls_lasso_opt}
\argmin_{\boldsymbol \beta} \|  \vecy - \textbf{X} \boldsymbol \beta \|^2_2 \quad \text{ subject to } \|\boldsymbol \beta \|_1\leq\lambda\,.
\end{equation}
\noindent Threshold parameter $\lambda > 0$ controls the extent of shrinkage applied to the estimate; that is, the number \lzc of coefficients set to zero. An appropriate $\lambda$ is important to get interpretable results. If $\hat{\boldsymbol \beta}^{\text{LS}}$ exists, as mentioned in \cite{Tibshirani_R_1996_j-r-stat-soc-b-stat-methodol_regression_sslasso}, then for a $\lambda \geq \|\hat{\boldsymbol \beta}^{ \text{LS} }\|_1$, the lasso estimate $\hat{\boldsymbol \beta}^{ \text{l} }$ is equal to the ordinary least square solution. And for $\lambda= \dfrac{\|\hat{\boldsymbol \beta}^{ \text{LS} }\|_1}{2}$, it selects on average half of the variables. We can reformulate \eqref{eq_dspls_lasso_opt} as
\begin{equation}\label{eq_dspls_lasso_opt_bis}
\argmin_{\boldsymbol \beta} \dfrac{1}{2}\|  \vecy - \textbf{X} \boldsymbol \beta\|_2^2 + t \|\boldsymbol \beta \|_1 \,.
\end{equation}
Note that there is a (non-explicit) correspondence between parameters $\lambda$ and $t$. In the orthonormal design case, i.e. $\textbf{X}^T\textbf{X}=I_P$, there exists $\hat{ \beta}^{\text{l}}$ closed form solution called \emph{soft thresholding} verifying
\begin{equation}\label{eq_dspls_soft_thresholding}
\hat{ \beta}^{\text{l}}_p=\text{sign} (\hat{\beta}^{\text{LS}}_p)( |\hat{\beta}^{\text{LS}}_p|-\lambda)_+ \quad \forall p \in \{1,\dots,P\}\,.
\end{equation}
\noindent Coefficients whose magnitude is smaller than $\lambda$ are set to zero. Amplitudes of the others are shrunk with respect of the threshold. While proved successful for numerous applications, some drawbacks are reported \cite{Zou_H_2005_j-r-stat-soc-b-stat-methodol_regularization_vsven,Hastie_T_2015_book_statistical_llslg}. Some are: 1) non strict convexity of the criterion when the number of predictors exceeds the number of observations $(P>N)$ 2) algorithm saturation when $N$ variables have been selected 3) with highly correlated variables, tendency to pick mildly representative ones. \\
Another shrinking method is ridge regression \cite{Hoerl_A_1970_j-technometrics_ridge_ranp} with optimization problem: 
\begin{equation}\label{eq_dspls_ridge_opt}
\argmin_{\boldsymbol \beta} \dfrac{1}{2}\|  \vecy - \textbf{X} \boldsymbol \beta\|_2^2 + t \|\boldsymbol \beta \|_2 \,.
\end{equation}
Its trick is to add a diagonal matrix to $(\matX^T \matX)$ in order to overcome the singularity problem. Therefore, the solution always exists, expressed as:
\begin{equation}\label{eq_dspls_ridge_sol}
\hat{ \boldsymbol \beta}^{\text{r}}=(\matX^T \matX + t I_P)^{-1} \matX^T \vecy \,.
\end{equation}
Compared to the lasso, it uses an $\ell_2$-norm instead of the $\ell_1$ penalization but retains most variables by design. 
\subsection{Blending methods: sparse Partial Least Squares (sPLS) \label{sec_dspls_spls}}
Sparse Partial Least Squares (sPLS) denotes a body of works adding a variable selection flavor to the standard PLS framework. We focus here on ones specifically using lasso inspired penalties. An $\ell_1$-norm can be incorporated in optimization problem \eqref{eq_dspls_PLS_opt}. Noting
\begin{equation}\label{eq_dspls_cov}
\widehat{\text{Cov}} (\textbf{X}\textbf{w}, \textbf{y})=\dfrac{1}{N}\textbf{w}^T\textbf{z},\quad \text{with} \quad \textbf{z} = \textbf{X}^T\textbf{y} =N \widehat{\text{Cov}} (\textbf{X},\textbf{y})\,,
\end{equation}
adding the coupling parameter $\lambda_s>0$ and orthogonality constraint on components, we get, for the first component:
\begin{equation} \label{eq_dspls_spls_opt}
\hat{\textbf{w}}=\argmin_{\textbf{w} \in \mathbb{R}^p} \{-\widehat{\text{Cov}} (\textbf{X}\textbf{w}, \textbf{y})+\lambda_s \| \textbf{w} \|_1 \},  \qquad \text{ for } \textbf{w}^T\textbf{w}=1\,.
\end{equation}
\noindent Problem \eqref{eq_dspls_spls_opt} is tackled in 2008 \cite{LeCao_2008_j-stat-appl-genet-mol-biol_sparse_plsvsiod} using sparse PCA \cite{Shen_H_2008_j-multivar-anal_sparse_pcarlrma}. Then iterative PLS \cite{Tenenhaus_M_1998_book_regression_plstp} is combined to singular value decomposition. We denote it as \spls{LeCao} after the first author. In 2010 \cite{Chun_H_2010_j-r-stat-soc-b-stat-methodol_sparse_plsrsdrvs}, Problem \eqref{eq_dspls_spls_opt} is reframed by imposing the $\ell_1$ penalty on a surrogate direction close to the original  vector $\textbf{w}$, providing an approximate solution with \spls{Chun}. In 2018,  \cite{Durif_G_2018_j-bioinformatics_high_dccasplslr} reformulates Problem \eqref{eq_dspls_spls_opt} using recent results from proximal optimization \cite{Bach_F_2012_j-found-trend-mach-learn_optimization_sip} with \spls{Durif}. In this last case, \spls{Durif} provides an exact and closed-form solution reminiscing the soft threshold operator. Moreover, they suggest an adaptive method for computing the sPLS weight vectors using classical PLS ones.\\
Along the lines of methods presented above, Dual-sPLS aims at inference and interpretability: accurate predictions combined with sparse  localization  features for better chemometrics performance. Following \cite{Durif_G_2018_j-bioinformatics_high_dccasplslr}, we also wish to provide a means to tuning the relative sparsity of the outcome. Finally, as different analytical chemistry modalities provide different insights on chemical mixtures, the Dual-sPLS is designed to naturally  allow the combination of heterogeneous datasets as a byproduct of the versatile  dual norm approach\footnote{ Application of this extention is not performed here and is subject to a later work.}. 

\section{Dual Sparse Partial Least Squares (Dual-sPLS)\label{sec_dspls_dspls}}
\subsection{Motivation and purposes\label{sec_dspls_motivation}}
The classical data fidelity $\ell_2$-norm is often used in penalties applications. Arbitrary norm choices may not lead to trackable algorithms. However, the concept of dual norm is a means to formulate a unifying optimization framework with practical algorithmic properties.
\begin{definition} \label{def_dspls_dual_norm}
Let $\Omega (\cdot)$ be a norm on $\mathbb{R}^P$. For any $\textbf{z} \in \R^P$, the associated dual norm, denoted $\Omega^*(\cdot)$, is defined as 
\begin{equation}\label{eq_dspls_dual_norm}
\Omega^*(\textbf{z})=\max_\textbf{w}(\textbf{z}^T\textbf{w}) \quad \mathrm{s.t.} \quad \Omega(\textbf{w})=1\,.
\end{equation}
\end{definition}
\noindent Comparing \eqref{eq_dspls_PLS_opt} and \eqref{eq_dspls_dual_norm}, we find that the optimization of the PLS objective function amounts to finding the vector $\vecw_1$ that fits the dual norm of the $\ell_2$-norm of $\vecz$, where $\vecz=\textbf{X}^T\vecy$. This gives us the incentive to evaluate different norm expressions that could be used as domain-related penalizations. Thus, for any norm $\Omega(.)$ used, the first component will be:
\begin{equation}
  \hat{\textbf{w}}=\argmin_{\textbf{w} \in \mathbb{R}^p} \{-\vecz^T \vecw \},  \quad \text{ s.t. } \Omega(\vecw)=1\,.
\end{equation}
Although formulation is generic, we emphasize four types of norms that make practical sense when dealing with measurements typically available in chemometrics. We provide the corresponding R \cite{RCoreTeam_0_2021_manual_r_lesc} package dual.spls \cite{Alsouki_L_2022_misc_package_dualspls_dsplsr} with a complete description. It contains the following main functions, each of them being associated to specific penalty:
\begin{enumerate}
\item \textbf{\dspls{l}} \emph{(pseudo-lasso norm, d.spls.lasso())}. Similar to the sPLS  Problem \eqref{eq_dspls_spls_opt}, an intuitive norm combines $\ell_2$ and $\ell_1$:
\begin{equation}\label{eq_dspls_l_pseudo_norm}
\Omega (\textbf{w})= \lambda \|\textbf{w}\|_1 + \|\textbf{w}\|_2\,.
\end{equation}
\item \textbf{\dspls{gl}} \emph{(pseudo-group lasso norm, d.spls.GL())}. Inspired by  group lasso \cite{Simon_N_2013_j-comp-graph-stat_sparse-group_l}, it combines groups of measurements.
It applies pseudo-lasso to each group individually while constraining the total set. For $G$ groups, $\textbf{w}_g$ represents the variables of the loading vector $\textbf{w}$ that belongs to group $g$. The corresponding norm is formulated as:
\begin{equation}\label{eq_dspls_gl_pseudo_norm}
\Omega(\textbf{w})=\sum_{g=1}^G \alpha_g \|\textbf{w}_g\|_2+ \lambda_g\|\textbf{w}_g\|_1\,,
\end{equation}
where $\alpha_g \geq 0, \forall g \in \{1, \dots, G\} $ and $\sum_{g \in {1, \dots, G}} \alpha_g=1$.
\item \textbf{\dspls{LS}} \emph{(pseudo-least squares norm, d.spls.LS())}. It introduces \textbf{N}$_1$, a matrix  of $p$ columns, and applies when \textbf{X} is not singular:
\begin{equation}\label{eq_dspls_LS_pseudo_norm}
\Omega (\textbf{w})= \lambda \|\mathbf{N_1w}\|_1 + \|\textbf{Xw}\|_2\,.
\end{equation}
The classical least squares solution is recovered for $\lambda=0$. 
\item\textbf{\dspls{r}}\emph{ (pseudo-ridge norm, d.spls.ridge())}. It deals with cases where $\textbf{X}$ is singular and resorts to a ridge-like penalization:
\begin{equation}\label{eq_dspls_r_pseudo_norm}
\Omega (\textbf{w})= \lambda_1 \|\mathbf{w}\|_1 +\lambda_2 \|\textbf{X}\mathbf{w}\|_2+ \|\textbf{w}\|_2\,.
\end{equation}
\end{enumerate}
\noindent The construction of weight vectors $\vecw_1, \dots, \vecw_M$ differs in each of the four cases. It however follows similar steps as for the PLS. Starting with a reformulation of optimization Problem \eqref{eq_dspls_dual_norm} and using  Lagrange multipliers, we aim at iteratively minimizing the function $L(\vecw)=-\vecz^T \vecw + \mu (\Omega(\vecw)-1)$, for $\mu>0$. As some norms are not differentiable, we resort to the more generic notion of subgradient $\nabla\Omega(\vecw)$ \cite{Bach_F_2012_j-found-trend-mach-learn_optimization_sip}. It identifies to the classical differential when it is defined. The subgradient of $L$ vanishes for 
\begin{equation}\label{eq_dspls_norm_diff}
\nabla \Omega(\textbf{w}) = \dfrac{\textbf{z}}{\mu}\,.
\end{equation}
It is then sufficient to substitute the gradient --- when it exists--- of the considered norm of $\Omega(\vecw)$ in \eqref{eq_dspls_norm_diff}. \\
We provide in the following a detailed analysis for the pseudo-lasso case of Dual-sPLS (see \eqref{eq_dspls_l_pseudo_norm}) and some remarks for the other norms. In all cases we impose that $\vecw$ and $\vecz$ lie in the same orthant; it generalizes, in $n$ dimensions, the quadrant in the 2D plane or the octant in the 3D space. In other words, corresponding coordinates of $\vecw$ and $\vecz$ have the same sign. 
\subsection{Norm options (lasso, group lasso, least squares and ridge)\label{sec_dspls_norm}}
\subsubsection{Pseudo-lasso \label{sec_dspls_dspls_l}}
\begin{figure}[H] 
    \centering
    \includegraphics[width=0.7\textwidth]{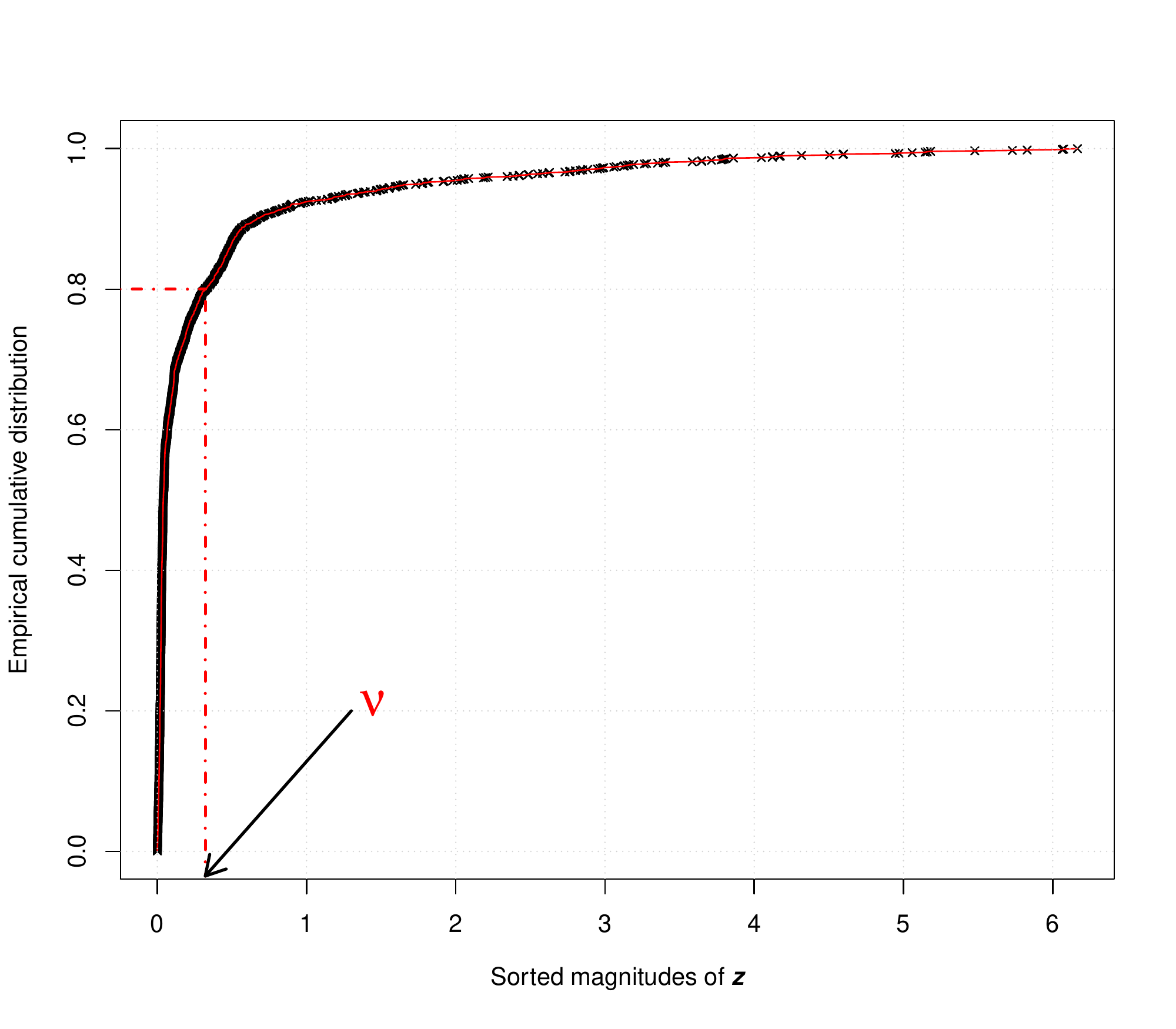}
    \caption{Empirical cumulative distribution of the sorted magnitude of $\vecz=\textbf{X}^T\textbf{y}$ from real data \DNIR to control sparsity.}
     \label{fig2.2.1}
\end{figure}

We rconsider Equation \eqref{eq_dspls_l_pseudo_norm}. Let $\delta$ be the sign vector of $\vecw$ and $\vecz$. By differentiating $\Omega(\textbf{w})$, we get
 \begin{equation}\label{eq_dspls_norm_diff_bis}
\nabla \Omega(\textbf{w}) = \lambda \boldsymbol \delta + \dfrac{w}{\|w\|_2}\,,
\end{equation}
\noindent and by substituting it in \eqref{eq_dspls_norm_diff}, we obtain
\begin{equation}\label{eq_dspls_diff_equality} 
\dfrac{\textbf{w}}{\|\textbf{w}\|_2}=\dfrac{\textbf{z}}{\mu}-\lambda \boldsymbol \delta\,.
\end{equation} 
\noindent The closed-form solution of the \dspls{l} optimization problem consists in zeroing coordinates whose magnitude is lower than the soft threshold $\lambda$ and in reducing the others toward zero. Thus, for $\nu=\lambda \mu$ and $ p \in \{1, \dots, P\}$, it can be expressed as:
\begin{equation}\label{eq_dspls_l_sol}
\dfrac{w_p}{\|\textbf{w}\|_2}=\dfrac{1}{\mu}\delta_p (|z_p|-\nu)_+\,.
\end{equation}
\noindent A common issue is the choice of the appropriate shrinking parameter. Cross-Validation \cite{Stone_M_1974_j-r-stat-soc-b-stat-methodol_cross-validation_casp}, evoked in Section \ref{sec_dspls_model_settings}, is popularly adopted in sparse regressions. We choose a more intuitive option. We obtain it adaptively, according to the proportion of variables that we would like to keep in the active set at each iteration. The procedure is illustrated in Figure \ref{fig2.2.1}. It represents the empirical cumulative distribution of sorted magnitudes of $|\textbf{X}^T\textbf{y}|$ from the real data \DNIR described later in Section (\ref{sec_dspls_DNIR}). Fixing a shriking ratio $\varsigma$ of expected zero coefficients (e.g. $\varsigma=\SI{80}{\percent}$), we select the threshold $\nu$ at iteration $m$ as depicted. As the cumulative destribution is non-decreasing, we choose the first $x$-axis value corresponding to  ordinate \num{0.8}. 
\noindent To guarantee the unit norm property for $\vecw$, we set $\mu=\|\textbf{z}_\nu\|_2$ where $\textbf{z}_\nu$ is the vector of coordinates $\delta_p (|z_p|-\nu)_+$ for $p \in \{1, \dots, P\}$. Consequently, $$\textbf{w} =\dfrac{\mu}{\nu \|\textbf{z}_\nu\|_1 + \|\textbf{z}_\nu\|_2^2}\textbf{z}_\nu.$$\\
The rationale behind constrainting the direction $\vecw$ instead of the regression coefficients $\hat{\boldsymbol \beta}$ is their collinearity. Indeed, the estimator writes $\hat{\boldsymbol \beta}=\textbf{W}(\textbf{T}^T\textbf{T})^{-1}\textbf{T}^T\vecy$. Being collinear, soft-thresholding $\vecw$ performs a variable selection at the same location in $\hat{\boldsymbol \beta}$ coordinates. The pseudo-lasso Dual-sPLS is described in Algorithm \ref{algo_dspls_l}.
\begin{algorithm}[H]
\caption{ \Dspls{l} }
\begin{algorithmic} \label{algo_dspls_l}
\STATE Input: $\textbf{X},\textbf{y},M\text{ (number of components desired)},\varsigma \text{  (shrinking ratio)}$
\STATE $\textbf{X}_1=\textbf{X}$
\FOR{ $ m= 1, \dots, \text{M}$ }
\STATE $\textbf{z}_m=\textbf{X}_m^T\textbf{y}$ (weight vector)
\STATE Find $\nu$ adaptatively according to $\varsigma$
\STATE$\textbf{z}_\nu= (\delta_p (|z_p|-\nu)_+)_{p}$ (applying the threshold), $p \in \{1, \dots, P\}$
\STATE$\mu=\|\textbf{z}_\nu\|_2 $ and $\lambda=\dfrac{\nu}{\mu }$
\STATE $\textbf{w}_p=\dfrac{\|\textbf{z}_\nu\|_2}{\nu \|\textbf{z}_\nu\|_1 + \|\textbf{z}_\nu\|_2^2}\textbf{z}_\nu $ (loadings)
\STATE $\textbf{t}_m=\textbf{X}_m\textbf{w}_m$ (component)
\STATE $\textbf{X}_{m+1}=\textbf{X}_m-\mathcal{P}_{\textbf{t}_m}\textbf{X}_m$ (deflation)
\ENDFOR
\STATE $\hat{\boldsymbol \beta}=\textbf{W}(\textbf{T}^T\textbf{T})^{-1}\textbf{T}^T\vecy$
\end{algorithmic}
\end{algorithm}
\noindent Note that as long as $\vecw$ and $\textbf{z}_\nu$ are collinear, the sparsity of the results only requires the computation of $\vecw$, up to a non-zero factor. 
\subsubsection{Pseudo-group lasso \label{sec_dspls_dspls_gl}}
Response $\textbf{y}$ may be explained separately  by explanatory variables of different nature with prediction models. Combining them appropriately is potentially beneficial both in predictive and interpretative powers. The same reasoning could be used to  partition the dataset into groups.\\
Physico-chemical motivation resides in segmenting a spectrum into homogenous bands or combining complementary modalities (e.g. IR and NMR) to predict the same property (e.g. viscosity, density). We consider $G$ groups, and $\textbf{z}_g$ sub-vector of  $\textbf{z}$ denotes variables belonging to group $g$. The group lasso inspired norm is expressed as in Equation \eqref{eq_dspls_gl_pseudo_norm}. The closed-form solution is collinear to the vector $\textbf{z}_{\nu_g}$. It is given by 
\begin{equation}\label{eq_dspls_gl_znug}
 \textbf{z}_{\nu_g} = \boldsymbol \delta_g (| \textbf{z}_g | - \nu_g)_+  \quad \text{and } \textbf{z}_\nu=\big( \textbf{z}_{\nu_g} \big)_{g  \in \{ 1, \dots, G\}}\,,
\end{equation}
$\boldsymbol\delta_g$ being the vector of signs of $\vecw_g$ and $\nu_g=\lambda_g \mu$ for $g  \in \{ 1, \dots, G\}$. Each group is driven by its own threshold $\nu_g$. The latter can be obtained similarly as in Section \ref{sec_dspls_dspls_l}. Note that this Dual-sPLS version reduces to the pseudo-lasso case when $G=1$.
\subsubsection{Pseudo-least squares and  pseudo-ridge \label{sec_dspls_dspls_ls_r}}
The above can be generalized in many ways, by defining more versatile norm shapes, including notably weighted norms. One such possibility is $\forall \vecw \in \R^P$
\begin{equation}\label{eq_dspls_pseudo_norm}
\Omega(\vecw)=\lambda_1 \| \textbf{N}_1 \vecw \|_1 + \| \textbf{N}_2 \vecw \|_2 + \lambda_2 \| \vecw \|_2\,.
\end{equation}
It is not easily solvable in general. However, an appropriate choice of matrices $\textbf{N}_1$ and $\textbf{N}_2$, and factors $\lambda_1$ and $\lambda_2$ allow us to recover the lasso and group lasso norms, but also several other already known concepts, like fused lasso, least squares or ridge. We focus here on two main situations whose optimization problem resolution can be obtained analytically. An obvious option heavily inspired by least squares regression sets $\textbf{N}_2=\textbf{X}$ and $\lambda_2=0$. Its resolution supplements the traditional least squares problem with a more selective shrinkage akin to that of our pseudo-lasso. Namely we first note 
\begin{equation}\label{eq_dspls_LS_pseudo_norm_2}
\Omega (\textbf{w})= \lambda \|\mathbf{N_1w}\|_1 + \|\textbf{Xw}\|_2\,.
\end{equation}
\noindent Then for $\nu=\mu \lambda$ and $\delta$ the vector of signs of $\mathbf{N_1}\vecw$ and $\mathbf{N_1}\vecz$, 
\begin{equation}\label{eq_dspls_LS_sol}
\dfrac{\textbf{w}}{\| \textbf{X} \textbf{w}\|_2}=(\textbf{X}^T \textbf{X})^{-1} \dfrac{\textbf{z}}{\boldsymbol \mu}-\lambda (\textbf{X}^T \textbf{X})^{-1} \textbf{N}_1^T \boldsymbol \delta\,,
\end{equation}
where we have implicitly assumed that \matX has full rank.
\noindent Consequently, we penalize $| \hat{\boldsymbol \beta}^{\text{LS}} |$ instead of $| \vecz |$. For equation \eqref{eq_dspls_LS_sol} to take a genuine pseudo-lasso form, it is sufficient that $\textbf{N}_1$ verifies 
\begin{equation}\label{eq_dspls_N1_condition}
(\textbf{X}^T \textbf{X})^{-1} \textbf{N}_1^T \boldsymbol \delta= \text{sign} \bigg( (\textbf{X}^T \textbf{X})^{-1} \textbf{z} \bigg)\,.
\end{equation} 
However, as it does not play a role in loadings' computation, it does not need to be computed explicitly. Thus, the coordinates of the simplified closed-form solution is:
 \begin{equation}\label{eq_dspls_LS_sol_bis}
\dfrac{\textbf{w}_p}{\| \textbf{X} \textbf{w}\|_2}= \dfrac{1}{\mu} \text{sign}(\hat{\beta}^{\text{LS}}_p) ( | \hat{\beta}^{\text{LS}}_p | - \nu )_+\,,
\end{equation}
\noindent When $\textbf{X}$ is singular, the above cannot hold. Meanwhile, this case can be addressed with a regularization inspired by the ridge \cite{Hoerl_A_1970_j-technometrics_ridge_ranp}. By choosing $\textbf{N}_1=I_P$, $\textbf{N}_2=\lambda_2 \matX$ and $\lambda_2=1$, equation \eqref{eq_dspls_pseudo_norm} writes 
\begin{equation}\label{eq_dspls_r_pseudo_norm_2}
\Omega (\textbf{w})= \lambda_1 \|\mathbf{w}\|_1 +\lambda_2 \|\textbf{X}\mathbf{w}\|_2+ \|\textbf{w}\|_2\,.
\end{equation}
It amounts to penalize $| \textbf{z}_{\nu_2}|$ where $\textbf{z}_{\nu_2}=\bigg( \nu_2 \textbf{X}^T\textbf{X} + I_P \bigg)^{-1}$ and $\nu_2= \lambda_2 \dfrac{\textbf{w}}{\|\textbf{X}\textbf{w}\|_2}$, instead of $| \vecz |$ like in the pseudo-lasso. For $\nu_1= \lambda_1 \mu$, the closed-form solution is formulated as:
\begin{equation}\label{eq_dspls_r_sol}
\dfrac{\textbf{w}}{\|\textbf{w}\|_2}=\dfrac{1}{\mu} \delta ( | \textbf{z}_{\nu_2}| - \nu_1 )_+\,.
\end{equation}
\noindent where  $\delta= \text{sign}(\textbf{z}_{\nu_2} \textbf{z})$. Adding the diagonal perturbation resolves the non-invertability of $\textbf{X}^T\textbf{X}$.

\section{Simulated and real data, model settings, evaluation \label{sec_dspls_data_model_evaluation}}
\subsection{Simulated sparse data: Gaussian mixtures \Ds and \Dsb \label{sec_dspls_DSIM}}
\begin{figure}
	\centering
	\begin{tabular}{cc}
		\begin{subfigure}{0.49\textwidth}
			\centering
			\includegraphics[width=\textwidth]{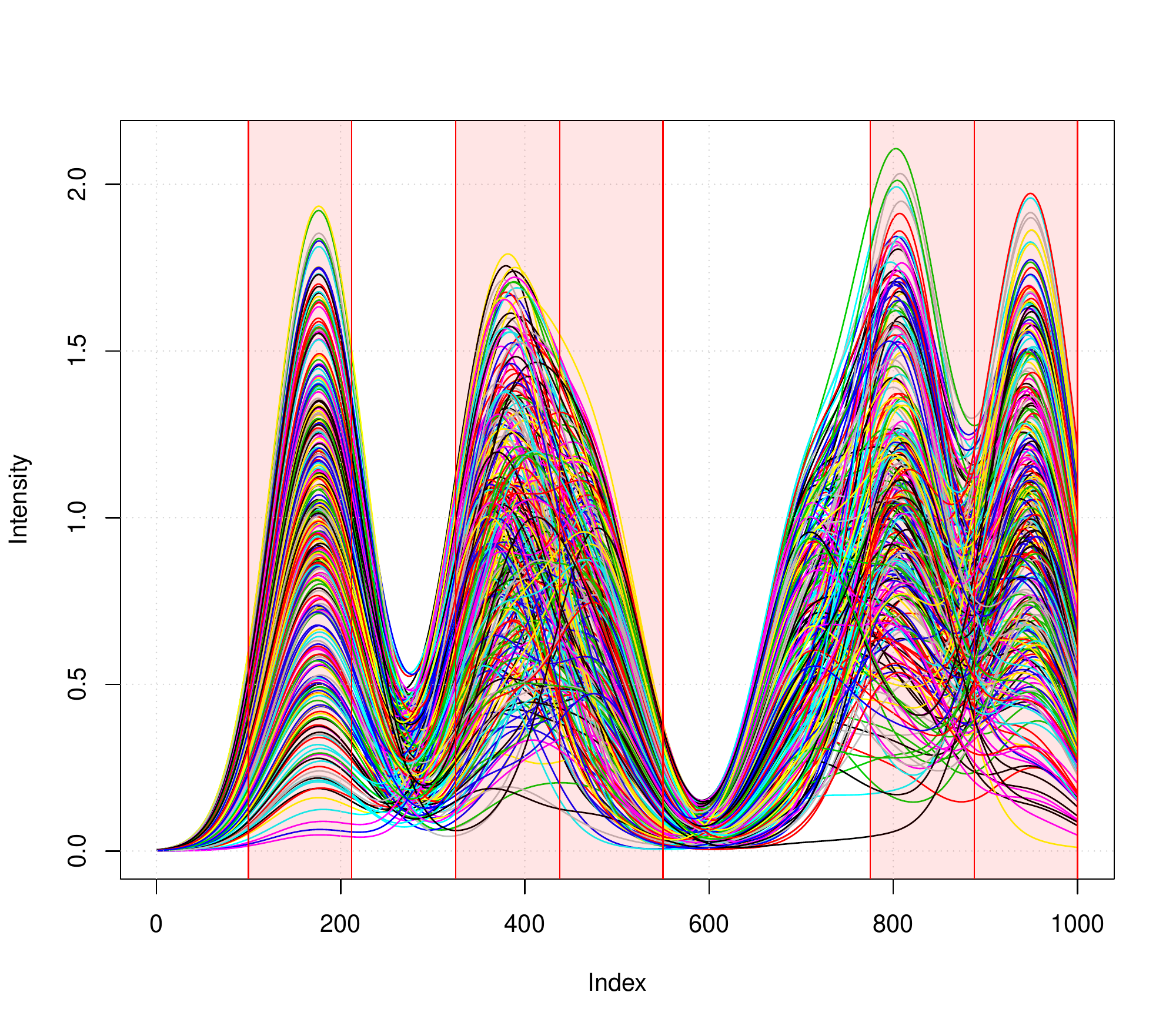}
		\end{subfigure}%
		\hfill
		\begin{subfigure}{0.49\textwidth}
			\centering
			\includegraphics[width=\textwidth]{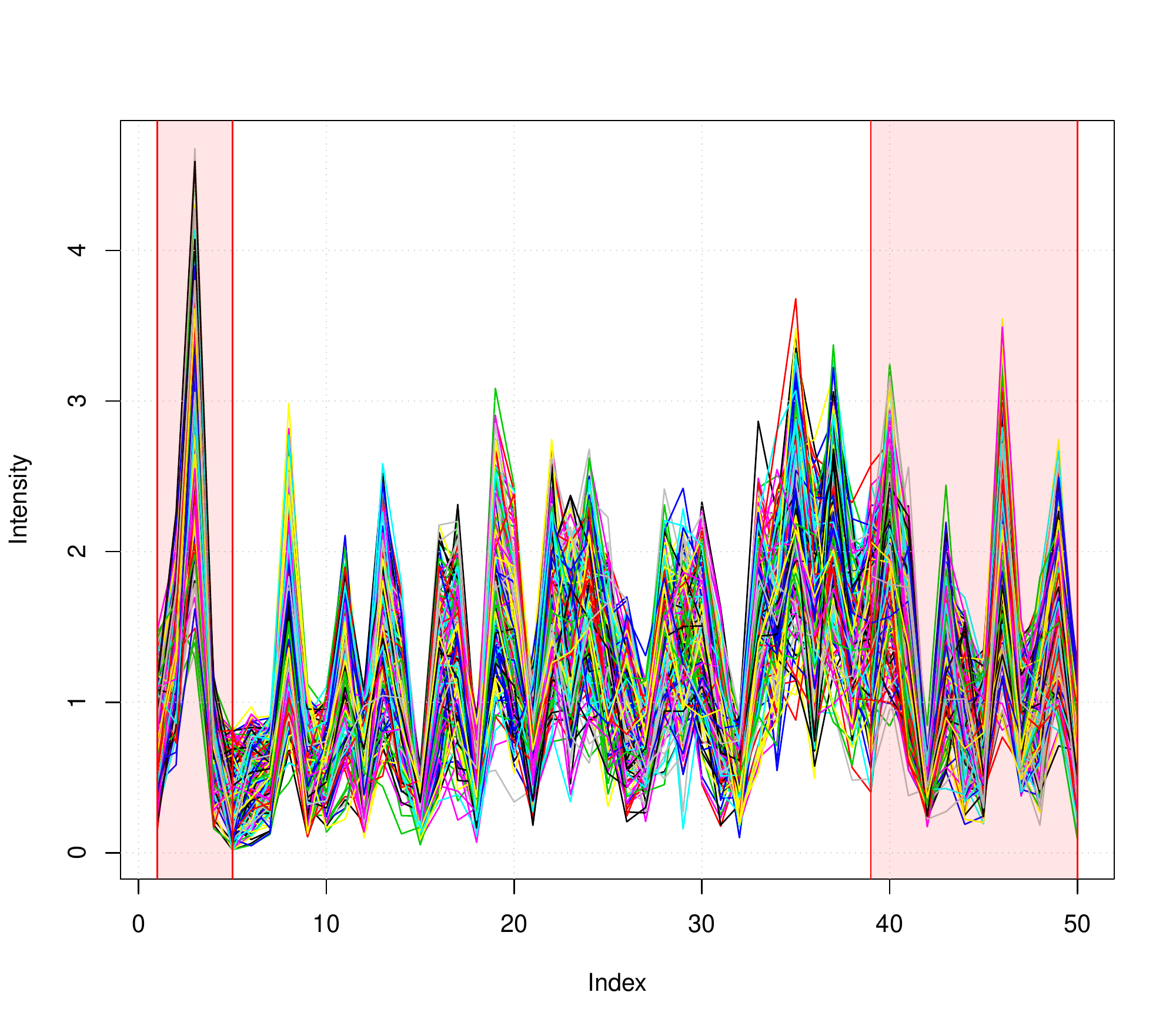}
		\end{subfigure}%
	\end{tabular}
         \caption{\Ds (left) and \Dsb (right) simulated data. }%
    \label{fig_dspls_DSIM}%
\end{figure}

For an in-depth analysis of machine learning algorithms, resorting to simulated data allows an unbiased access to ground truth. We thereby propose a parametrized model. It is thought to provide similarities with common analytical chemistry data, with all sparse parameters controlled. We choose a positively weighted mixture of $K$ Gaussians peaks with preset scale $\sigma^2$ and randomly picked amplitudes $A$ and locations $\mu$. They are summed as follows and uniformly sampled:
\begin{equation}
\sum_{k=1}^K A_k\exp(-\dfrac{x-\mu_k}{2\sigma^2}) \,.
\end{equation}
They are affected by a stochastic Gaussian contamination. The response vector $\textbf{y}$ is defined by an explicit linear model composed of weighted sums of $\textbf{X}$ values. Weights can be random or fixed quantities by range of indices. \\
In this work, to evaluate Dual-sPLS in both precision and information location, we devise a sparse additive model with only $S\ll P$ positive weights and $P-S$ null weights. Namely, only $S$ variables are responsible in the construction of response $\vecy$. This information is especially  beneficial to demonstrate the strength of variable selection in sparse methods. Since we deal with high-dimensional  situations, we simulated \Ds: 300 mixtures of 30 Gaussians represented by 1000 variables (Figure \ref{fig_dspls_DSIM} (left)). Highlighted red areas denote variables involved in the computation of response $\vecy$. The corresponding matrix of \Ds is singular and used in the evaluation of \dspls{l} and \dspls{r}. Since the \dspls{LS} is only operational with invertible matrices, we also simulated  non-singular data matrix \Dsb, 200 mixtures of 100 Gaussians represented by 50 variables. The response $\vecy$ corresponding to \Dsb depends only on the first five and last twelve variables as shown in Figure \ref{fig_dspls_DSIM} (right). 
\subsection{Real data: near-infrared (NIR) spectroscopy \DNIR \label{sec_dspls_DNIR}}
\begin{figure}[H]
    \centering
    \includegraphics[width=0.6\textwidth]{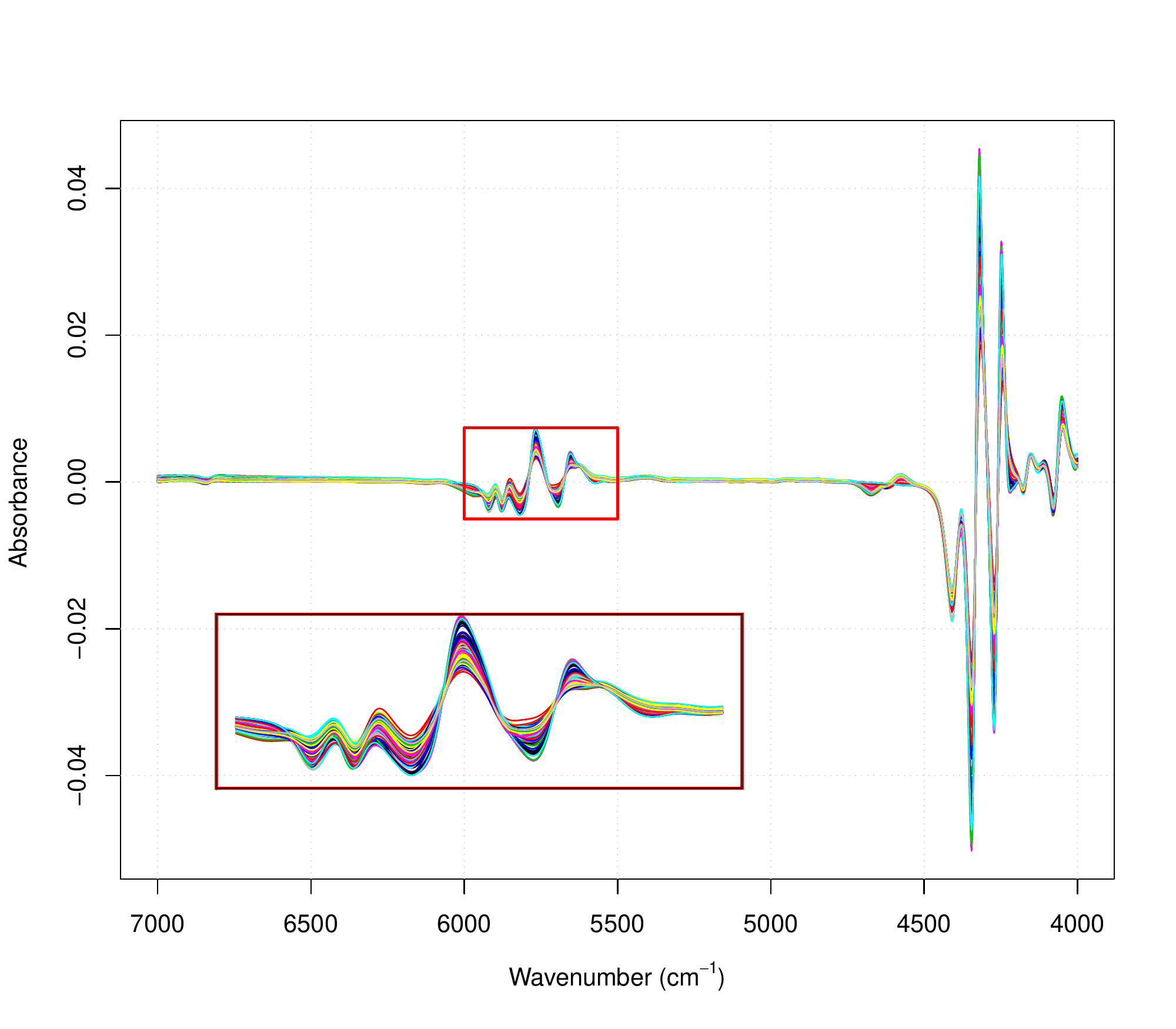}
    \caption{\DNIR: first derivative of the NIR spectra of 208 samples.}
     \label{fig_dspls_DNIR}
\end{figure}

In chemistry, complex mixtures of molecules are analyzed with different physico-chemical methods. Besides, determining macroscopic properties is important to their use. \\
The evaluation on real data is done using NIR spectra of hydrocarbon samples.
NIR is based on the principle of absorption of radiation (infrared) by  matter \cite{Chalmers_J_2002_book_handbook_vs}. Infrared radiations correspond to wavenumbers directly greater than those of the visible light spectrum. 
The absorption of radiation depends on chemical bonds, therefore a NIR spectrum encodes information about the composition of the sample. We focus on the density property which is obtained by standardized methods. 
The IFPEN dataset \DNIR was partly exposed in \cite{Laxalde_J_2011_j-anal-chim-acta_characterization_hounisoppmvs,Laxalde_J_2012_phd_analyse_plpsi}. It is available at \url{http://www.laurent-duval.eu} and subject to a forthcoming publication \cite{Duval_L_2023_PREPRINT_ifpen_nisdpp208nirhsdr}. It is composed of 208 samples with 1557 variables. The corresponding matrix \matX is singular. Many chemical data require adequate preprocessing: normalization, baseline removal \cite{Ning_X_2014_j-chemometr-intell-lab-syst_chromatogram_bedusbeads}, deconvolution \cite{Cherni_A_2020_j-ieee-tsp_spoq_lpolqrssrams}. Here we simply apply a discrete derivative  obtained with a Savitzky--Golay smoothing filter  \cite{Savitzky_A_1964_j-anal-chem_smoothing_ddslsp} of degree 2 and length 15. It serves as both a crude baseline filter and diversity enhancement operator \cite{DeNoyer_L_2002_incoll_smoothing_ds}. The NIR  preprocessed dataset \DNIR is represented  in Figure \ref{fig_dspls_DNIR}. 
\subsection{Model settings: number of latent component selection \label{sec_dspls_model_settings}}
Selecting the appropriate number $M$ of latent components is crucial when building a regression model. It balances between model complexity and prediction accuracy (degrees of freedom), preventing the risk of overfitting. This issue is especially important when using PLS and its extensions in chemometrics. The commonly used procedure is based on cross validation. \\
First, observations are split randomly several times into a calibration and a validation set. Second, candidate models using the calibration sets are constructed with different numbers of latent components. Third, each prediction is evaluated on the validation set with MSE. The latter are averaged for each model. Finally, the smallest one with the lowest averaged MSE reveals an adequate number of latent components.
\subsection{Calibration and validation\label{sec_dspls_calval}}
The evaluation of prediction models traditionally divides the dataset into two representative sets called calibration and validation. Three main methods are used. In the first one, observations are randomly selected. The second only considers the distribution of values in the response $\textbf{y}$ \cite[Stratified sampling]{Sarndal_C_2003_book_model_ass}. 
A third class is known as Kennard and Stone method \cite{Kennard_R_1969_j-technometrics_computer_ade}. It optimizes relative distances between observations according to variables of $\textbf{X}$. In chemometrics, one may expect the existence  of a yet unknown dependence between analytical measurements  and properties. Taking both $\textbf{X}$ and $\textbf{y}$ values for a proper calibration and validation split would be desirable. The attempts of \cite{Tian_H_2018_j-infrared-phys-technol_weighted_spxymcsscabnis} to consider $\textbf{X}$ and $\textbf{y}$ in a single distance with appropriate weights is not straightforward. It is difficult to adequately  weight variables that do not belong to the same space. We have recently proposed a CalValXy for that purpose. It consists in dividing the dataset into subgroups according to the repartition of $\textbf{y}$ and applying the Kennard and Stone to each subgroup. It is summarized in Algorithm \ref{algo_dspls_calval}, and extensively described in \cite{Alsouki_L_2022_PREPRINT_well-balanced_scvsubpxry}.
\begin{algorithm} [H]
\caption{ Calibration and validation CalValXy} 
\begin{algorithmic} \label{algo_dspls_calval}
\STATE Input: $\textbf{X},\textbf{X}_{type} \text{(index of which set belongs each observation of \textbf{X})},$
\STATE $\textbf{Listecal} \text{ (number of calibration points to pick from each subset)}$
\STATE $G=\text{mean}(\textbf{X}) $  (centroid)
\STATE C$_1=\max_n \|\textbf{x}_G-\textbf{x}_n\|,$ $n \in \{1,\dots,N\}$ (first calibration point)
\STATE$ s =$  subset where C$_1$ is located
\WHILE {\textbf{Listecal} is not empty}
\STATE $s \leftarrow s+1$
\STATE Find the minmax point C in subset $s$ 
\STATE Remove C from \textbf{X} and \textbf{Listecal}
\STATE Store C in a vector of calibration index \textbf{cal}
\ENDWHILE
\end{algorithmic}
\end{algorithm}

\section{Comparative evaluation and discussion \label{sec_dspls_results}}
We benchmark each proposed Dual-SPLS regression flavor (respectively pseudo-lasso, least squares and ridge) against its classical counterpart, and comparable sparse SPLSs, when applicable.
We follow a common procedure to state the main results. First, we split observations into calibration (\SI{80}{\percent}) and validation (\SI{20}{\percent}). We replaced the traditional  Kennard and Stone method (KS) \cite{Kennard_R_1969_j-technometrics_computer_ade}  --- using explanatory variables  \textbf{X} only --- with CalValXy (cf. Section \ref{sec_dspls_calval} and \cite{Alsouki_L_2022_PREPRINT_well-balanced_scvsubpxry}). The latter  incorporates  the response variable $\vecy$ in the splitting and proves slightly better than KS. Comparative performance is assessed in both accuracy and quality of interpretation. For the first one, common objective metrics are  root mean squared error (RMSE),  mean absolute error (MAE), or determination coefficient (R$^2$) (see end of Section \ref{sec_dspls_introduction}). As metrics yield similar outcomes, we only compare, in the topmost figures, RMSE values for either calibration (left) or validation (right), as we increase the number $M$ of latent components from one to ten. For the second one, we assess both variable selection and localization by vertically stacking regression coefficients for each compared algorithm in the bottom figure.
Results are extensively discussed   on simulated and real data for Dual-SPLS\textsubscript{l}, and in less details for the least squares and ridge flavors. Complementary outcomes are provided in the supplementary materials.
\subsection{Dual-sPLS pseudo-lasso evaluation (\Ds, \DNIR) \label{sec_dspls_l_eval} }
\begin{figure}
    \centering
    \begin{tabular}{cc}
    \begin{subfigure}{0.49\textwidth}
        \centering
        \includegraphics[width=\textwidth]{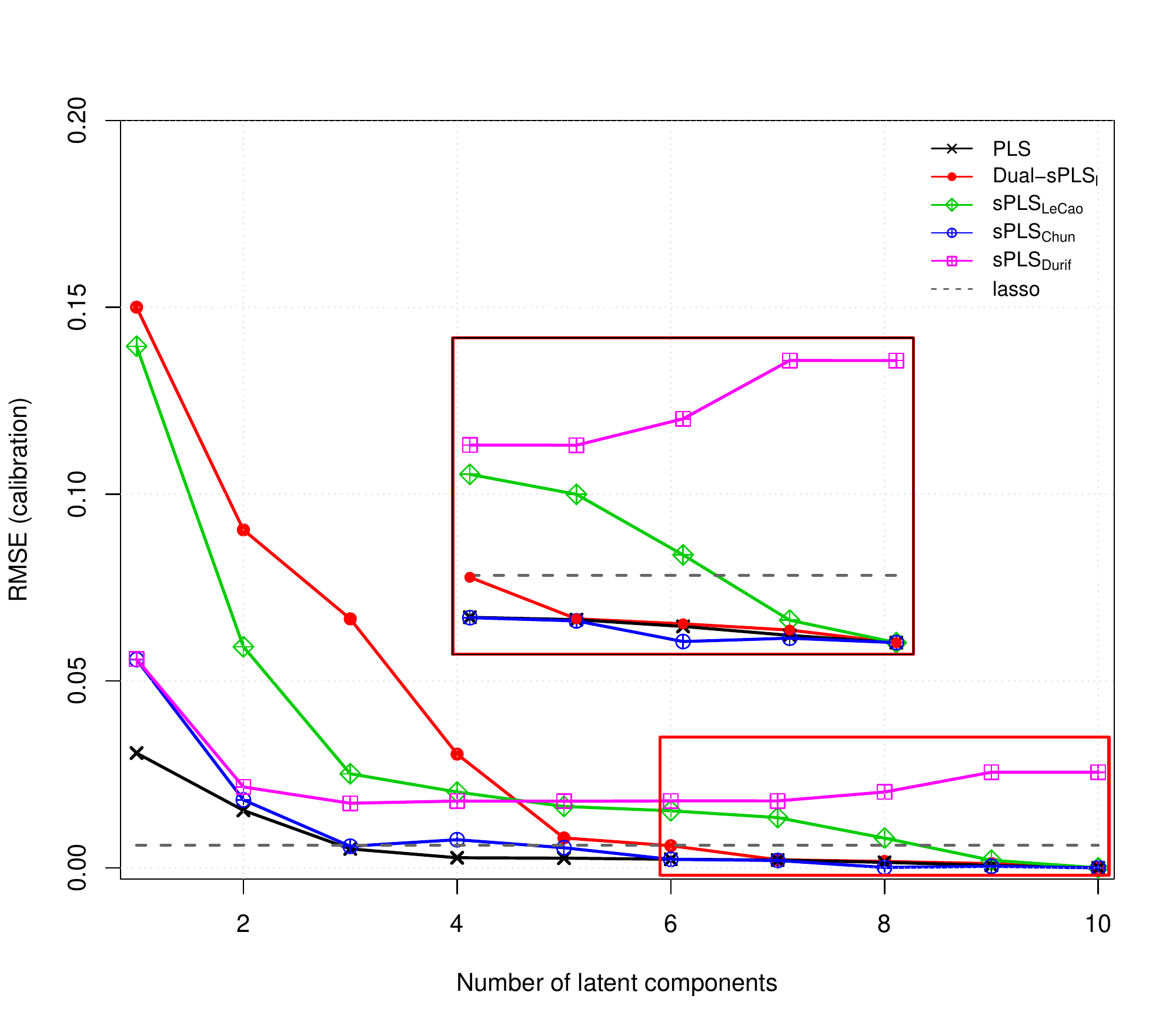}
    \end{subfigure}
    \hfill
        \begin{subfigure}{0.49\textwidth}
        \centering
       \includegraphics[width=\textwidth]{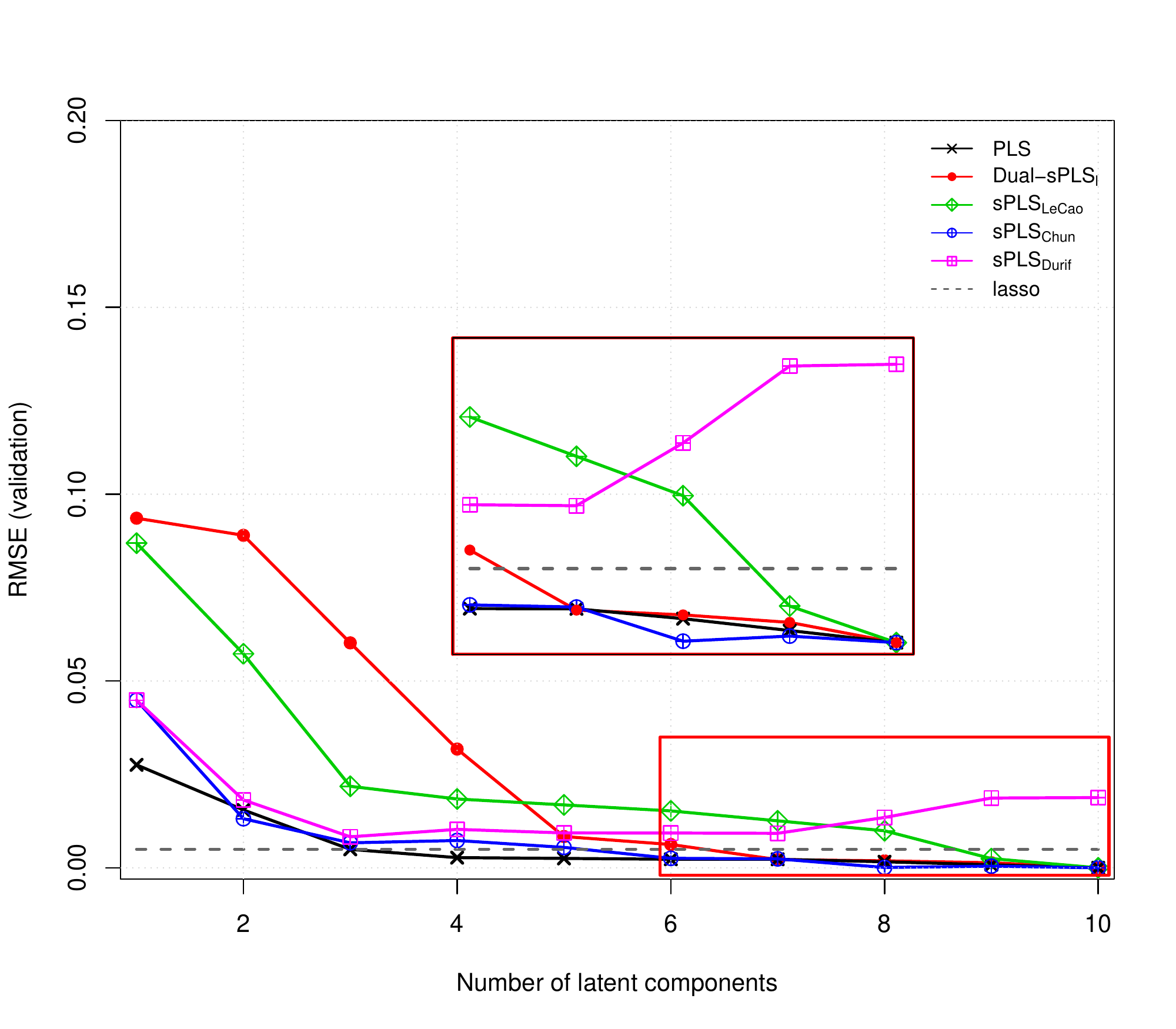}
    \end{subfigure}
    \\
    \begin{subfigure}{0.95\textwidth}
        \centering
        \includegraphics[width=\textwidth]{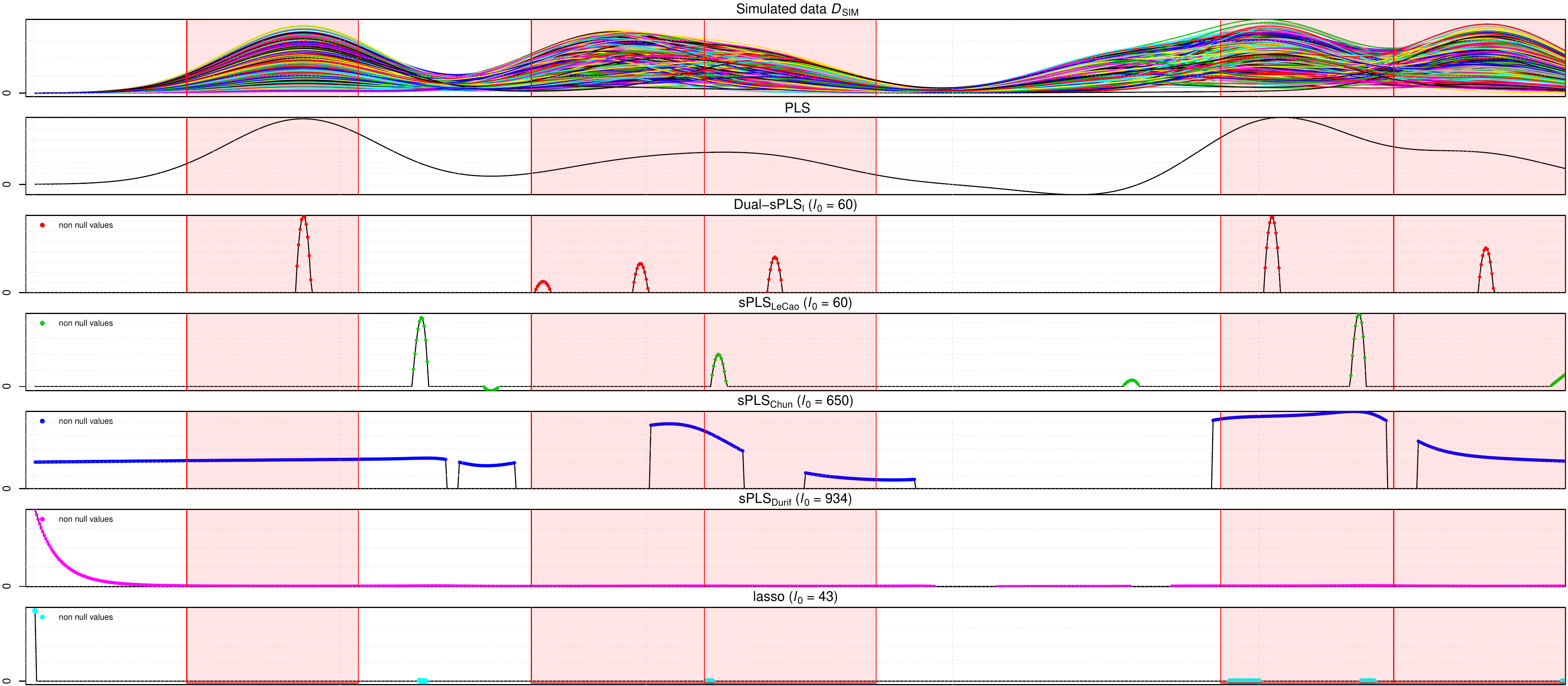}
    \end{subfigure}
    \end{tabular}
    \caption{\dspls{l} evaluation on simulated data \Ds. (Top) RMSE values for calibration (left) and validation (right) with respect to the number of latent components. (Bottom) From top to bottom:  simulated data \Ds, regression coefficients of PLS,  \dspls{l}, \spls{LeCao}, \spls{Chun}, \spls{Durif} for six components, and  lasso.} 
    \label{fig_dspls_l_DSIM}
\end{figure}

\begin{figure}
	\centering
	\begin{tabular}{cc}
		\begin{subfigure}{0.49\textwidth}
			\centering
			\includegraphics[width=\textwidth]{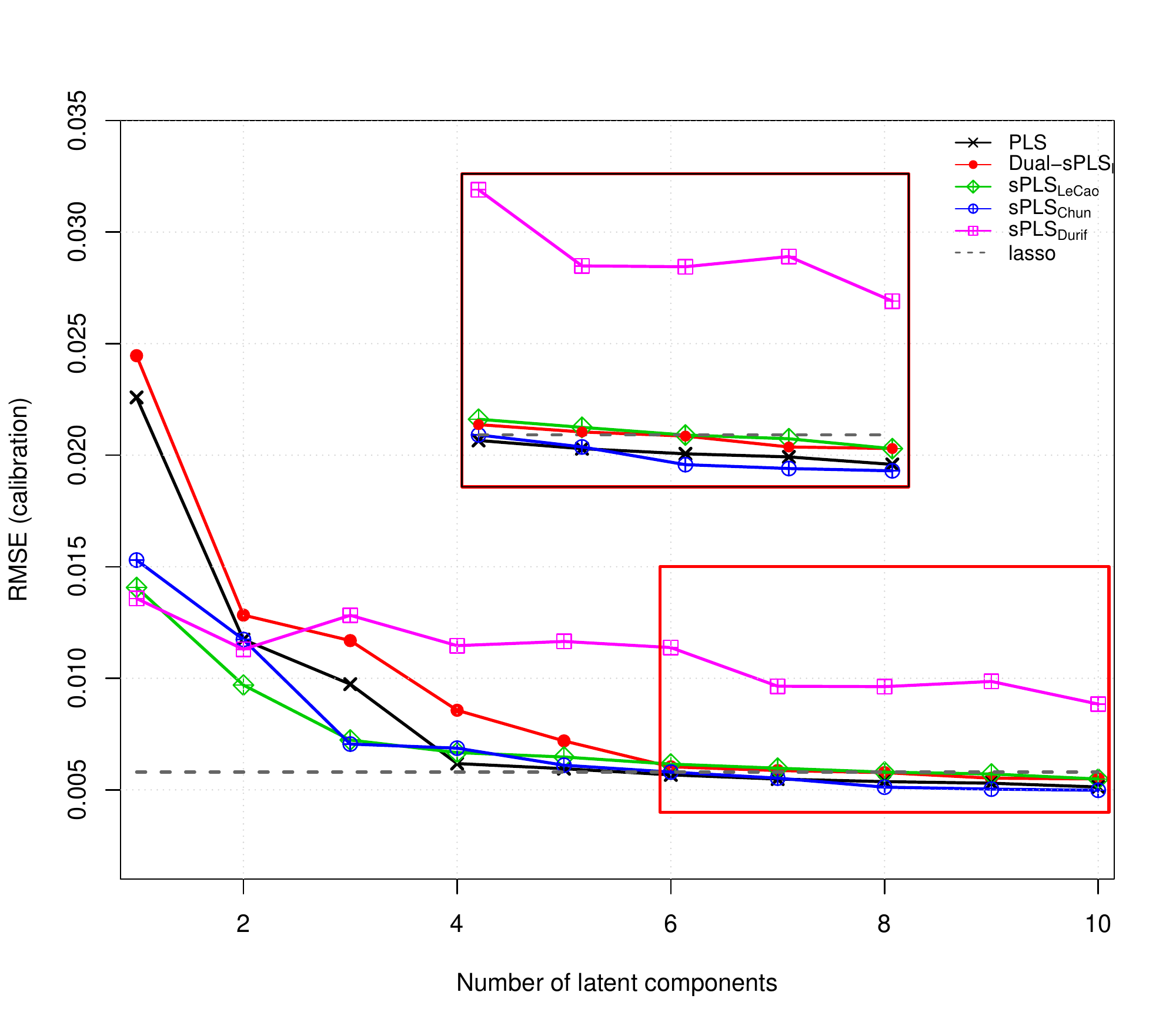}
		\end{subfigure}
		\hfill
		\begin{subfigure}{0.49\textwidth}
			\centering
			\includegraphics[width=\textwidth]{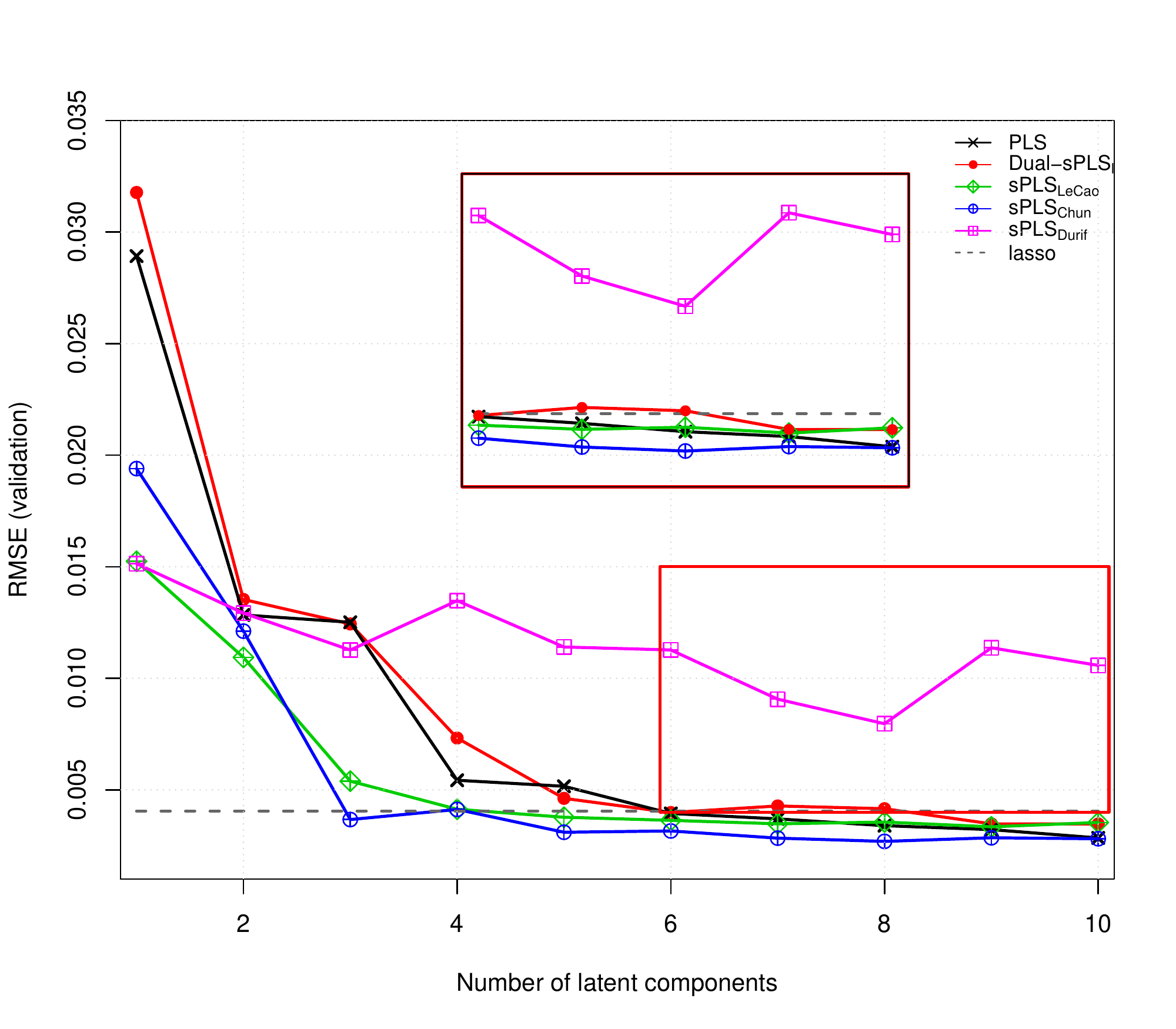}
		\end{subfigure}
		\\
			\begin{subfigure}{0.95\textwidth}
				\centering
				\includegraphics[width=\textwidth]{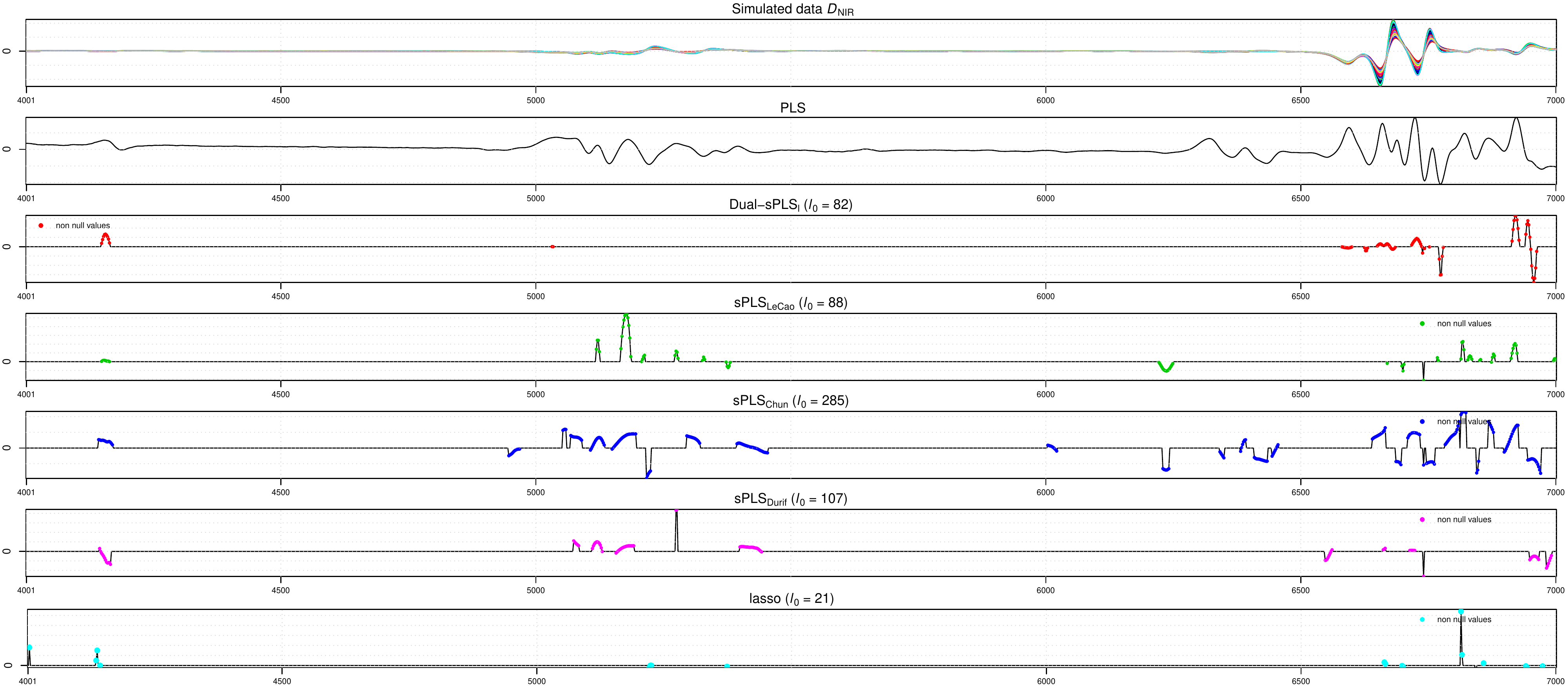}
			\end{subfigure}
	\end{tabular}
    \caption{\dspls{l} evaluation on real data \DNIR. (Top) RMSE values for calibration (left) and validation (right) with respect to the number of latent components. (Bottom) From top to bottom:  original data \DNIR, regression coefficients of PLS,  \dspls{l}, \spls{LeCao}, \spls{Chun}, \spls{Durif} for six components, and  lasso.} 
	\label{fig_dspls_l_DNIR}
\end{figure}

\dspls{l} is compared to standard PLS, three alternative sparse PLS (\spls{LeCao} \cite{LeCao_2008_j-stat-appl-genet-mol-biol_sparse_plsvsiod}, \spls{Chun} \cite{Chun_H_2010_j-r-stat-soc-b-stat-methodol_sparse_plsrsdrvs}, \spls{Durif} \cite{Durif_G_2018_j-bioinformatics_high_dccasplslr}) and lasso \cite{Tibshirani_R_1996_j-r-stat-soc-b-stat-methodol_regression_sslasso}. Their respective parameters are selected by cross validation (Section \ref{sec_dspls_DSIM}). Both \spls{LeCao} and \dspls{l} explicitely specify a sparsity parameter: the (approximate) proportion of variables $\varsigma$ to be discarded ($\lzc/P$). We set it here  to \SI{99}{\percent}.\\
We first evaluate \dspls{l} on simulated data  \Ds (Section \ref{sec_dspls_DSIM}) in  Figure \ref{fig_dspls_l_DSIM}. Top-left and right plots entail that accuracy (RMSE) globally improves as the number of latent variables $M$ increases for all five PLS-related methods --- in both calibration and validation. The lasso performance, independent on the number of components, is represented by the sixth dotted curve. From six to ten latent variables, 
all curves tend to plateau, with close RMSE values. Dual-SPLS\textsubscript{l},  \spls{Chun} and PLS provide the best results (lowest curves). Thus, adding more components seems uncessary. We choose six latent variables to compare coefficient localization. On Figure  \ref{fig_dspls_l_DSIM}-bottom, we stack seven panels:  original spectra (1) and the coefficients for: PLS (2), \dspls{l} (3), \spls{LeCao} (4), \spls{Chun} (5), \spls{Durif} (6), lasso (7). PLS  coefficients (panel 2) match the shape of the simulated data (panel 1). However, it fails to localize the most important variables, unlike sparse PLS. The \lz criterion (Section \ref{sec_dspls_introduction}) quantifies the sparsity induced by each method. \Dspls{l},  \spls{LeCao} and  lasso  perform best, selecting as expected  a small number of variables, with an \lz value around \numrange{40}{60}. It however is not sufficient to hint at improvements in interpretability. Looking only at variables affecting the response (shaded red background in panel 1), most compared methods exhibit significant coefficients in many (useless) areas (transparent background). Only \dspls{l},  \spls{LeCao} present concentrated coefficients that can help chemical interpretation. On this rudimentary yet explainable  model, we hint that  \dspls{l} provides a predictive quality comparable to its challengers, and is the best in providing  at the same time accurate localization on simulated data, with a verifiable (yet simplified) prediction model.\\
\noindent We are now able to evaluate the performance of \dspls{l} on real near-infrared data \DNIR (\ref{sec_dspls_DNIR}) for density prediction. Similarly to  \Ds, RMSE curves in Figure \ref{fig_dspls_l_DNIR} for calibration (top-left) and validation (top-right) globally decrease with an increasing number of components. Errors plateau after six components, indicating that additional latent structure orders might be weakly helpful. The performance gap for \spls{Durif} could occur as it was mainly designed for classification. Again, we assess model interpretation in Figure \ref{fig_dspls_l_DNIR} (bottom) for six latent vectors. By nature,  location of the most influential features of  spectra for a specific property is yet to be unveiled. One may expect that most of the meaningful variables  are located in the active parts of the signal, e.g. spectral bands with relatively higher intensities, with some others possibly in quieter wavenumber ranges. On the top panel, NIR spectra are mainly active\footnote{\label{note_chem_expl}We do not endeavour a chemical explanation here. It ought to be substanciated in forthcoming paper \cite{Duval_L_2023_PREPRINT_ifpen_nisdpp208nirhsdr}} from \SIrange{4000}{4800}{\centi\meter^{-1}} and  \SIrange{5500}{6000}{\centi\meter^{-1}}. Meaningful PLS  coefficients are visible on a much wider support, provoking  ambiguity on the identification of spectral bands related to density. All sPLS actually have smaller support, \spls{LeCao} and \dspls{l} being the sparsest with  \lz respectively equal to \numlist{88;82}. The first singularity of \dspls{l} is the contiguous and smoothness of its coefficients. By contrast,  \spls{Chun} and  \spls{Durif} coefficients location appear to be more scattered across the wavenumber axis, in non-contiguous small chunks and even isolated spikes. The second is the absence of response in the \SIrange{5500}{6000}{\centi\meter^{-1}} bands\footnoteref{note_chem_expl} in \dspls{l}. We are not able to chemically explain the discrepancy of absence/presence results in this band. However, \dspls{l} does not need it to remain almost as accurate as its competitors.

\subsection{Dual-sPLS pseudo-least squares evaluation (\Dsb) \label{sec_dspls_ls_eval}}

The \dspls{LS} requires data to be represented by a non-singular matrix \textbf{X}, as explained in Section \ref{sec_dspls_dspls_ls_r}. Since real data \DNIR is singular, we use simulated data \Dsb presented in Section \ref{sec_dspls_DSIM}. As the number of variables in \Dsb is already small, we only shrink  \SI{60}{\percent} of its variables to evaluate the \dspls{LS} against classical least squares. The latter is denoted by dashes, as the number of latent components is meaningless in this case. \\

\begin{figure}
    \centering
    \begin{tabular}{cc}
    \begin{subfigure}{0.49\textwidth}
        \centering
        \includegraphics[width=\textwidth]{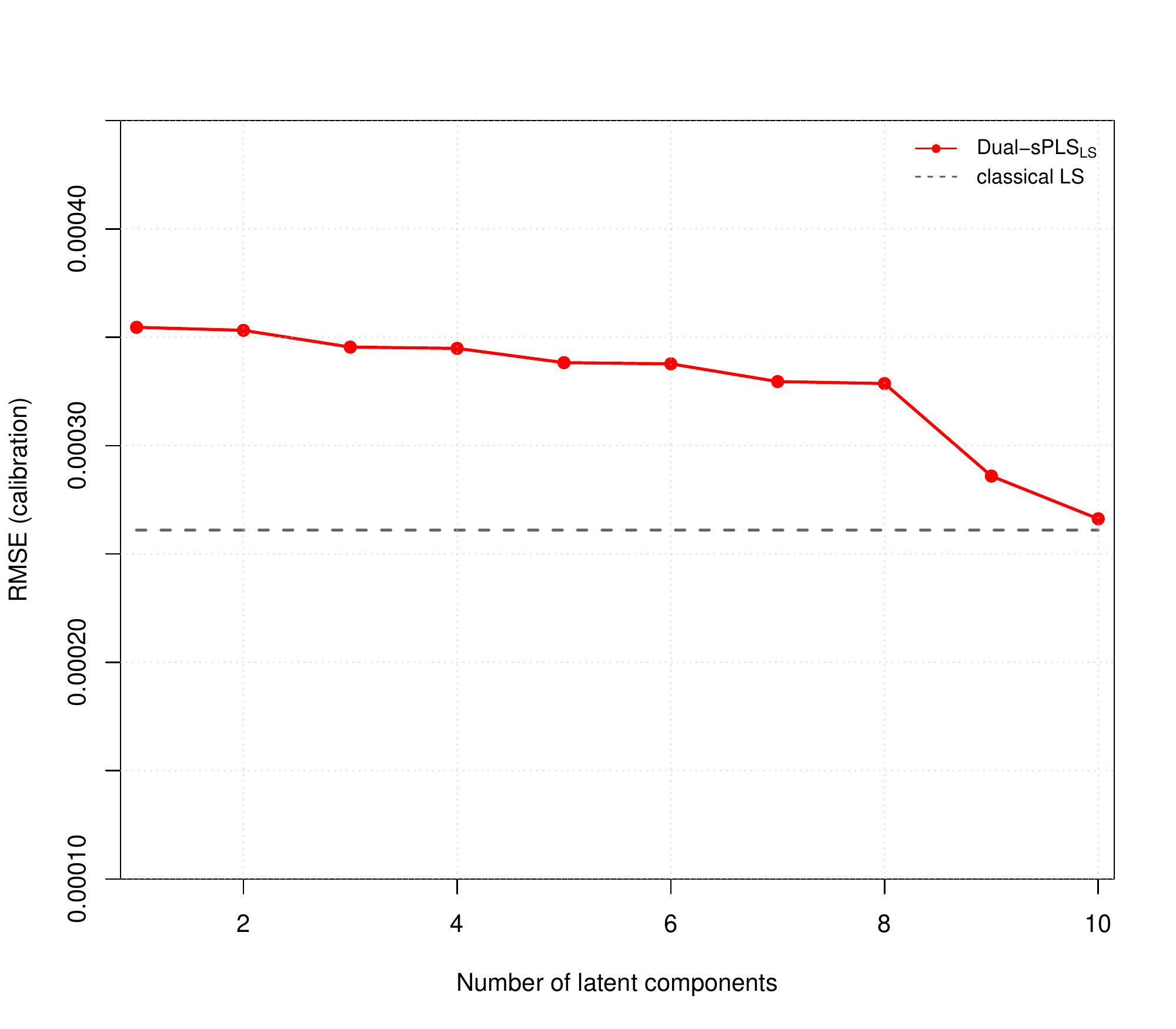}
    \end{subfigure}%
    \hfill
        \begin{subfigure}{0.49\textwidth}
        \centering
       \includegraphics[width=\textwidth]{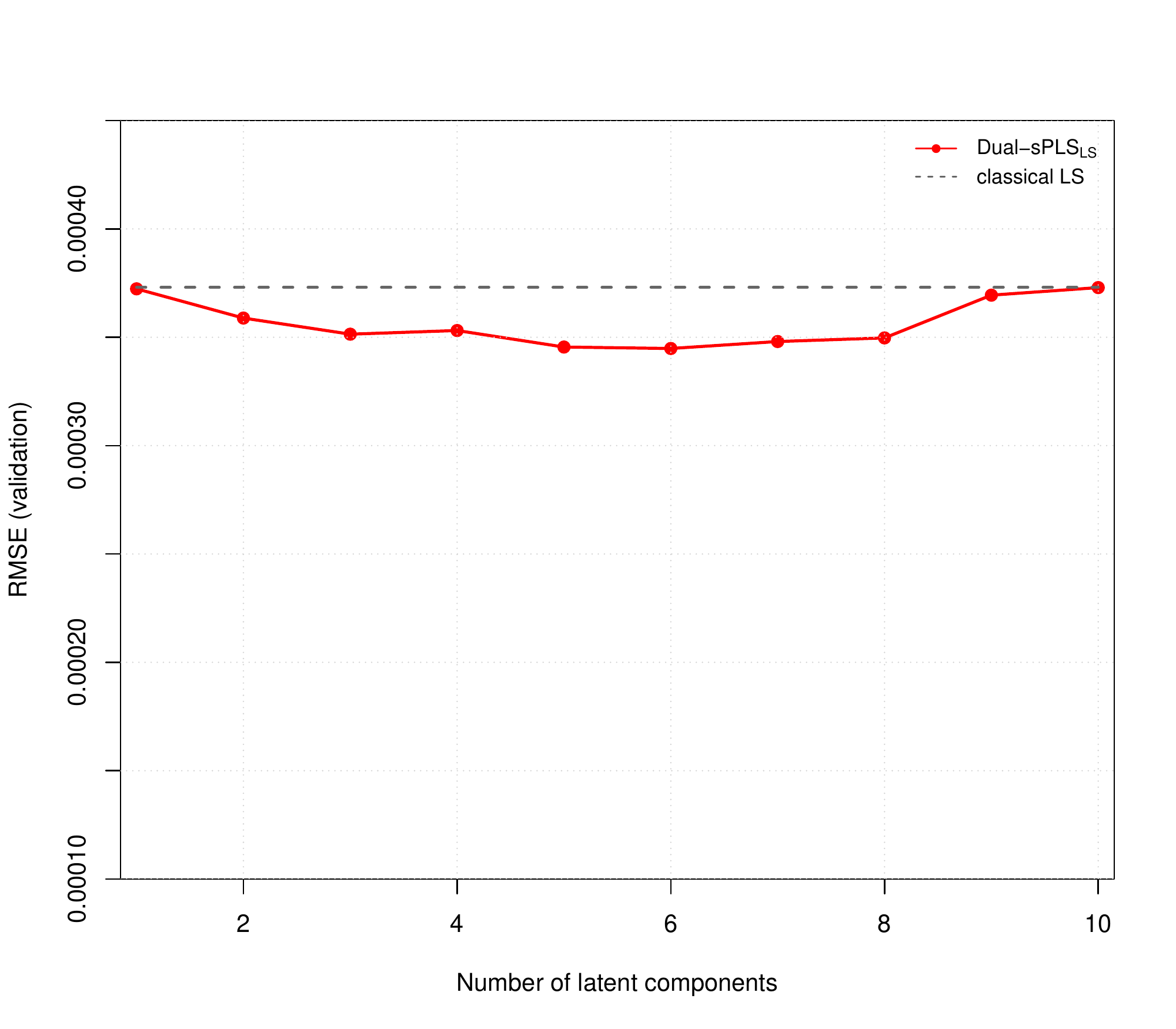}
    \end{subfigure}%
    \\
    \begin{subfigure}{0.95\textwidth}
        \centering
        \includegraphics[width=\textwidth]{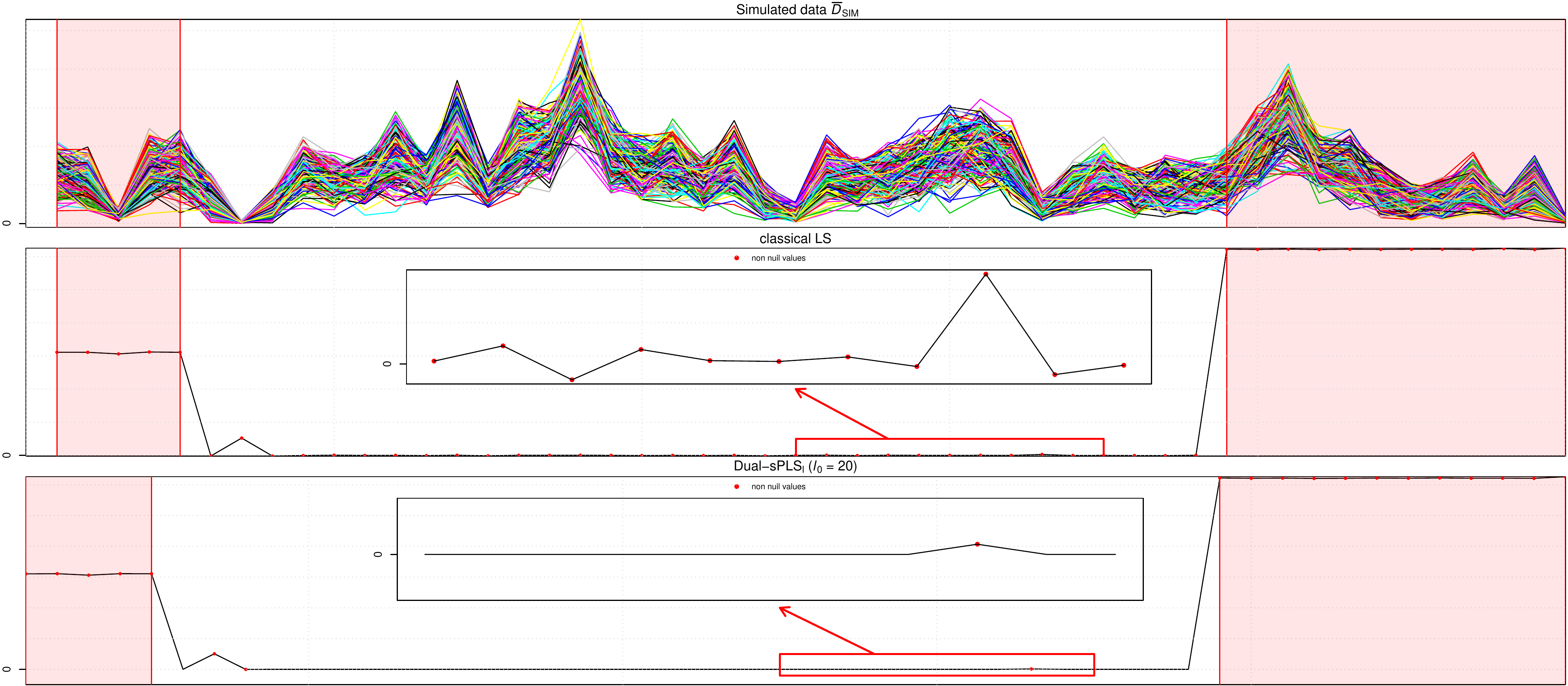}
    \end{subfigure}
    \end{tabular}
         \caption{\dspls{LS} evaluation on simulated data \Dsb. (Top) RMSE values for calibration (left) and validation (right) with respect to the number of latent components. (Bottom) From top to bottom:  simulated data \Dsb, regression coefficients of least squares and \dspls{LS} for five components.} 
    \label{fig_dspls_LS}
\end{figure}

\noindent For calibration (Figure \ref{fig_dspls_LS} top-left) the RMSE for \dspls{LS} decreases mildly as the number of components increases. It approaches the least squares performance. For validation  (Figure \ref{fig_dspls_LS} top-right) \dspls{LS} performs similarly or better than least squares all over model orders. This contrast in performance might be explained by a tendency to overfit for least squares. A better prediction performance is expected with our model. Similarly to the \dspls{l}, we also choose to evaluate it with six components in the bottom of Figure \ref{fig_dspls_LS}. Again, redish regions indicate active variables for the unknown linear model. We observe an overall similarity in the dynamics of both regression coefficients: strong amplitude in the first five and last ten variables corresponding to active regions. The main difference resides in the intermediate part, irrelevant to the response. Least squares as expected shrinks inactive variables towards zero but not as much as \dspls{LS} does. This is exemplified  in the zoomed panels, where \dspls{LS} exhibit much less non-zero coefficients. 

\subsection{Dual-sPLS pseudo-ridge evaluation (\Ds, \DNIR) \label{sec_dspls_r_eval}}
\begin{figure}
	\centering
	\begin{tabular}{cc}
		\begin{subfigure}{0.49\textwidth}
			\centering
			\includegraphics[width=\textwidth]{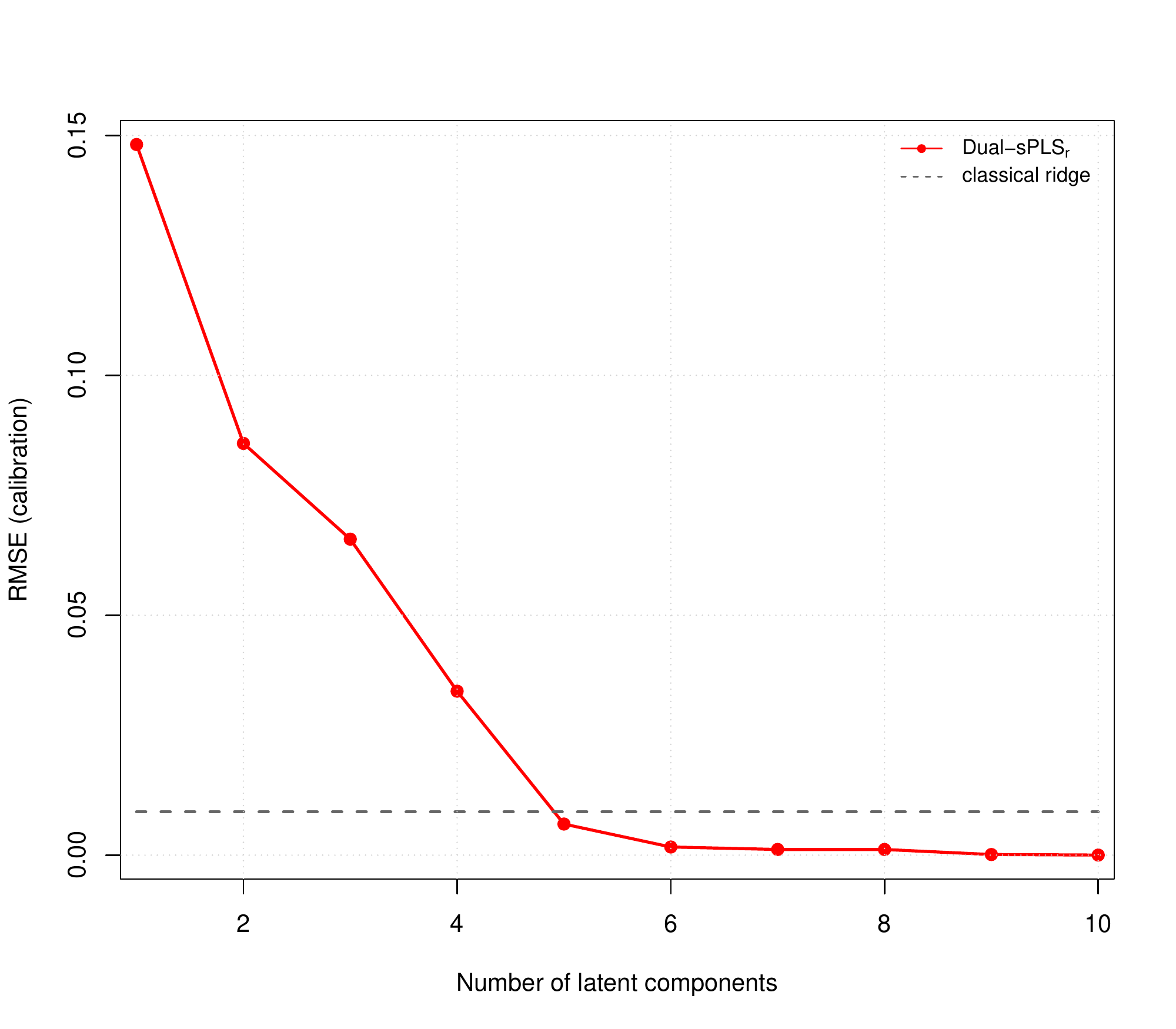}
		\end{subfigure}
		\hfill
		\begin{subfigure}{0.49\textwidth}
			\centering
			\includegraphics[width=\textwidth]{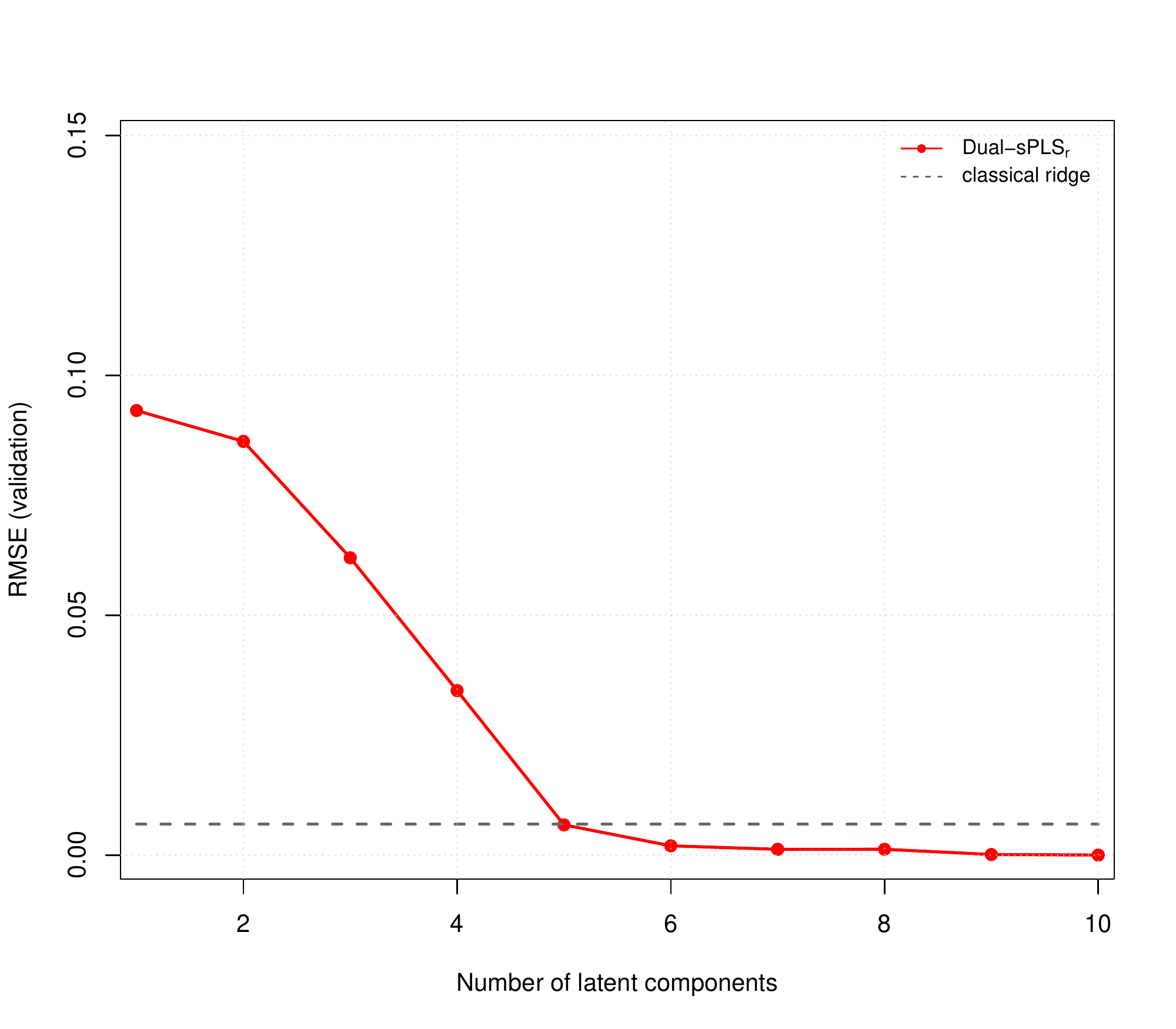}
		\end{subfigure}
		\\
			\begin{subfigure}{0.95\textwidth}
				\centering
				\includegraphics[width=\textwidth]{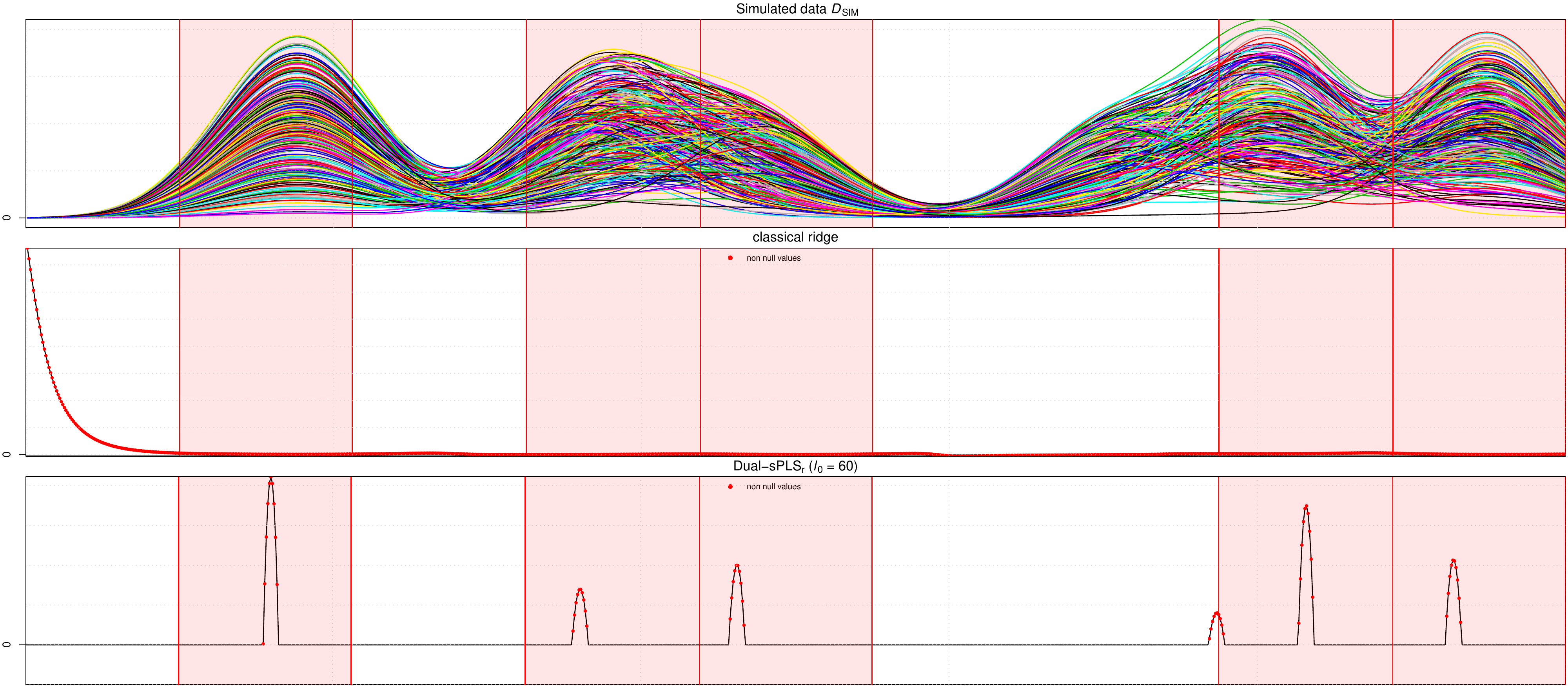}
			\end{subfigure}
	\end{tabular}
	         \caption{\dspls{r} evaluation on simulated data \Ds. (Top) RMSE values for calibration (left) and validation (right) with respect to the number of latent components. (Bottom) From top to bottom:  original data \Ds, regression coefficients of ridge and \dspls{r} for five components.} 
	\label{fig_dspls_r_DSIM}
\end{figure}

\begin{figure}
	\centering
	\begin{tabular}{cc}
		\begin{subfigure}{0.49\textwidth}
			\centering
			\includegraphics[width=\textwidth]{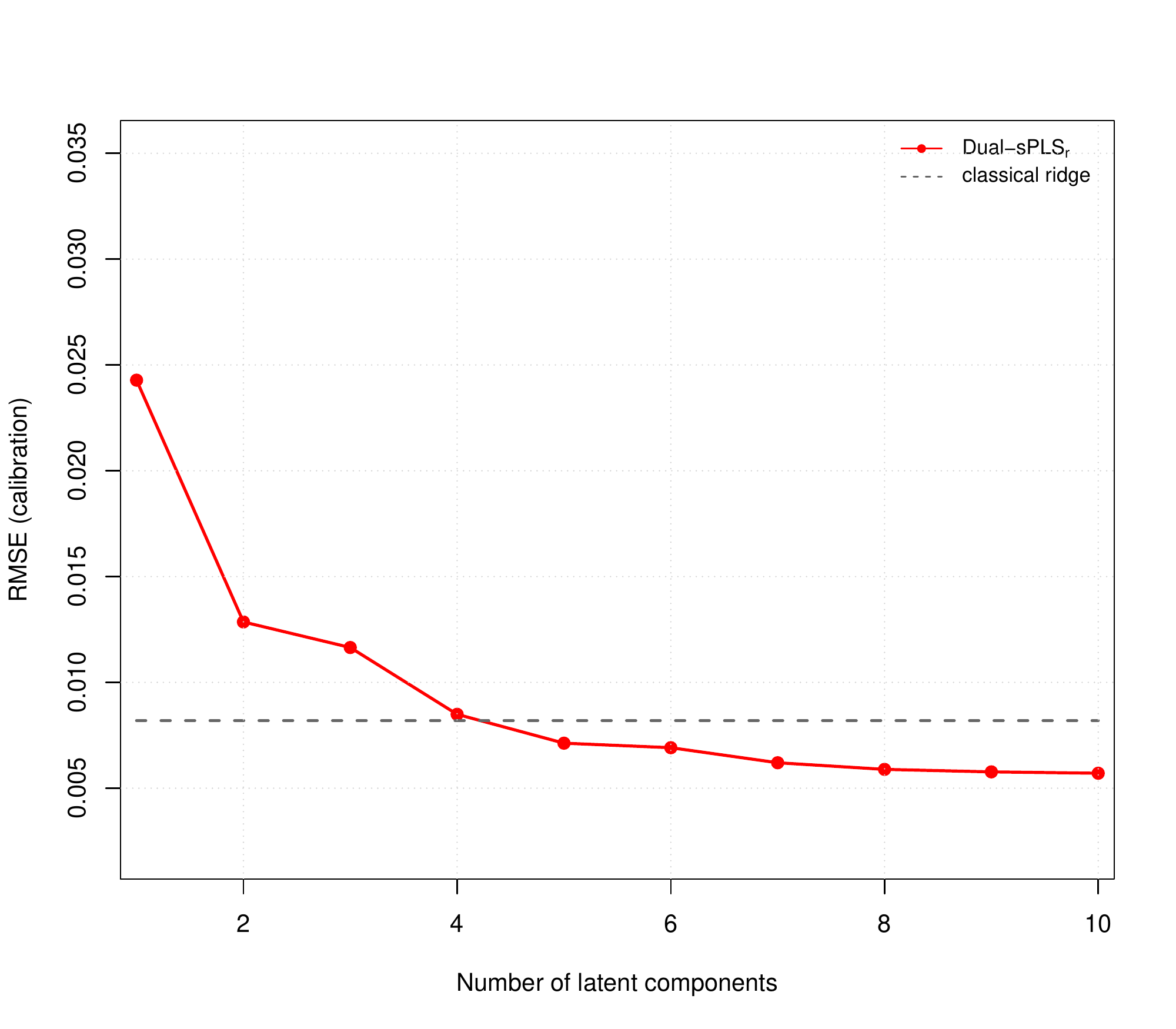}
		\end{subfigure}
		\hfill
		\begin{subfigure}{0.49\textwidth}
			\centering
			\includegraphics[width=\textwidth]{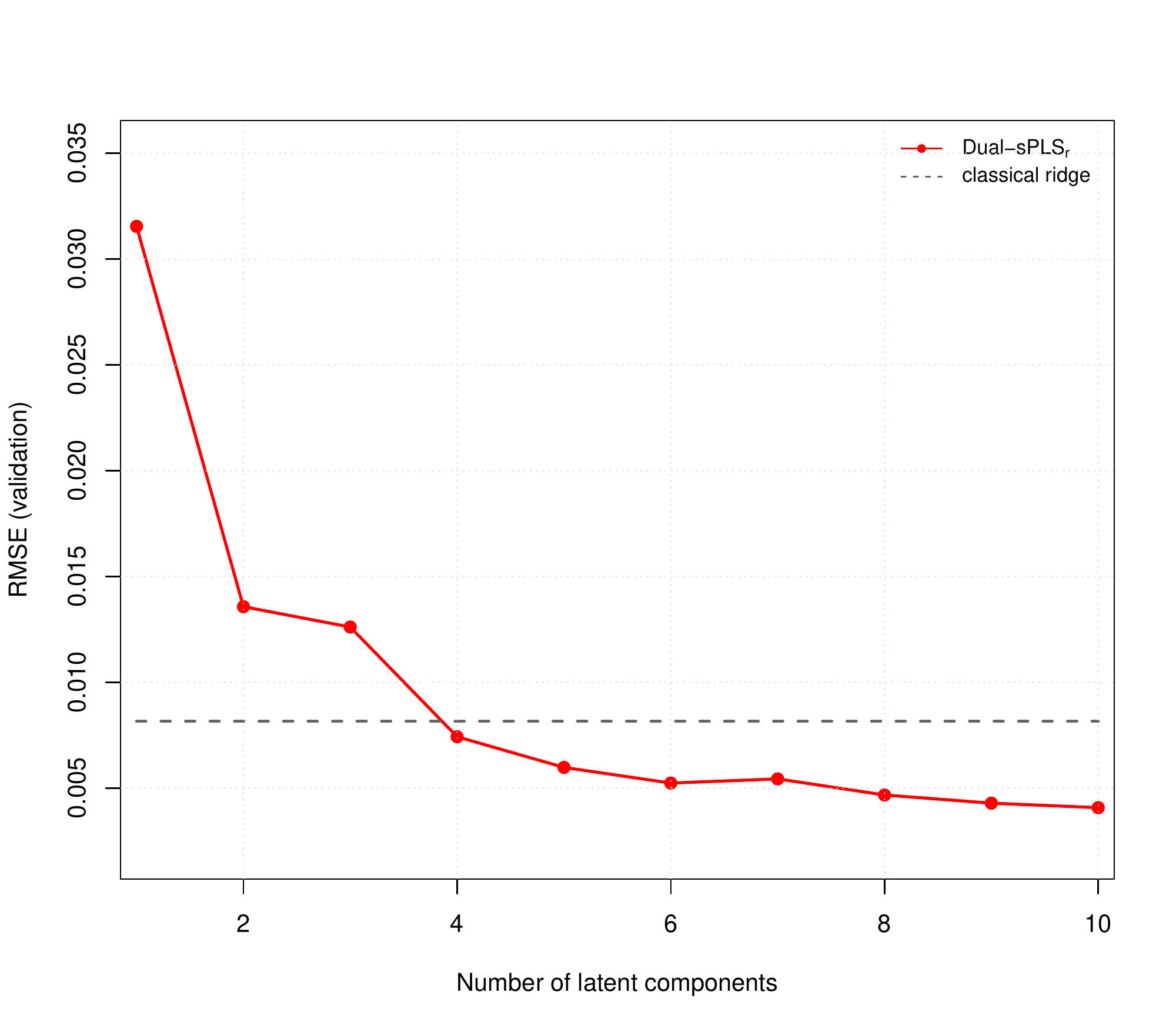}
		\end{subfigure}
		\\
			\begin{subfigure}{0.95\textwidth}
				\centering
				\includegraphics[width=\textwidth]{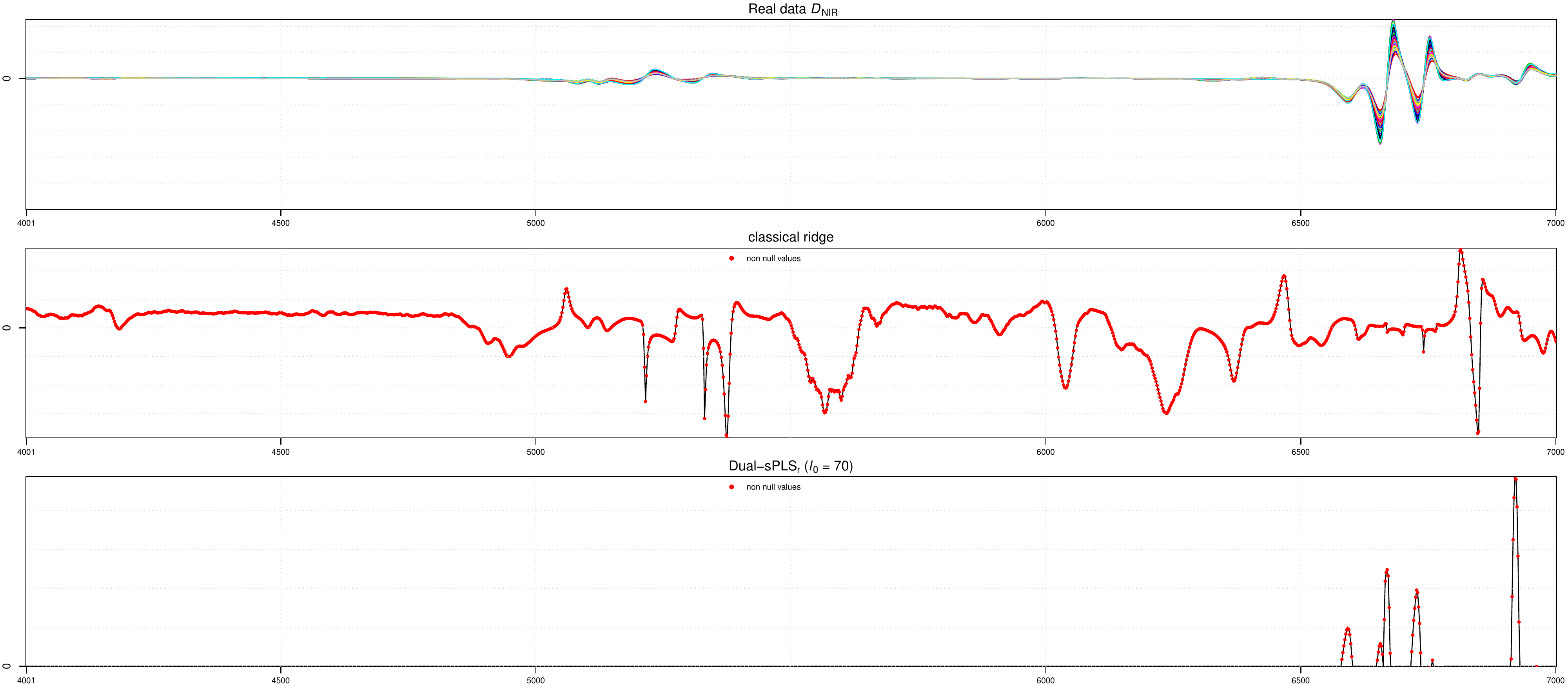}
			\end{subfigure}
		
	\end{tabular}
         \caption{\dspls{r} evaluation on real data \DNIR. (Top) RMSE values for calibration (left) and validation (right) with respect to the number of latent components. (Bottom) From top to bottom: original data \DNIR, regression coefficients of ridge and \dspls{r} for five components.} 
	\label{fig_dspls_r_DNIR}
\end{figure}

\Dspls{r} is compared to classical ridge regression (Section \ref{sec_dspls_lasso}) either applied to simulated data \Ds or real data \DNIR. Ridge hyper parameter $t$ (equation \eqref{eq_dspls_ridge_opt}) is fixed using cross validation. We set $\lambda_2$ for \dspls{r} (equation \eqref{eq_dspls_r_pseudo_norm}) to $\dfrac{1}{t}$ for easier comparison. All other parameters are kept as for \dspls{l} (Section \ref{sec_dspls_l_eval}).
Looking at top-left and -right in Figure \ref{fig_dspls_r_DSIM} \dspls{r} reaches a plateau for \Ds after five latent components. Moreover, its RMSE values are slightly lower than ridge's for both calibration and validation. We can safely select six latent components as before. Reference coefficients for ridge are misleading because the largest ones do not reside in influencing areas. They therefore can not be used for data interpretation. By selecting only fifty variables, located in red regions governing the model, \dspls{r} better succeeds in both prediction and localization. Similar conclusions can be drawn for real data \DNIR on RMSE values. \Dspls{r} even better predicts the response $\vecy$ with only four components. Regression coefficients (Figure \ref{fig_dspls_r_DNIR} bottom) yield comments akin to above. While ridge apparently emphasizes unimportant features, \dspls{r} seems more reliable in identifiying of relevant variables to predict density using chemical data. 

\section{Conclusion and perspectives \label{sec_dspls_conclusion}}

We propose a family of dual sparse Partial Least Squares algorithms that broadens the compass of standard PLS. Along with competitive prediction accuracy with respect to PLS as used in chemometrics, we expect additional benefits in dimension reduction or model interpretability. This is achieved by supplementing the traditional optimization problem with well-chosen dual norms.	\\
We chiefly validate this approach by borrowing three classical regression penalties: lasso, least-squares, ridge. Each proposed Dual-sPLS draws close to the reference in calibration/validation performance with a reduced number of latent components. This is assessed in a benchmark on both  realistic simulated models and real near infrared spectroscopy data, against a standard baseline and sparse contenders. Coefficients are sieved with a  user-defined sparsity target. They are well-located in influential data ranges, suggesting a means for better interpretability of the trained prediction reduced model. Pseudo-lasso and ridge Dual-sPLS avatars exhibit close colocation of selected features in both datasets despite different penalties. This suggests a  robust identification of meaningful information in  signals.	\\
The Dual-sPLS framework is thus a good candidate for a host of applications. We provide it as an open-source package in R \cite{Alsouki_L_2022_misc_package_dualspls_dsplsr}. It can be prolonged to other field-favorite penalties, for instance elastic net. We plan to evaluate the alluded ``pseudo-group lasso'' option, to refine  feature selection on important contiguous areas, or to combine  datasets providing complementary information on the predicted response. To improve prediction robustness or reduce the number of necessary latent components (toward three or four instead of six), we explore additional diversity enhancement preprocessing, such as higher-order derivatives and discrete wavelet transforms. Last, as PLS deserves sounder statistical foundations, we endeavor a study of asymptotic convergence bounds. 


\section{Declaration of competing interest \label{sec_dspls_declaration}}
The authors declare that they have no known competing financial interests or personal relationships that could have appeared to influence the work reported in this paper.

\section{Acknowledgements \label{sec_dspls_acknowledgements}}
This work was performed within the framework of the LABEX MILYON (ANR-10-LABX- 0070) of Universit\'e de Lyon, within the program "Investissements d’Avenir" (ANR-11-IDEX- 0007) operated by the French National Research Agency (ANR). We acknowledge the financial support of the Research Council of the Saint Joseph University of Beirut.
Ghislain Durif contributed in the code validation procedure in R. IFPEN provided the real data set used in the applications. We thank No\'emie Caillol, Luca Castelli and Ir\`ene Gannaz for useful comments.
Our mentor, colleague or friend Fran\c{c}ois Wahl passed away unexpectedly during the writing of this paper. He was the driving force of our team. 


\bibliographystyle{elsarticle-num} 

\begin{thebibliography}{10}
	\expandafter\ifx\csname url\endcsname\relax
	\def\url#1{\texttt{#1}}\fi
	\expandafter\ifx\csname urlprefix\endcsname\relax\def\urlprefix{URL }\fi
	\expandafter\ifx\csname href\endcsname\relax
	\def\href#1#2{#2} \def\path#1{#1}\fi
	
	\bibitem{RamirezVerduzco_L_2012_j-fuel_predicting_cnkvdhhvbfamec}
	L.~F. Ram{\'{\i}}rez-Verduzco, J.~E. Rodr{\'{\i}}guez-Rodr{\'{\i}}guez, A.~del
	Rayo Jaramillo-Jacob, Predicting cetane number, kinematic viscosity, density
	and higher heating value of biodiesel from its fatty acid methyl ester
	composition, Fuel 91~(1) (2012) 102--111.
	\newblock \href {https://doi.org/10.1016/j.fuel.2011.06.070}
	{\path{doi:10.1016/j.fuel.2011.06.070}}.
	
	\bibitem{Willard_H_1965_book_methodes_pac}
	H.~H. Willard, L.~L. Merritt, J.~A. Dean, M{\'e}thodes physiques de l'analyse
	chimique, Dunod, 1965.
	
	\bibitem{Verdier_S_2009_j-fuel_critical_avi}
	S.~Verdier, J.~A.~P. Coutinho, A.~M.~S. Silva, O.~F. Alkilde, J.~A. Hansen, A
	critical approach to viscosity index, Fuel 88~(11) (2009) 2199--2206.
	\newblock \href {https://doi.org/10.1016/j.fuel.2009.05.016}
	{\path{doi:10.1016/j.fuel.2009.05.016}}.
	
	\bibitem{Wold_S_1995_j-chem-int-lab-syst_chemometrics_wdwmwiwdwwfi}
	S.~Wold, Chemometrics; what do we mean with it, and what do we want from it?,
	Chemometr. Intell. Lab. Syst. 30 (1995) 109--115.
	
	\bibitem{Wold_S_2001_j-chem-int-lab-syst_pls-regression_btc}
	S.~Wold, M.~Sj\"ostr\"om, L.~Eriksson, {PLS}-regression: a basic tool of
	chemometrics, Chemometr. Intell. Lab. Syst. 58~(2) (2001) 109--130.
	\newblock \href {https://doi.org/10.1016/s0169-7439(01)00155-1}
	{\path{doi:10.1016/s0169-7439(01)00155-1}}.
	
	\bibitem{Tibshirani_R_1996_j-r-stat-soc-b-stat-methodol_regression_sslasso}
	R.~Tibshirani,
	\href{http://www.ams.org/mathscinet-getitem?mr=1379242}{Regression shrinkage
		and selection via the lasso}, J. R. Stat. Soc. Ser. B Stat. Methodol. 58~(1)
	(1996) 267--288.
	\newline\urlprefix\url{http://www.ams.org/mathscinet-getitem?mr=1379242}
	
	\bibitem{Cherni_A_2020_j-ieee-tsp_spoq_lpolqrssrams}
	A.~Cherni, E.~Chouzenoux, L.~Duval, J.-C. Pesquet, {SPOQ}
	$\ell_p$-over-$\ell_q$ regularization for sparse signal recovery applied to
	mass spectrometry, IEEE Trans. Signal Process. 68 (2020) 6070--6084.
	\newblock \href {https://doi.org/10.1109/TSP.2020.3025731}
	{\path{doi:10.1109/TSP.2020.3025731}}.
	
	\bibitem{MateosAparicioMorales_G_2011_j-commun-stat-theory-methods}
	G.~Mateos-Aparicio~Morales, Partial least squares ({PLS}) methods: Origins,
	evolution, and application to social sciences, Commun. Stat. Theory Methods
	40~(13) (2011) 2305--2317.
	\newblock \href {https://doi.org/10.1080/03610921003778225}
	{\path{doi:10.1080/03610921003778225}}.
	
	\bibitem{Mehmood_T_2016_j-chemometrics_diversity_aplso}
	T.~Mehmood, B.~Ahmed, The diversity in the applications of partial least
	squares: an overview, J. Chemometrics 30~(1) (2016) 4--17.
	\newblock \href {https://doi.org/10.1002/cem.2762}
	{\path{doi:10.1002/cem.2762}}.
	
	\bibitem{Boulesteix_A_2007_j-brief-bioinform_partial_lsvtahdgd}
	A.-L. Boulesteix, K.~Strimmer,
	\href{https://doi.org/10.1093/bib/bbl016}{Partial least squares: a versatile
		tool for the analysis of high-dimensional genomic data}, Brief. Bioinform.
	8~(1) (2007) 32--44.
	\newblock \href
	{http://arxiv.org/abs/https://academic.oup.com/bib/article-pdf/8/1/32/737013/bbl016.pdf}
	{\path{arXiv:https://academic.oup.com/bib/article-pdf/8/1/32/737013/bbl016.pdf}},
	\href {https://doi.org/10.1093/bib/bbl016} {\path{doi:10.1093/bib/bbl016}}.
	\newline\urlprefix\url{https://doi.org/10.1093/bib/bbl016}
	
	\bibitem{Krishnan_A_2011_j-neuroimage_partial_lsplsmntr}
	A.~Krishnan, L.~J. Williams, A.~R. McIntosh, H.~Abdi, Partial least squares
	({PLS}) methods for neuroimaging: A tutorial and review, Neuromimage 56~(2)
	(2011) 455--475.
	\newblock \href {https://doi.org/10.1016/j.neuroimage.2010.07.034}
	{\path{doi:10.1016/j.neuroimage.2010.07.034}}.
	
	\bibitem{Wright_J_2022_book_high-dimensional_daldmpca}
	J.~Wright, Y.~Ma,
	\href{https://www.ebook.de/de/product/41126998/john_wright_yi_ma_high_dimensional_data_analysis_with_low_dimensional_models_principles_computation_and_applications.html}{High-Dimensional
		Data Analysis with Low-Dimensional Models: Principles, Computation, and
		Applications}, Cambridge University Press, 2022.
	\newline\urlprefix\url{https://www.ebook.de/de/product/41126998/john_wright_yi_ma_high_dimensional_data_analysis_with_low_dimensional_models_principles_computation_and_applications.html}
	
	\bibitem{Wold_H_1975_incoll_path_mlvnipalsa}
	H.~Wold, Path models with latent variables: The {NIPALS} approach, in:
	Quantitative Sociology. International Perspectives on Mathematical and
	Statistical Modeling, Elsevier, 1975, pp. 307--357.
	\newblock \href {https://doi.org/10.1016/b978-0-12-103950-9.50017-4}
	{\path{doi:10.1016/b978-0-12-103950-9.50017-4}}.
	
	\bibitem{DeJong_S_1993_j-chemometr-intell-lab-syst_simpls_aaplsr}
	S.~de~Jong, {SIMPLS}: An alternative approach to partial least squares
	regression 18~(3) (1993) 251--263.
	\newblock \href {https://doi.org/10.1016/0169-7439(93)85002-x}
	{\path{doi:10.1016/0169-7439(93)85002-x}}.
	
	\bibitem{Zou_H_2005_j-r-stat-soc-b-stat-methodol_regularization_vsven}
	H.~Zou, T.~Hastie, Regularization and variable selection via the elastic net,
	J. R. Stat. Soc. Ser. B Stat. Methodol. 67~(2) (2005) 301--320.
	\newblock \href {https://doi.org/10.1111/j.1467-9868.2005.00503.x}
	{\path{doi:10.1111/j.1467-9868.2005.00503.x}}.
	
	\bibitem{Hastie_T_2015_book_statistical_llslg}
	T.~Hastie, R.~Tibshirani, M.~Wainwright, Statistical Learning with Sparsity:
	The Lasso and Generalizations, CRC Press, 2015.
	
	\bibitem{Hoerl_A_1970_j-technometrics_ridge_ranp}
	A.~E. Hoerl, R.~W. Kennard,
	\href{http://dx.doi.org/10.1080/00401706.1970.10488635}{Ridge regression:
		Applications to nonorthogonal problems}, Technometrics 12~(1) (1970) 69--82.
	\newblock \href {https://doi.org/10.1080/00401706.1970.10488635}
	{\path{doi:10.1080/00401706.1970.10488635}}.
	\newline\urlprefix\url{http://dx.doi.org/10.1080/00401706.1970.10488635}
	
	\bibitem{LeCao_2008_j-stat-appl-genet-mol-biol_sparse_plsvsiod}
	K.-A. L{\^{e}}~Cao, D.~Rossouw, C.~Robert-Grani{\'{e}}, P.~Besse, A sparse
	{PLS} for variable selection when integrating omics data, Stat. Appl. Genet.
	Mol. Biol. 7~(1) (2008) 35.
	\newblock \href {https://doi.org/10.2202/1544-6115.1390}
	{\path{doi:10.2202/1544-6115.1390}}.
	
	\bibitem{Shen_H_2008_j-multivar-anal_sparse_pcarlrma}
	H.~Shen, J.~Z. Huang, Sparse principal component analysis via regularized low
	rank matrix approximation, J. Multivar. Anal. 99~(6) (2008) 1015--1034.
	\newblock \href {https://doi.org/10.1016/j.jmva.2007.06.007}
	{\path{doi:10.1016/j.jmva.2007.06.007}}.
	
	\bibitem{Tenenhaus_M_1998_book_regression_plstp}
	M.~Tenenhaus, La r{\'{e}}gression {PLS}. Th{\'{e}}orie et pratique,
	{\'{E}}ditions Technip, 1998.
	
	\bibitem{Chun_H_2010_j-r-stat-soc-b-stat-methodol_sparse_plsrsdrvs}
	H.~Chun, S.~Kele\c{s},
	\href{http://dx.doi.org/10.1111/j.1467-9868.2009.00723.x}{Sparse partial
		least squares regression for simultaneous dimension reduction and variable
		selection}, J. R. Stat. Soc. Ser. B Stat. Methodol. 72~(1) (2010) 3--25.
	\newblock \href {https://doi.org/10.1111/j.1467-9868.2009.00723.x}
	{\path{doi:10.1111/j.1467-9868.2009.00723.x}}.
	\newline\urlprefix\url{http://dx.doi.org/10.1111/j.1467-9868.2009.00723.x}
	
	\bibitem{Durif_G_2018_j-bioinformatics_high_dccasplslr}
	G.~Durif, L.~Modolo, J.~Michaelsson, J.~E. Mold, S.~Lambert-Lacroix, F.~Picard,
	High dimensional classification with combined adaptive sparse {PLS} and
	logistic regression, Bioinformatics 34~(3) (2018) 485--493.
	\newblock \href {https://doi.org/10.1093/bioinformatics/btx571}
	{\path{doi:10.1093/bioinformatics/btx571}}.
	
	\bibitem{Bach_F_2012_j-found-trend-mach-learn_optimization_sip}
	F.~Bach, R.~Jenatton, J.~Mairal, G.~Obozinski, Optimization with
	sparsity-inducing penalties, Found. Trends Mach. Learn. 4~(1) (2012) 1--106.
	\newblock \href {https://doi.org/10.1561/2200000015}
	{\path{doi:10.1561/2200000015}}.
	
	\bibitem{RCoreTeam_0_2021_manual_r_lesc}
	{R Core Team}, \href{https://www.R-project.org/}{R: A Language and Environment
		for Statistical Computing}, R Foundation for Statistical Computing, Vienna,
	Austria (2021).
	\newline\urlprefix\url{https://www.R-project.org/}
	
	\bibitem{Alsouki_L_2022_misc_package_dualspls_dsplsr}
	L.~Alsouki, F.~Wahl, G.~Durif,
	\href{https://CRAN.R-project.org/package=dual.spls}{{dual.spls}: Dual sparse
		partial least squares regression}, CRAN, R package version 0.1.2 (Oct. 2022).
	\newline\urlprefix\url{https://CRAN.R-project.org/package=dual.spls}
	
	\bibitem{Simon_N_2013_j-comp-graph-stat_sparse-group_l}
	N.~Simon, J.~Friedman, T.~Hastie, R.~Tibshirani, A sparse-group lasso, J. Comp.
	Graph. Stat.) 22~(2) (2013) 231--245.
	\newblock \href {https://doi.org/10.1080/10618600.2012.681250}
	{\path{doi:10.1080/10618600.2012.681250}}.
	
	\bibitem{Stone_M_1974_j-r-stat-soc-b-stat-methodol_cross-validation_casp}
	M.~Stone, Cross-validatory choice and assessment of statistical predictions, J.
	R. Stat. Soc. Ser. B Stat. Methodol. 36~(2) (1974) 111--133.
	\newblock \href {https://doi.org/10.1111/j.2517-6161.1974.tb00994.x}
	{\path{doi:10.1111/j.2517-6161.1974.tb00994.x}}.
	
	\bibitem{Chalmers_J_2002_book_handbook_vs}
	J.~M. Chalmers, P.~R. Griffiths (Eds.), Handbook of Vibrational Spectroscopy,
	Wiley, 2002.
	
	\bibitem{Laxalde_J_2011_j-anal-chim-acta_characterization_hounisoppmvs}
	J.~Laxalde, C.~Ruckebusch, O.~Devos, N.~Caillol, F.~Wahl, L.~Duponchel,
	Characterisation of heavy oils using near-infrared spectroscopy: Optimisation
	of pre-processing methods and variable selection, Anal. Chim. Acta 705~(1-2)
	(2011) 227--234.
	\newblock \href {https://doi.org/10.1016/j.aca.2011.05.048}
	{\path{doi:10.1016/j.aca.2011.05.048}}.
	
	\bibitem{Laxalde_J_2012_phd_analyse_plpsi}
	J.~Laxalde, Analyse des produits lourds du p{\'{e}}trole par spectroscopie
	infrarouge, Ph.D. thesis, Universit{\'{e}} de Lille 1 (2012).
	
	\bibitem{Duval_L_2023_PREPRINT_ifpen_nisdpp208nirhsdr}
	L.~Duval, L.~Alsouki, F.~Wahl, J.~Laxalde, N.~Caillol, {IFPEN} near-infrared
	spectroscopy dataset for property prediction: 208 {NIR} hydrocarbon spectra
	and density response, PREPRINT (2023).
	
	\bibitem{Ning_X_2014_j-chemometr-intell-lab-syst_chromatogram_bedusbeads}
	X.~Ning, I.~W. Selesnick, L.~Duval, Chromatogram baseline estimation and
	denoising using sparsity ({BEADS}) 139 (2014) 156--167.
	\newblock \href {https://doi.org/10.1016/j.chemolab.2014.09.014}
	{\path{doi:10.1016/j.chemolab.2014.09.014}}.
	
	\bibitem{Savitzky_A_1964_j-anal-chem_smoothing_ddslsp}
	A.~Savitzky, M.~J.~E. Golay, Smoothing and differentiation of data by
	simplified least squares procedures, Anal. Chem. 36~(8) (1964) 1627---1639.
	
	\bibitem{DeNoyer_L_2002_incoll_smoothing_ds}
	L.~K. DeNoyer, J.~G. Dodd, Smoothing and derivatives in spectroscopy.
	\newblock \href {https://doi.org/10.1002/0470027320.s4501}
	{\path{doi:10.1002/0470027320.s4501}}.
	
	\bibitem{Sarndal_C_2003_book_model_ass}
	C.-E. S{\"{a}}rndal, B.~Swensson, J.~Wretman, Model Assisted Survey Sampling,
	Springer, 2003.
	
	\bibitem{Kennard_R_1969_j-technometrics_computer_ade}
	R.~W. Kennard, L.~A. Stone, Computer aided design of experiments, Technometrics
	11~(1) (1969) 137--148.
	\newblock \href {https://doi.org/10.1080/00401706.1969.10490666}
	{\path{doi:10.1080/00401706.1969.10490666}}.
	
	\bibitem{Tian_H_2018_j-infrared-phys-technol_weighted_spxymcsscabnis}
	H.~Tian, L.~Zhang, M.~Li, Y.~Wang, D.~Sheng, J.~Liu, C.~Wang, Weighted {SPXY}
	method for calibration set selection for composition analysis based on
	near-infrared spectroscopy, Infrared Phys. Technol. 95 (2018) 88--92.
	\newblock \href {https://doi.org/10.1016/j.infrared.2018.10.030}
	{\path{doi:10.1016/j.infrared.2018.10.030}}.
	
	\bibitem{Alsouki_L_2022_PREPRINT_well-balanced_scvsubpxry}
	L.~Alsouki, L.~Duval, R.~El~Haddad, C.~Marteau, F.~Wahl, {CalValXy}:
	well-balanced and stratified calibration/validation splitting using both
	predictors ${X}$ and response $y$, PREPRINT (2022).
	
\end{thebibliography}


\appendix \label{sec_dspls_supp}
\section{Detailed resolution of Dual-sPLSs \label{sec_dspls_details}}
\subsection{Dual-sPLS pseudo-group lasso\label{sec_dspls_gl_resolution}}
\noindent We recall Equation \eqref{eq_dspls_gl_pseudo_norm}: the \dspls{gl} norm case applied to optimization Problem~\eqref{eq_dspls_dual_norm}.
Note that here
\begin{itemize}
\item $g$ represents a group of $P(g)$  index extracted from $\{1, \dots, P\}$;
\item $G$ represents the number of groups;
\item $\textbf{w}_g$ represents the values of index $g$ in the loading vector $\textbf{w}$.
\end{itemize}
\noindent We denote $\textbf{z}_g$ the variables of $\textbf{z}$ belonging to group $g$. We impose $\textbf{z}_g$ and $\textbf{w}_g$ to be in the same orthant. Let $\boldsymbol\delta_g$ be their vector of signs. By differentiating equation \eqref{eq_dspls_gl_pseudo_norm} we obtain 
\begin{equation} \label{eq_dspls_gl_diff_norm}
\dfrac{\partial \Omega(\textbf{w})}{\partial w_{g}}= \dfrac{\alpha_g w_{g}}{\|\textbf{w}_g\|_2} +\alpha_g\lambda_g\delta_{g}\,.
\end{equation}
Using Lagrange multipliers as in Section \ref{sec_dspls_motivation}, we compare \eqref{eq_dspls_norm_diff} to \eqref{eq_dspls_gl_diff_norm} and obtain for $g \in \{1, \dots, G\}$:

\begin{equation}\label{eq_dspls_gl_sol}
\dfrac{\textbf{w}_g}{\|\textbf{w}_g\|_2} = \dfrac{\textbf{z}_g}{\alpha_g \mu} - \lambda_g \delta_g\,,
\end{equation}
which is simplified by 
\begin{equation}\label{eq_dspls_gl_sol_bis}
\dfrac{\textbf{w}_g}{\|\textbf{w}_g\|_2} = \dfrac{1}{\mu \alpha_g} \textbf{z}_{\nu_g}\,,
\end{equation}
where 
\begin{equation}\label{eq_dspls_gl_znu}
 \textbf{z}_{\nu_g} = \boldsymbol \delta_g (| \textbf{z}_g | - \nu_g)_+ \quad \text{for }  g  \in \{ 1, \dots, G\} \,.
\end{equation}
Here $\nu_g=\mu \alpha_g \lambda_g$ and controls the amount of variables that we would like to shrink to zero. By applying $\ell_2$-norm to \eqref{eq_dspls_gl_sol_bis}, we conclude that $\text{for }  g  \in \{ 1, \dots, G\},$

\begin{equation}\label{eq_dspls_gl_mu_nu}
\mu=\sum_{g=1}^{G} \|\textbf{z}_{\nu_g}\|_2 \qquad \text{and}  \qquad \alpha_g=\dfrac{\|\textbf{z}_{\nu_g}\|_2}{\mu}\,.
\end{equation} 
The term \noindent $\|\textbf{w}_g\|_2$ is more involved. Thus, we simply use grid search. For each group $g$, ten possible values are chosen to be tested. The selection is done by detecting the maximum value of $\|\textbf{w}_g\|_2$ for each group $g$, denoted $\|\textbf{w}_g\|_2^{max}$. The latter is computed by zeroing $\|\textbf{w}_{g'}\|_2$ for all groups $g' \neq g$ and is expressed as: 
\begin{equation}\label{eq_dspls_gl_norm2_w}
\|\textbf{w}_g\|_2^{max}=\dfrac{\mu}{\Omega_g(\textbf{z}_{\nu_g})}\,.
\end{equation}
Then, ten values of each group $g$ are selected inside the interval $[0,\|\textbf{w}_g\|_2^{max}]$. The grid search tests all the possible combinations and retains the one that allows the smallest error. We summarize the methodology with Algorithm \ref{algo_dspls_gl}.

\begin{algorithm}[H] 
\caption{ \Dspls{gl} algorithm } 
\begin{algorithmic} \label{algo_dspls_gl}
\STATE Input: $\textbf{X}^1, \dots,\textbf{X}^G,\textbf{ y}$, $M$ (number of components desired), $\varsigma$ (shrinking ratio), $\alpha_1, \dots, \alpha_g$.
\FOR{ $ m= 1, \dots, M$ }
\STATE $\textbf{X}_m=(\textbf{X}^1, \dots,\textbf{X}^G)$ (combining data)
\STATE $\textbf{z}_m=\textbf{X}_m^T\textbf{y}$ (weight vector)
\STATE Find $\nu$ adaptively according to $\varsigma$ for each group seperatly
\STATE$\textbf{z}_{\nu_g} = \boldsymbol \delta_g (| \textbf{z}_g | - \nu_g)_+ \quad \text{for }  g  \in \{ 1, \dots, G\}$ (applying the threshold) 
\STATE$\mu=\sum_{g=1}^{G} \|  \textbf{z}_{\nu_g} \|_2 $ 
\STATE $\alpha_g=\dfrac{\|  \textbf{z}_{\nu_g} \|_2}{\mu}$   and $\lambda_g=\dfrac{\nu_g}{\alpha_g \mu }\quad \text{for }  g  \in \{ 1, \dots, G\}$
\STATE $\|\textbf{w}_g\|_2^{max}=\dfrac{\mu}{\Omega_g(\textbf{z}_{\nu_g})} \quad \text{for }  g  \in \{ 1, \dots, G\}$
\STATE selection of the values of $\|\textbf{w}_g\|_2$ for each group
\STATE $\textbf{w}_g = \dfrac{\|\textbf{w}_g\|_2}{\mu \alpha_g} \textbf{z}_{\nu_g} \quad \text{for }  g  \in \{ 1, \dots, G\}$ (loadings)
\STATE $\textbf{w}_g=\bigg( \textbf{w}_g \bigg)_{g=1}^G$
\STATE $\textbf{t}_m=\textbf{X}_m\textbf{w}_m$ (component)
\STATE $\textbf{X}_{m+1}=\textbf{X}_m-\mathcal{P}_{\textbf{t}_m}\textbf{X}_m$ (deflation)
\ENDFOR
\STATE Compute $\hat{\boldsymbol \beta}$.
\end{algorithmic}
\end{algorithm}

\subsection{Dual-sPLS pseudo-least squares\label{sec_dspls_LS_resolution}}
\noindent We recall Equation \eqref{eq_dspls_LS_pseudo_norm}: the \dspls{LS} pseudo case applied to optimization Problem~\eqref{eq_dspls_dual_norm}.\\
We impose $\textbf{N}_1\textbf{z}$ and $\textbf{N}_1\textbf{w}$ to be in the same orthant. Let $\boldsymbol \delta_2$ be their vector of signs. By differentiating \eqref{eq_dspls_LS_pseudo_norm} we obtain 
\begin{equation}\label{eq_dspls_LS_diff_norm}
\nabla \Omega (\textbf{w})= \lambda \textbf{N}_1^T \boldsymbol \delta_2 + \dfrac{\textbf{X}^T \textbf{X}\textbf{w}}{\| \textbf{X} \textbf{w}\|_2}\,.
\end{equation}
Using Lagrange multipliers as in Section \ref{sec_dspls_motivation}, we compare \eqref{eq_dspls_norm_diff} to \eqref{eq_dspls_LS_diff_norm} and obtain 
\begin{equation}\label{eq_dspls_LS_sol_2}
\dfrac{\textbf{w}}{\| \textbf{X} \textbf{w}\|_2}=(\textbf{X}^T \textbf{X})^{-1} \dfrac{\textbf{z}}{\boldsymbol \mu}-\lambda (\textbf{X}^T \textbf{X})^{-1} \textbf{N}_1^T \boldsymbol \delta_2\,,
\end{equation}
 imposing the invertibility of $\textbf{X}^T \textbf{X}$. We choose $\textbf{N}_1$ such as 
\begin{equation}\label{eq_dspls_LS_N1_condition_2}
(\textbf{X}^T \textbf{X})^{-1} \textbf{N}_1^T \boldsymbol \delta_2= \text{sign} \bigg( (\textbf{X}^T \textbf{X})^{-1} \textbf{z} \bigg)\,.
\end{equation}
The resolution steps are be similar to the ones from \dspls{l} but instead of applying the threshold on $\textbf{z}$, we apply it on $(\textbf{X}^T \textbf{X})^{-1} \textbf{z}$ which is exactly the classical Least Squares regression coefficients $\hat{\boldsymbol \beta}^{LS}$. So, the simplified solution is 
\begin{equation}\label{eq_dspls_LS_sol_bis_2}
\dfrac{\textbf{w}}{\| \textbf{X} \textbf{w}\|_2}= \dfrac{1}{\mu} \text{sign}(\hat{\boldsymbol \beta}_{LS_j}) ( | \hat{\boldsymbol \beta}_{LS_j} | - \nu )_+\,,
\end{equation}
\noindent where $\nu$ is chosen adaptively.\\
For a simpler algorithm, $\| \textbf{X} \textbf{w}\|_2$ is not computed as it is not mandatory in this case. Additionally, $\textbf{w}$ only depends on $\nu$ and $\hat{\beta}_{LS}$, which means $\textbf{N}_1$ does not intervene in the computation of the optimal solution. Thus, proving that $\textbf{N}_1$ exists is enough. \eqref{eq_dspls_LS_N1_condition_2} implies the following
\begin{equation}\label{eq_dspls_LS_N1_resol}
 \textbf{N}_1^T \boldsymbol \delta_2=(\textbf{X}^T \textbf{X}) \text{sign} \bigg( (\textbf{X}^T \textbf{X})^{-1} \textbf{z} \bigg)\,.
\end{equation}
Let $\textbf{w}$ be an eignvector of $\textbf{N}_1$, and $\textbf{N}'_1$ be such as 
\begin{equation}\label{eq_dspls_LS_N1_resol_bis}
\textbf{N}'_1=\textbf{N}_1-\textbf{w}^T \textbf{w}  \quad \text{ and } \textbf{N}'_1\textbf{w}=0\,.
\end{equation}

\noindent Therefore, using \eqref{eq_dspls_LS_N1_resol} we have
\begin{equation}\label{eq_dspls_ls_N1_resol_final}
\textbf{N}'_1 \boldsymbol \delta_2 = (\textbf{X}^T \textbf{X}) \text{sign}\bigg(  (\textbf{X}^T \textbf{X})^{-1} \textbf{z}  \bigg) - \textbf{ww}^T \boldsymbol \delta_2 \quad \text{ with } \textbf{N}'_1\textbf{w}=0\,.
\end{equation}
\noindent With $\textbf{N}_1$  a square matrix of $P$ variables, \eqref{eq_dspls_ls_N1_resol_final} is a system of $P^2$ unknowns, $P$ equations and $P$ contraints. It can be verified by an infinite number of solutions.

\noindent The following algorithm reformulates the previous steps:
\begin{algorithm}[H]\label{algo_dspls_LS}
\caption{ \textsc{\dspls{LS} algorithm} }
\begin{algorithmic} 
\STATE Input: $\textbf{X},\textbf{y},M \text{ (number of components desired)},\varsigma \text{ (shrinking ratio)}$
\STATE $\textbf{X}_1=\textbf{X}$
\FOR{ $ m= 1, \dots, M$ }
\STATE $\textbf{z}_m=\textbf{X}_m^T\textbf{y}$ (weight vector)
\STATE $\hat{\beta}_{LS}=(\textbf{X}^T \textbf{X})^{-1} \textbf{z}$
\STATE Find $\nu$ adaptively according to $\varsigma$ and $\hat{\beta}_{LS}$
\STATE$\textbf{z}_\nu= (\text{sign}(\hat{\boldsymbol \beta}_{LS}) ( | \hat{\boldsymbol \beta}_{LS} | - \nu )_+)$ (applying the threshold) 
\STATE $\textbf{w}_m=\dfrac{\textbf{z}_\nu}{\mu}$ (loadings)
\STATE $\textbf{w}_m=\dfrac{\textbf{w}_m}{\| \textbf{w} \|_2}$ (normalizing loadings)
\STATE $\textbf{t}_m=\textbf{X}_m\textbf{w}_m$ (component)
\STATE $\textbf{X}_{m+1}=\textbf{X}_m-\mathcal{P}_{\textbf{t}_m}\textbf{X}_m$ (deflation)
\ENDFOR
\STATE Compute $\hat{\boldsymbol \beta}$.
\end{algorithmic}
\end{algorithm}

\subsection{Dual-sPLS pseudo-ridge \label{sec_dspls_r_resolution}}
We recall Equation \eqref{eq_dspls_r_pseudo_norm}: the \dspls{r} pseudo case applied to optimization Problem \eqref{eq_dspls_dual_norm}.
We impose $\textbf{z}$ and $\textbf{w}$ to be in the same orthant. Let $\boldsymbol \delta$ be their vector of signs. By differentiating  \eqref{eq_dspls_r_pseudo_norm}, we obtain 
\begin{equation}\label{eq_dspls_r_diff_norm}
\nabla \Omega (\textbf{w})=\lambda_1 \delta +\lambda_2 \dfrac{\matX^T\matX \textbf{w}}{\|\textbf{X}\textbf{w}\|_2}+\dfrac{\textbf{w}}{\|\textbf{w}\|_2}\,.
\end{equation}
Using Lagrange multipliers as in Section \ref{sec_dspls_motivation}, we compare \eqref{eq_dspls_norm_diff} to \eqref{eq_dspls_r_diff_norm} and obtain 
\begin{equation}\label{eq_dspls_r_sol_bis}
\dfrac{\textbf{w}}{\|\textbf{w}\|_2}=\bigg( \nu_2 \matX ^T\matX + I_P \bigg)^{-1} ( \textbf{z} - \nu_1 \boldsymbol \delta)\,,
\end{equation}
\noindent where $\nu_1= \lambda_1 \mu$ and $\nu_2= \lambda_2 \dfrac{\|\textbf{w}\|_2}{\|\matX \textbf{w}\|_2}.$\\ 
In line with \dspls{l}, we note $\textbf{z}_{\matX,\nu_2}=\bigg( \nu_2 \matX^T\matX + I_P \bigg)^{-1}  \textbf{z}$ and  $\boldsymbol \delta_{\matX}$ its vector of signs.
We exhibit a solution imposing that $\textbf{w}$ and $\textbf{z}_{\matX,\nu_2}$ are in the same orthant, which leads to the following reformulation of~\eqref{eq_dspls_r_sol_bis}:  
\begin{equation}\label{eq_dspls_r_sol_2}
\dfrac{\textbf{w}}{\|\textbf{w}\|_2}=\dfrac{1}{\mu} \delta_{\textbf{X}} ( | \textbf{z}_{\matX,\nu_2}| - \nu_1 )_+\,.
\end{equation}
The threshold $\nu_1$ is chosen with the adaptive procedure described in Section \ref{sec_dspls_dspls_l} and Figure \ref{fig2.2.1}. However, in this case, we compare $\nu_1$ to $| \textbf{z}_{\matX,\nu_2} |$. Since the latter is colinear to $\textbf{z}$, the shrinkage is adequate.  Denoting $\textbf{z}_\nu= \delta_{\textbf{X}} ( | \textbf{z}_{\matX,\nu_2}| - \nu_1 )_+$, simple computations lead to 
\begin{equation}\label{eq_dspls_r_mu}
\mu=\| \textbf{z}_\nu \|_2\,,
\end{equation}
and 
\begin{equation}\label{eq_dspls_r_w}
 \textbf{w}=\dfrac{\mu}{\nu_1 \| \textbf{z}_{\nu} \|_1 + \nu_2 \| \textbf{X} \textbf{z}_{\nu} \|_2^2 + \mu^2}\,.
\end{equation}
\noindent It is summarized in Algorithm \ref{algo_dspls_r}:
\begin{algorithm}[H]
\caption{\textsc{\dspls{r} algorithm}} 
\begin{algorithmic} 
\STATE Input: $\textbf{X}, \textbf{y}, M \text{ (number of components desired)}, \varsigma \text{ (shrinking ratio)}, \nu_2$
\STATE $\textbf{X}_1=\textbf{X}$
\FOR{ $ m= 1, \dots, M$ }
\STATE $\textbf{z}_m=\textbf{X}_m^T\textbf{y}$ (weight vector)
\STATE $\textbf{z}_{\matX,\nu_2}=\bigg( \nu_2 \matX^T\matX + I_P \bigg)^{-1}  \textbf{z}$
\STATE Find $\nu$ adaptively according to $\varsigma$ and $| \textbf{z}_{\matX,\nu_2} |$
\STATE $\boldsymbol \delta_{\matX}$ vector of signs of $\textbf{z}_{\matX,\nu_2}$
\STATE$\textbf{z}_\nu= \delta_{\textbf{X}} ( | \textbf{z}_{\matX,\nu_2}| - \nu_1 )_+$ (applying the threshold) 
\STATE$\mu=\|\textbf{z}_\nu\|_2 $ and $\lambda=\dfrac{\nu}{\mu }$
\STATE $\textbf{w}_m=\dfrac{\mu}{\nu_1 \| \textbf{z}_{\nu} \|_1 + \nu_2 \| \textbf{X} \textbf{z}_{\nu} \|_2^2 + \mu^2}$ (loadings)
\STATE $\textbf{t}_m=\textbf{X}_m\textbf{w}_m$ (component)
\STATE $\textbf{X}_{m+1}=\textbf{X}_m-\mathcal{P}_{\textbf{t}_m}\textbf{X}_m$ (deflation)
\ENDFOR
\STATE Compute $\hat{\boldsymbol \beta}$.
\end{algorithmic}\label{algo_dspls_r}
\end{algorithm}

\section{Complementary plots \label{sec_dspls_additi_plot}}
As mentioned in Section \ref{sec_dspls_results}, metrics MAE and R$^2$ were also computed. They support our findings based on RMSE, as they yield similar results.
\begin{figure}
    \centering
    \begin{tabular}{cc}
    \begin{subfigure}{0.49\textwidth}
        \centering
        \includegraphics[width=\textwidth]{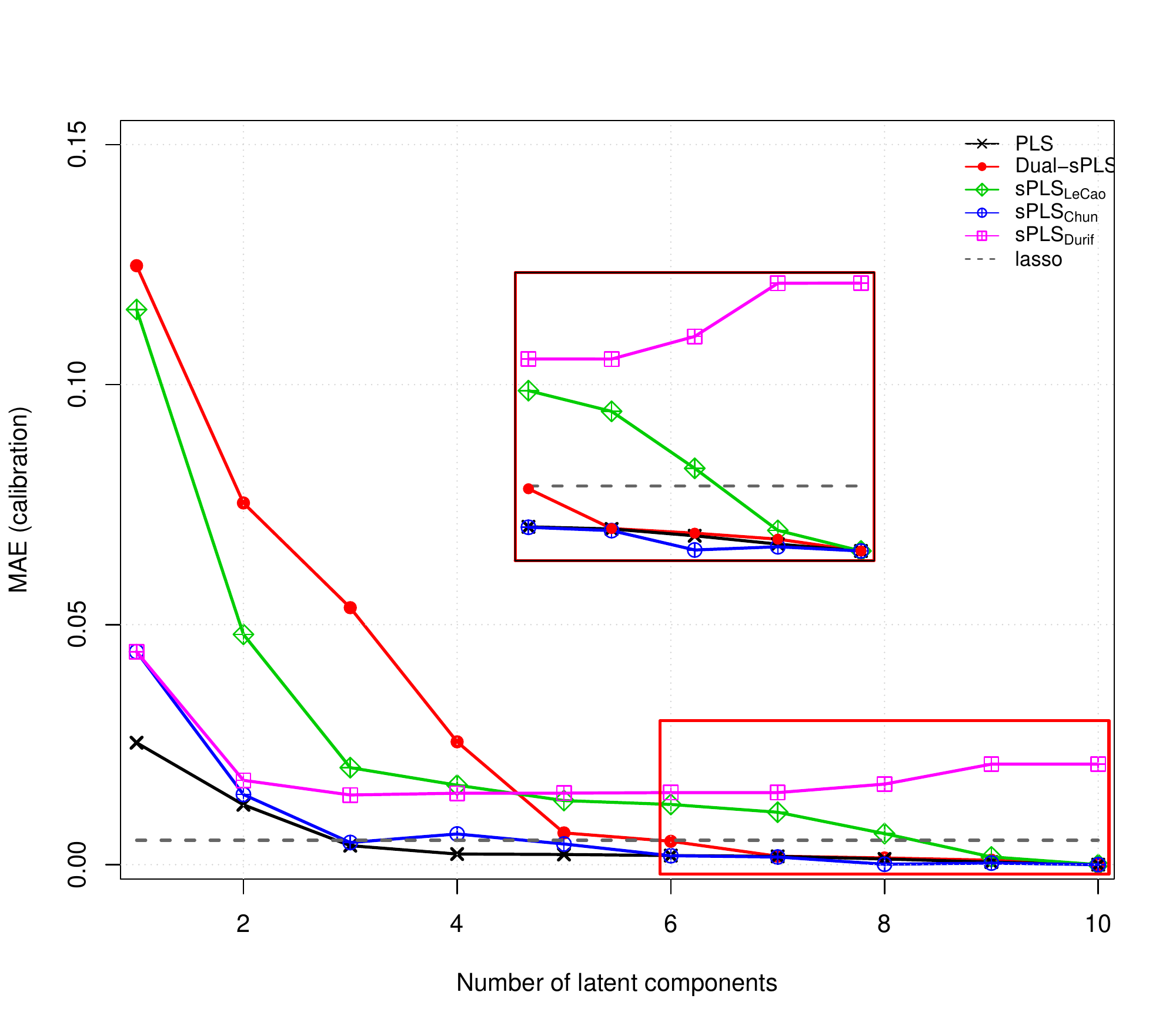}
    \end{subfigure}
    \hfill
        \begin{subfigure}{0.49\textwidth}
        \centering
       \includegraphics[width=\textwidth]{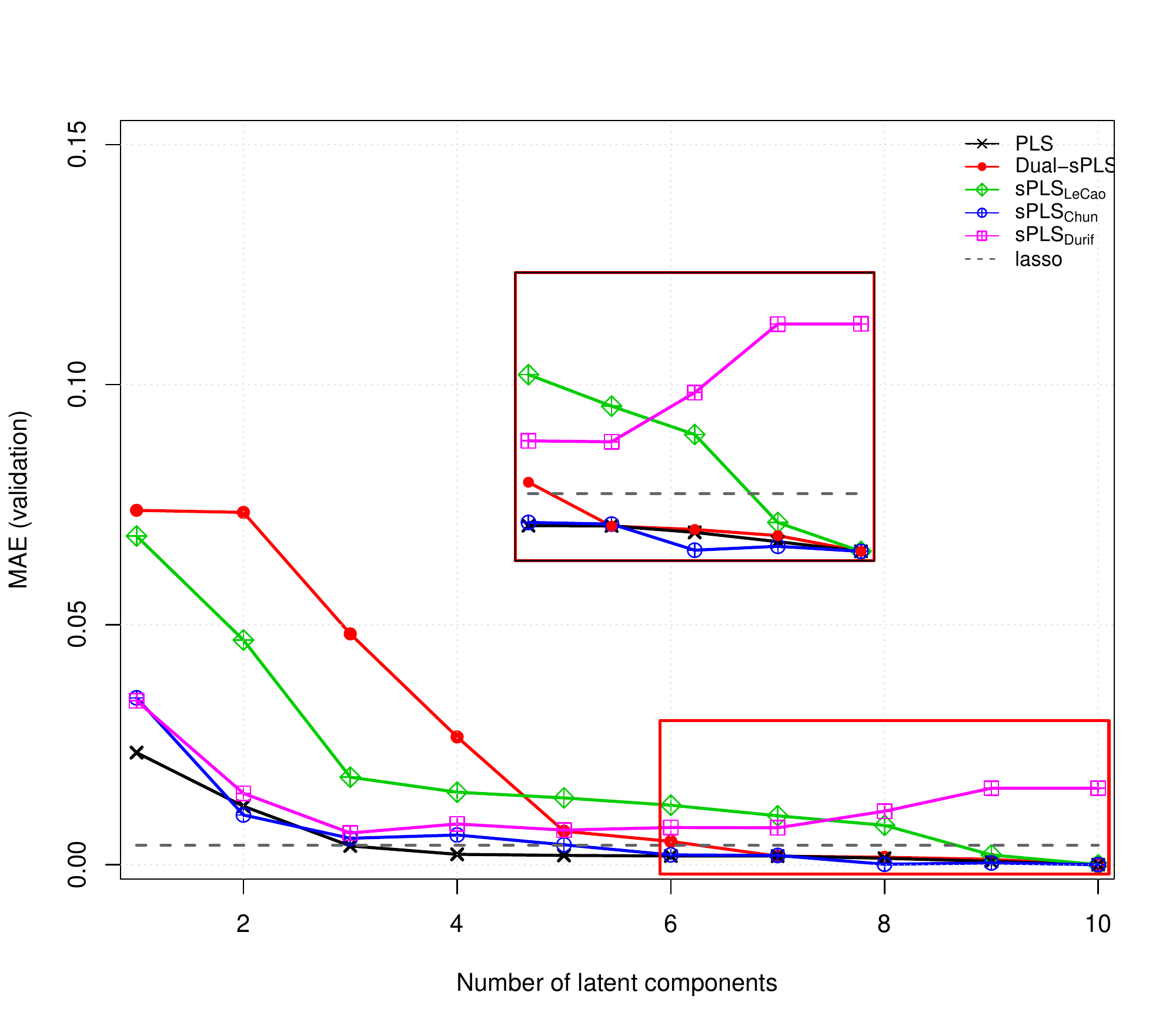}
    \end{subfigure}
    \\
     \begin{subfigure}{0.49\textwidth}
        \centering
        \includegraphics[width=\textwidth]{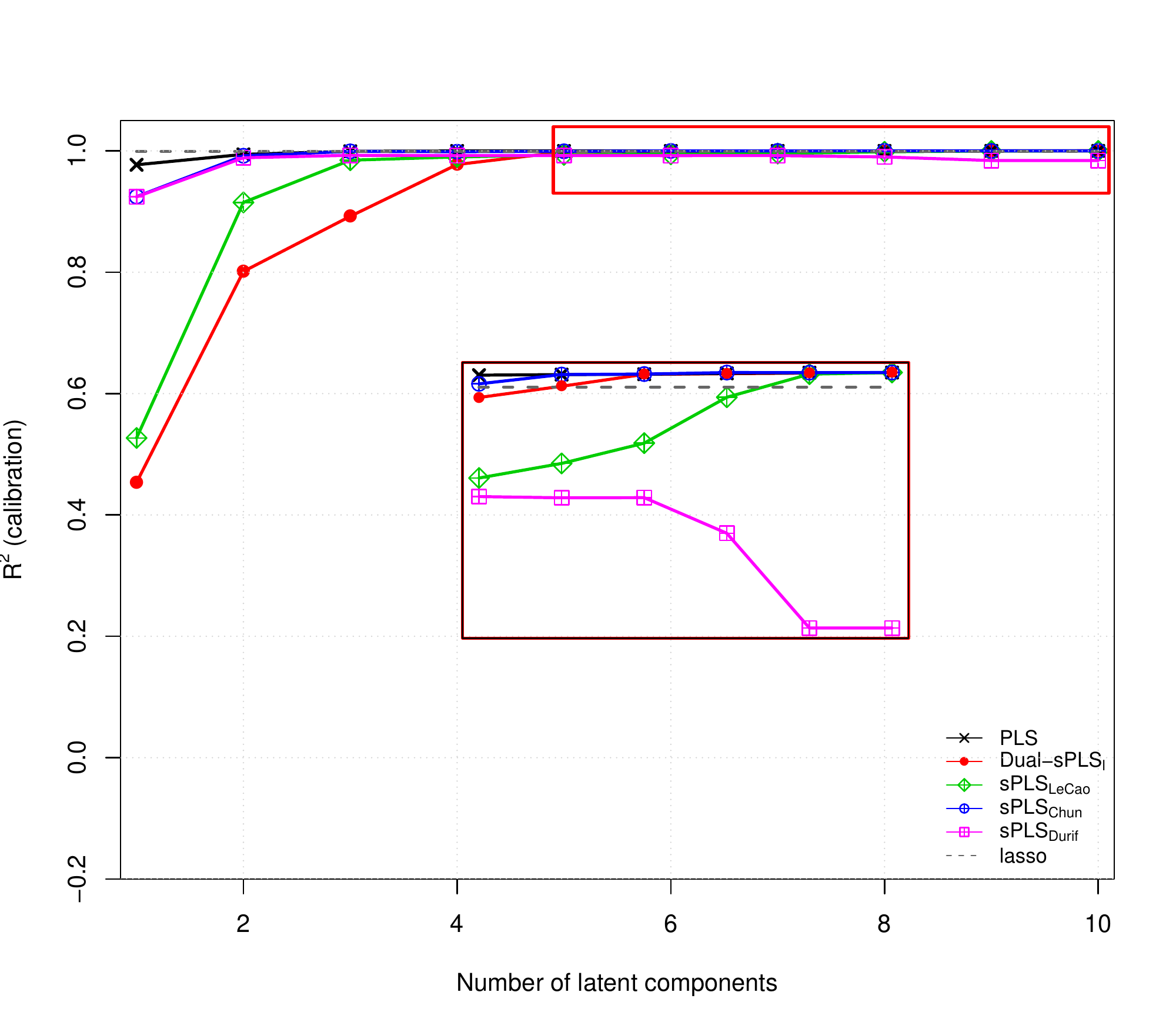}
    \end{subfigure}
    \hfill
        \begin{subfigure}{0.49\textwidth}
        \centering
       \includegraphics[width=\textwidth]{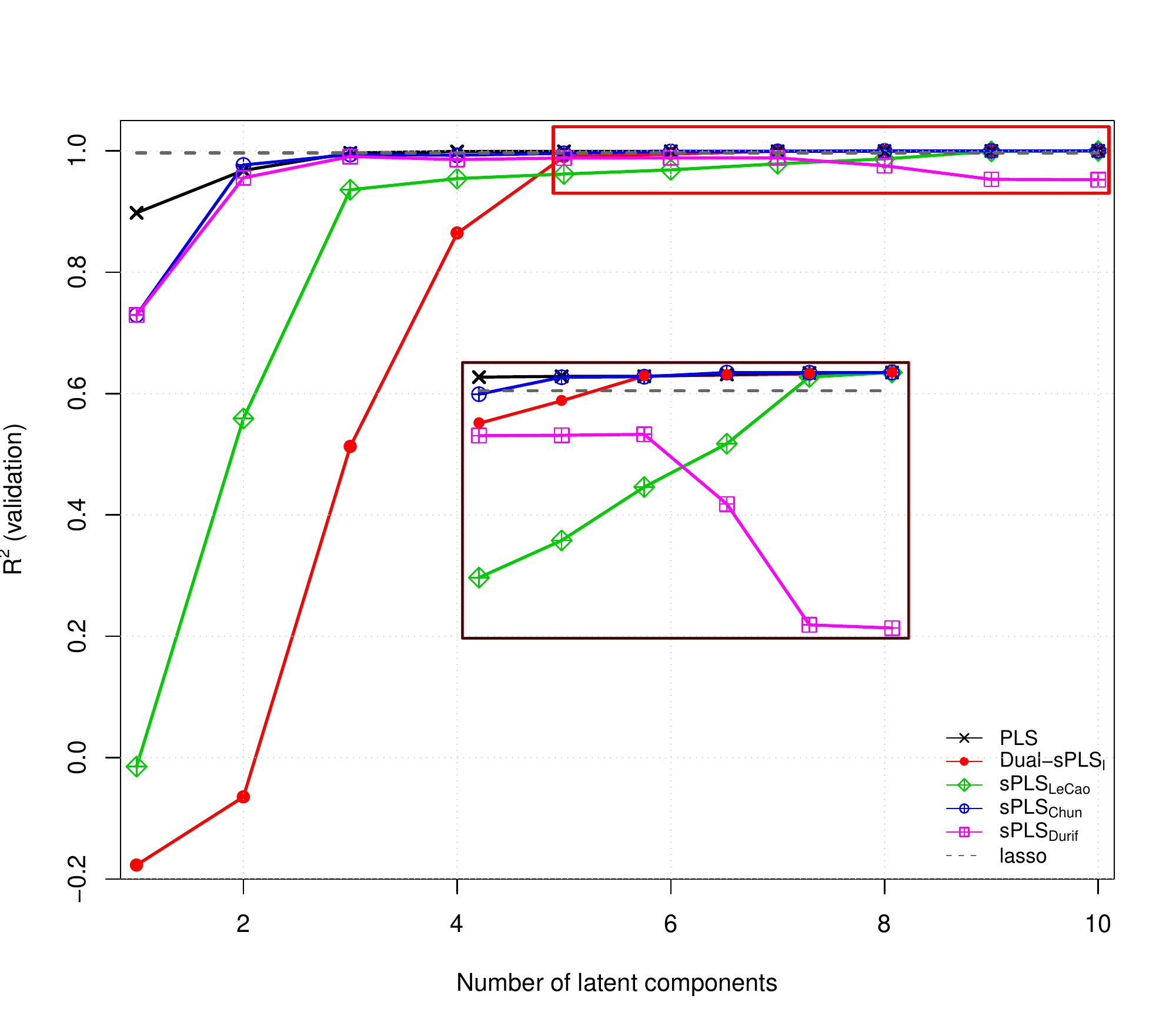}
    \end{subfigure}
    \end{tabular}
    \caption{\dspls{l} evaluation on simulated data \Ds. MAE (top) and R$^2$ (bottom) values for calibration (left) and validation (right) with respect to the number of latent components derived from PLS,  \dspls{l}, \spls{LeCao}, \spls{Chun}, \spls{Durif} and  lasso regressions.} 
    \label{fig_dspls_l_DSIM_MAE}
\end{figure}

\begin{figure}
    \centering
    \begin{tabular}{cc}
    \begin{subfigure}{0.49\textwidth}
        \centering
        \includegraphics[width=\textwidth]{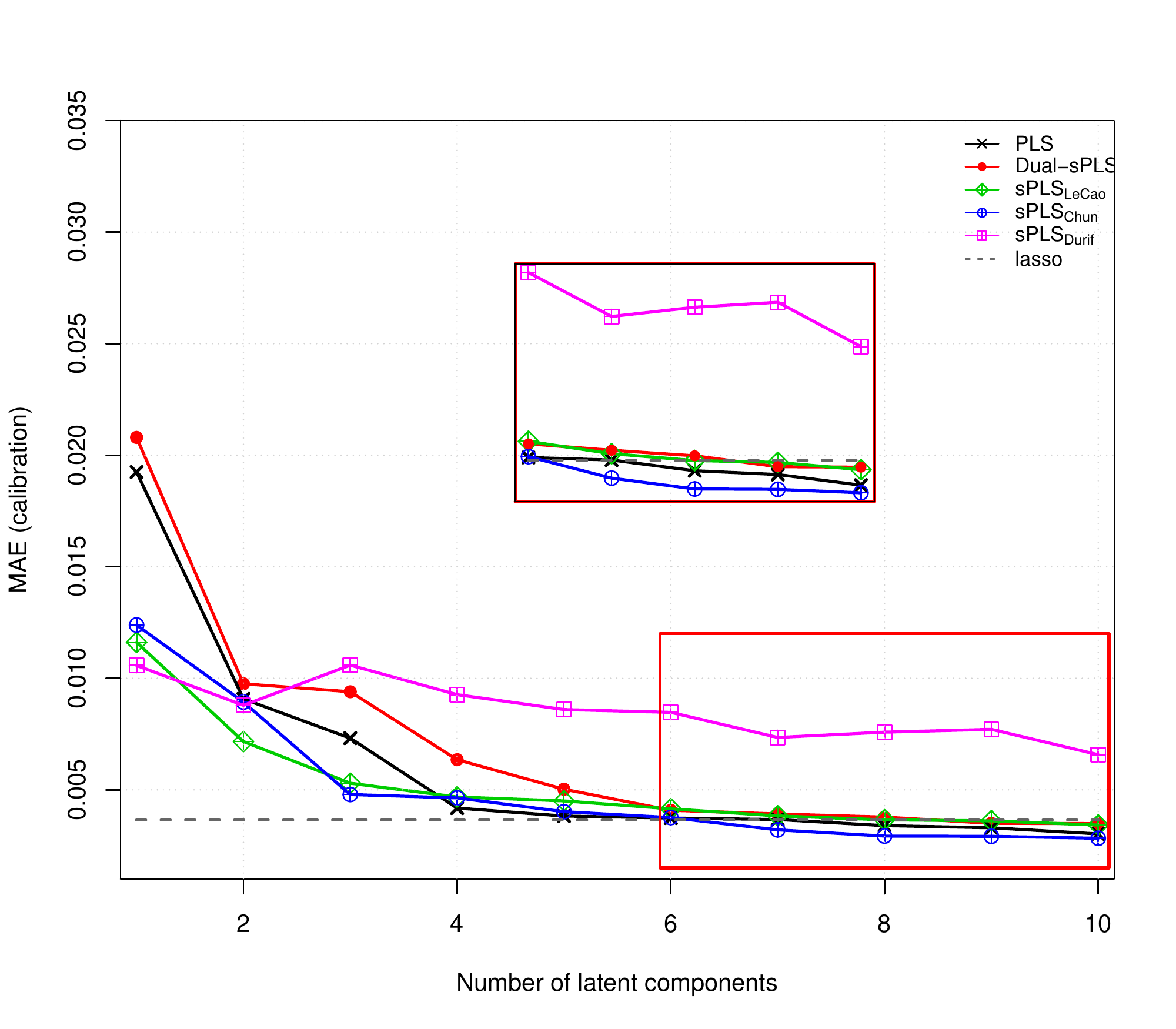}
    \end{subfigure}
    \hfill
        \begin{subfigure}{0.49\textwidth}
        \centering
       \includegraphics[width=\textwidth]{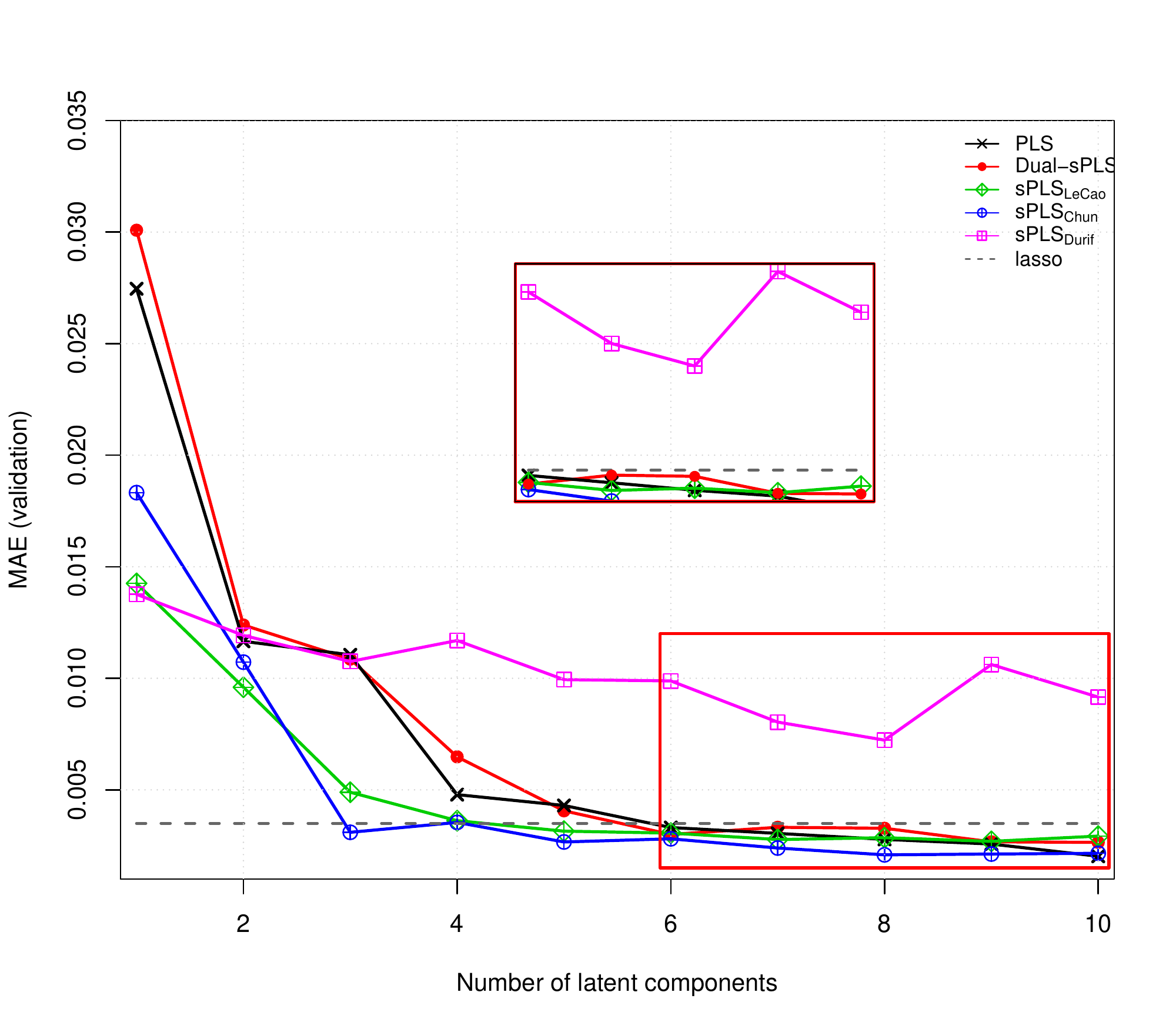}
    \end{subfigure}
    \\
     \begin{subfigure}{0.49\textwidth}
        \centering
        \includegraphics[width=\textwidth]{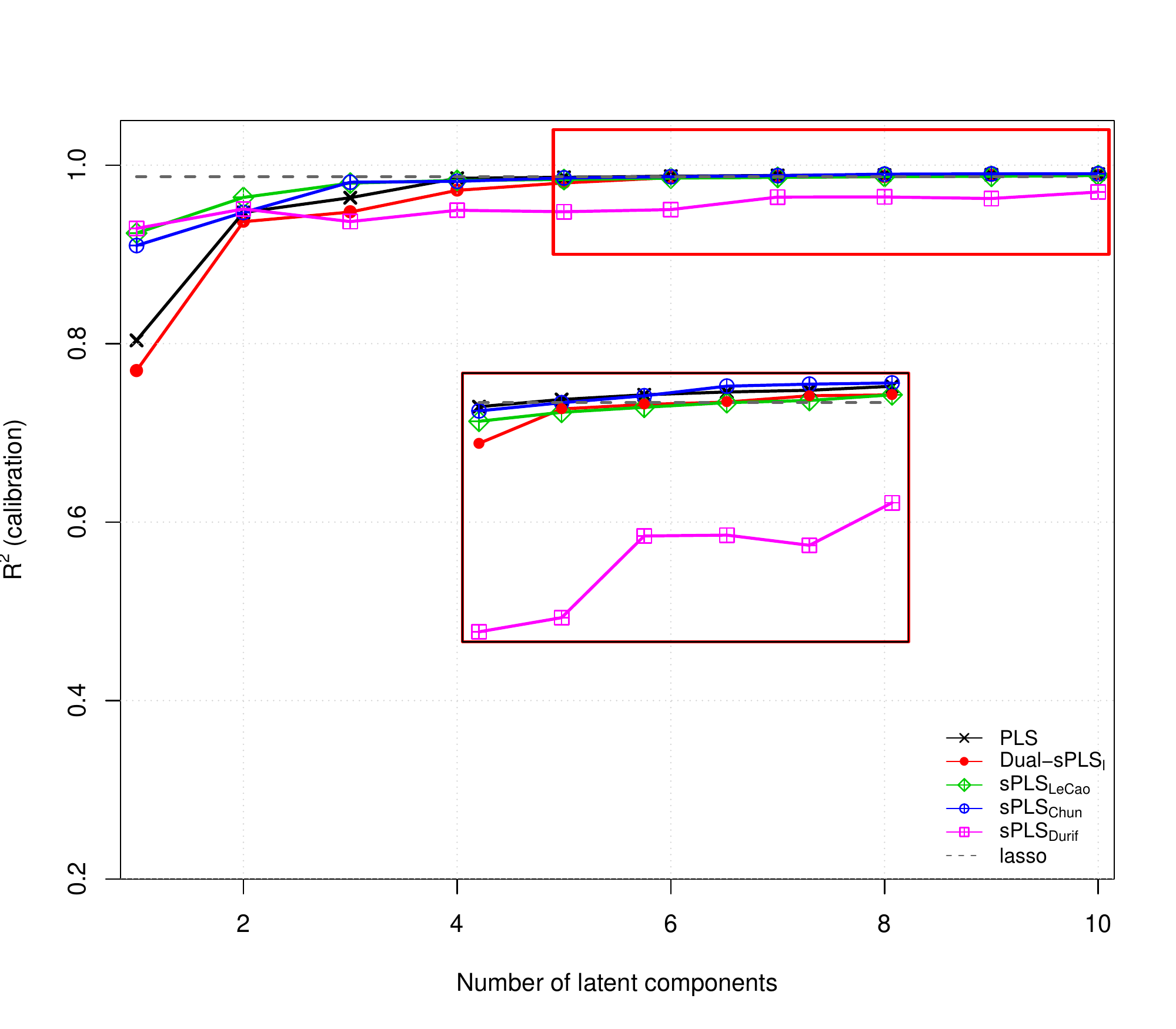}
    \end{subfigure}
    \hfill
        \begin{subfigure}{0.49\textwidth}
        \centering
       \includegraphics[width=\textwidth]{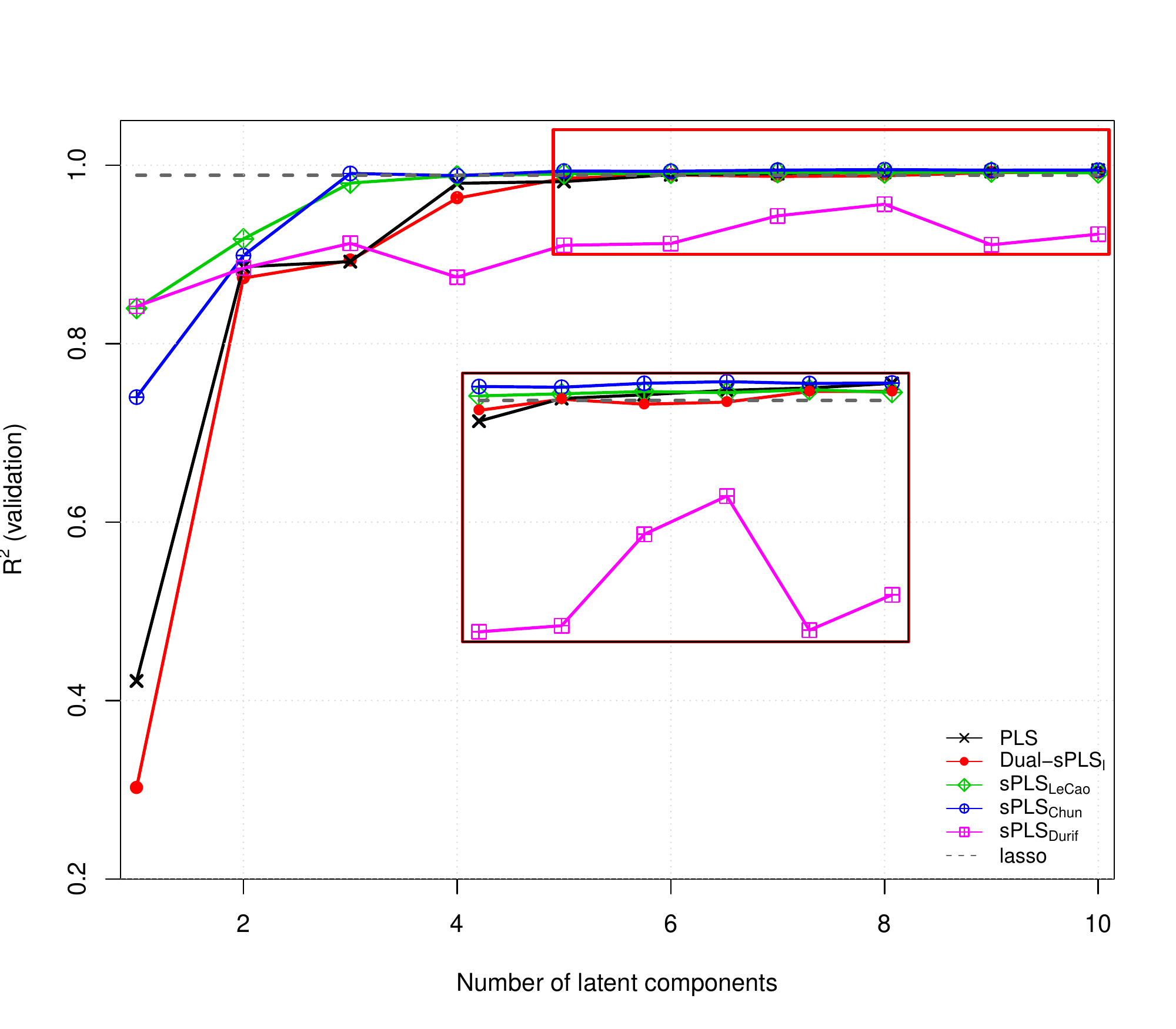}
    \end{subfigure}
    \end{tabular}
    \caption{\dspls{l} evaluation on real data \DNIR . MAE (top) and R$^2$ (bottom) values for calibration (left) and validation (right) with respect to the number of latent components derived from PLS,  \dspls{l}, \spls{LeCao}, \spls{Chun}, \spls{Durif} and  lasso regressions.} 
    \label{fig_dspls_l_DNIR_MAE}
\end{figure}

\begin{figure}
    \centering
    \begin{tabular}{cc}
    \begin{subfigure}{0.49\textwidth}
        \centering
        \includegraphics[width=\textwidth]{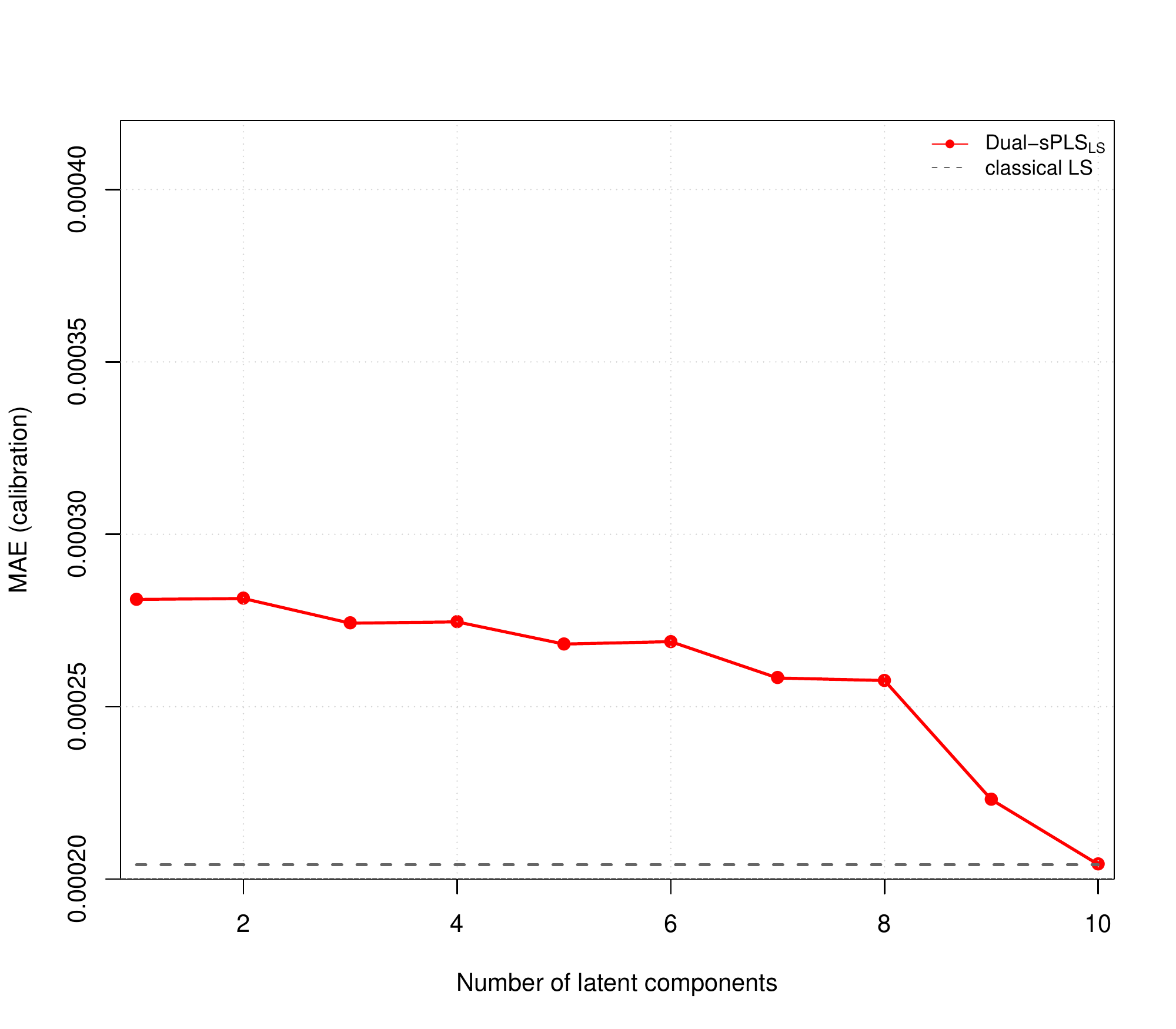}
    \end{subfigure}
    \hfill
        \begin{subfigure}{0.49\textwidth}
        \centering
       \includegraphics[width=\textwidth]{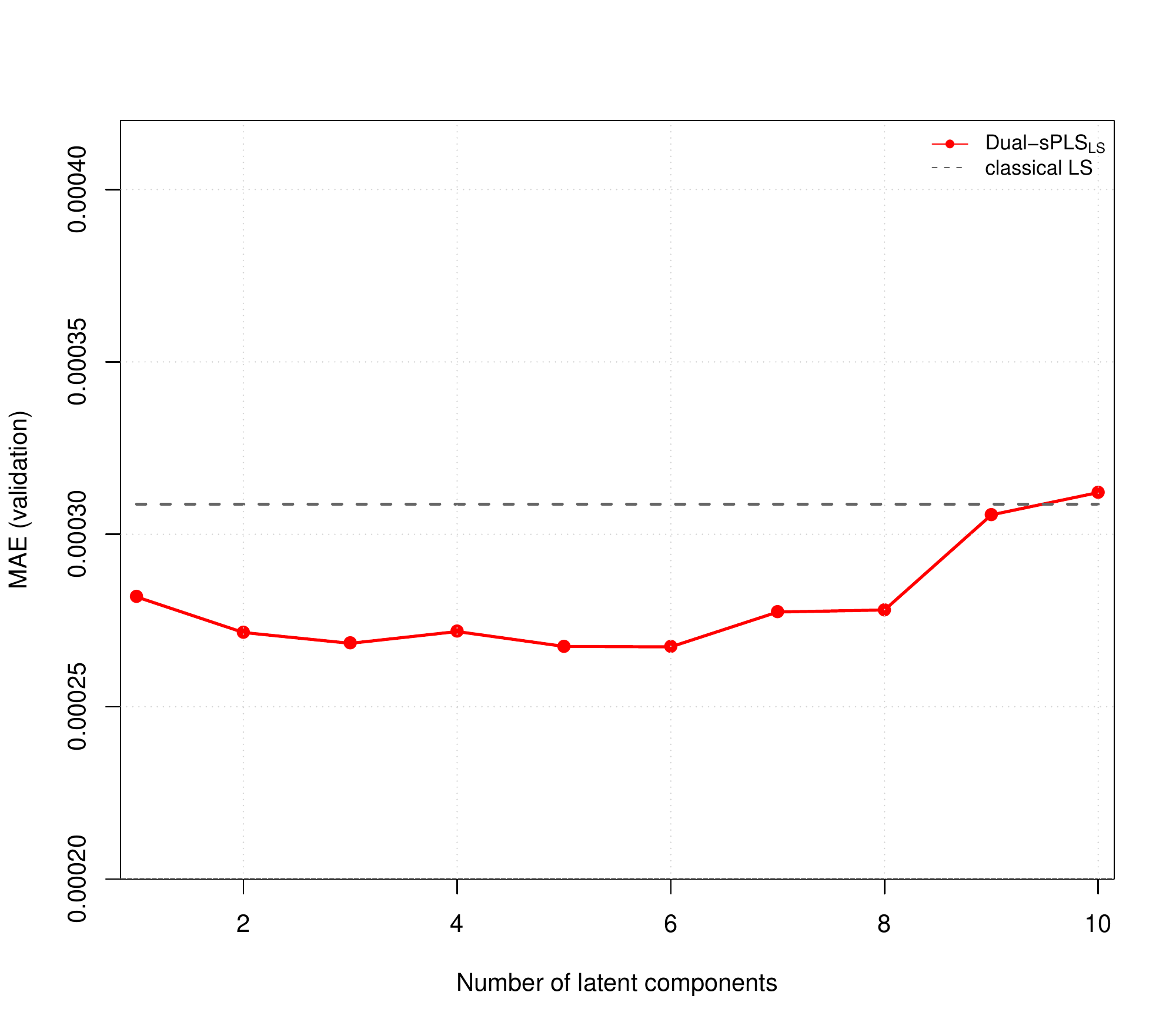}
    \end{subfigure}
    \\
     \begin{subfigure}{0.49\textwidth}
        \centering
        \includegraphics[width=\textwidth]{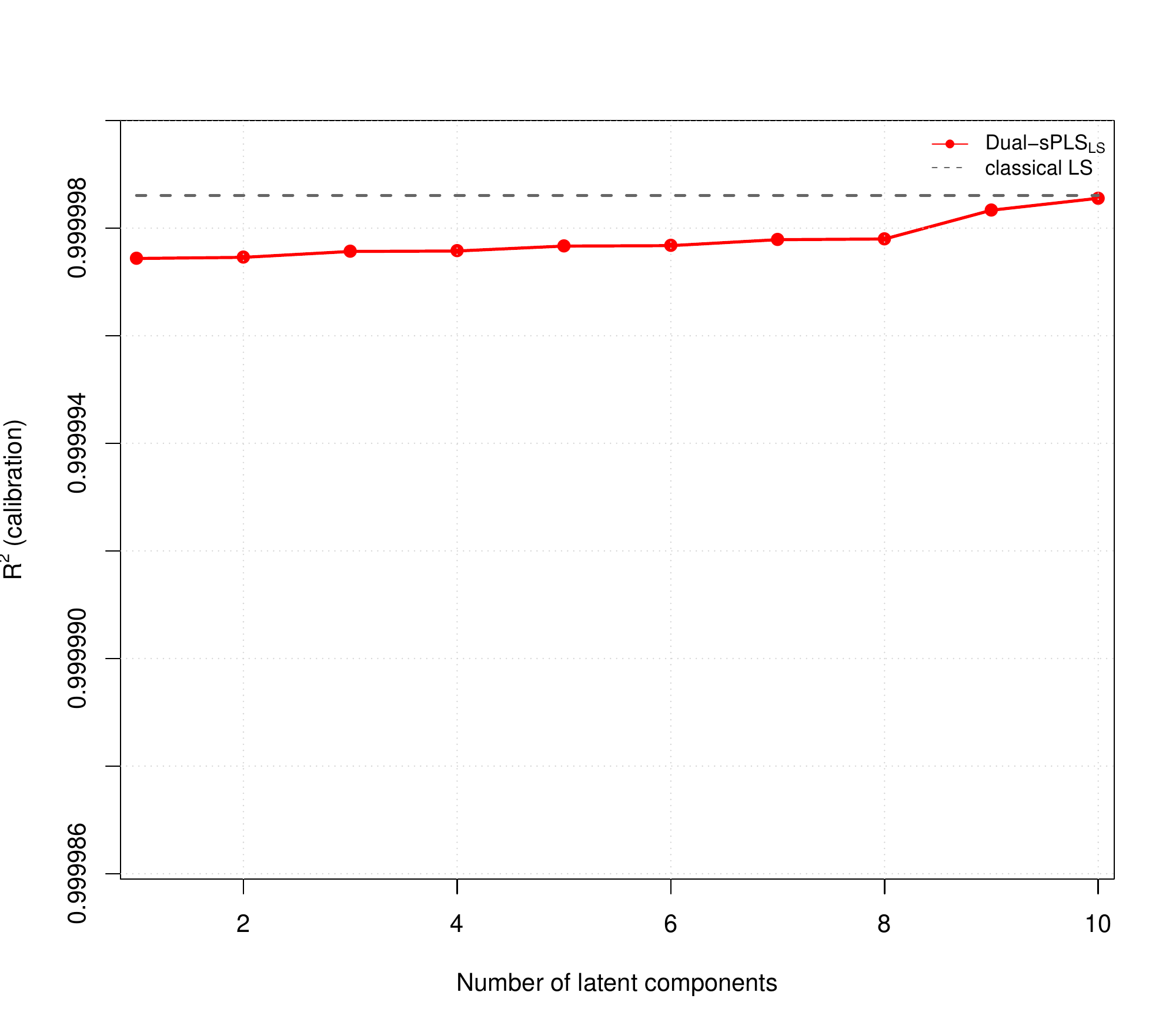}
    \end{subfigure}
    \hfill
        \begin{subfigure}{0.49\textwidth}
        \centering
       \includegraphics[width=\textwidth]{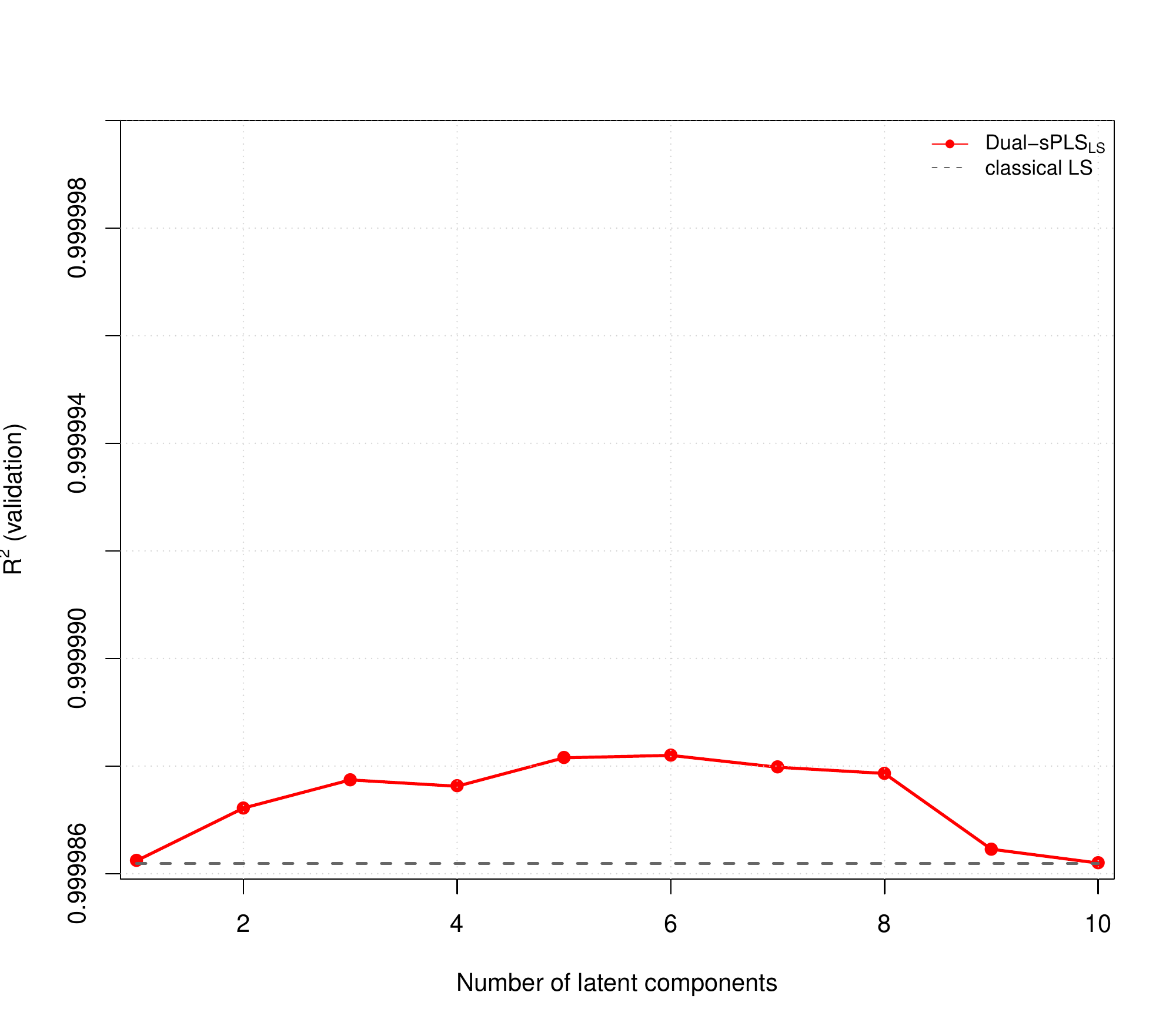}
    \end{subfigure}
    \end{tabular}
    \caption{\dspls{LS} evaluation on simulated data \Dsb. MAE (top) and R$^2$ (bottom) values for calibration (left) and validation (right) with respect to the number of latent components derived from \dspls{LS} and  least squares regressions.} 
    \label{fig_dspls_LS_DSIM_MAE}
\end{figure}

\begin{figure}
    \centering
    \begin{tabular}{cc}
    \begin{subfigure}{0.49\textwidth}
        \centering
        \includegraphics[width=\textwidth]{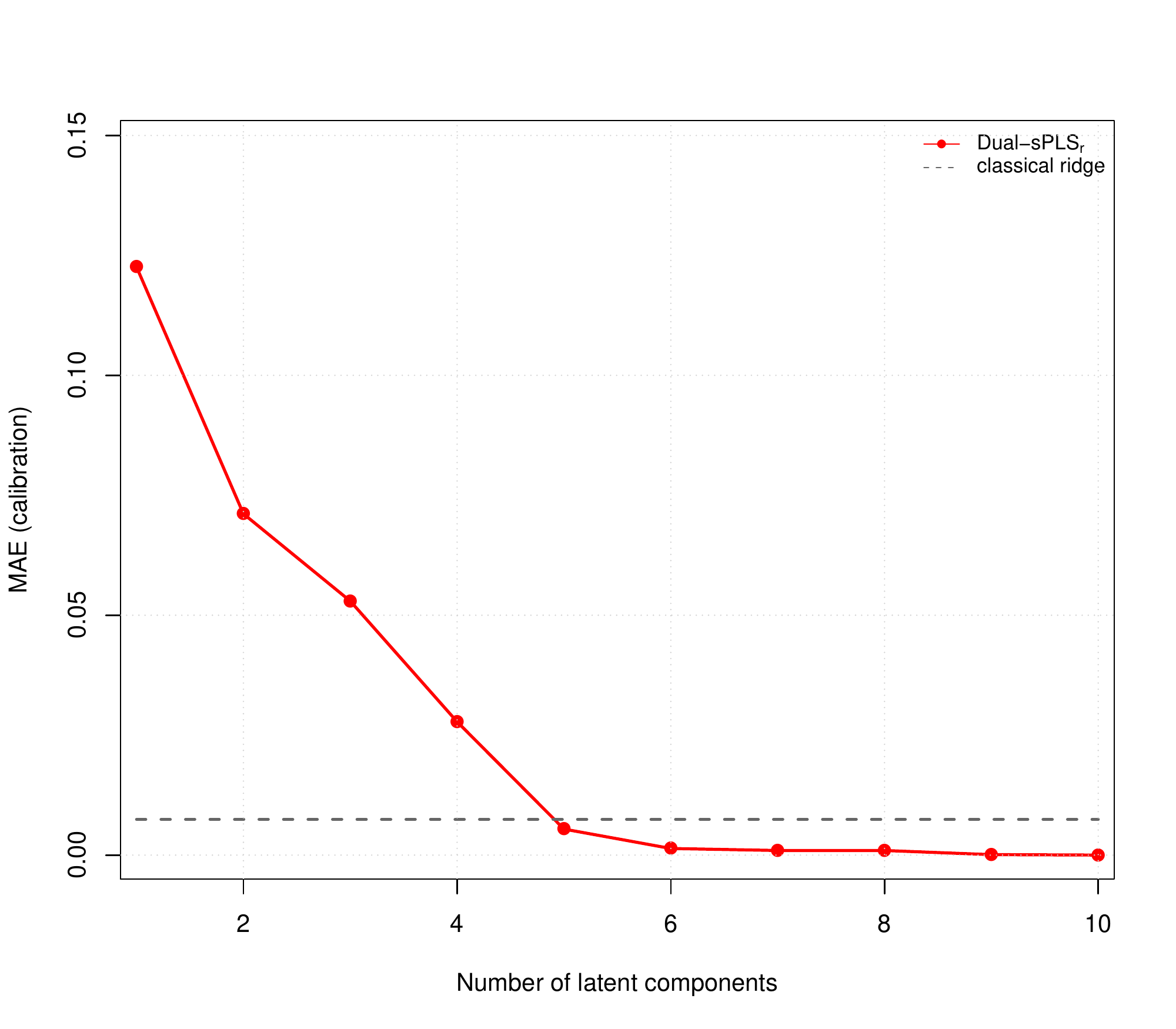}
    \end{subfigure}
    \hfill
        \begin{subfigure}{0.49\textwidth}
        \centering
       \includegraphics[width=\textwidth]{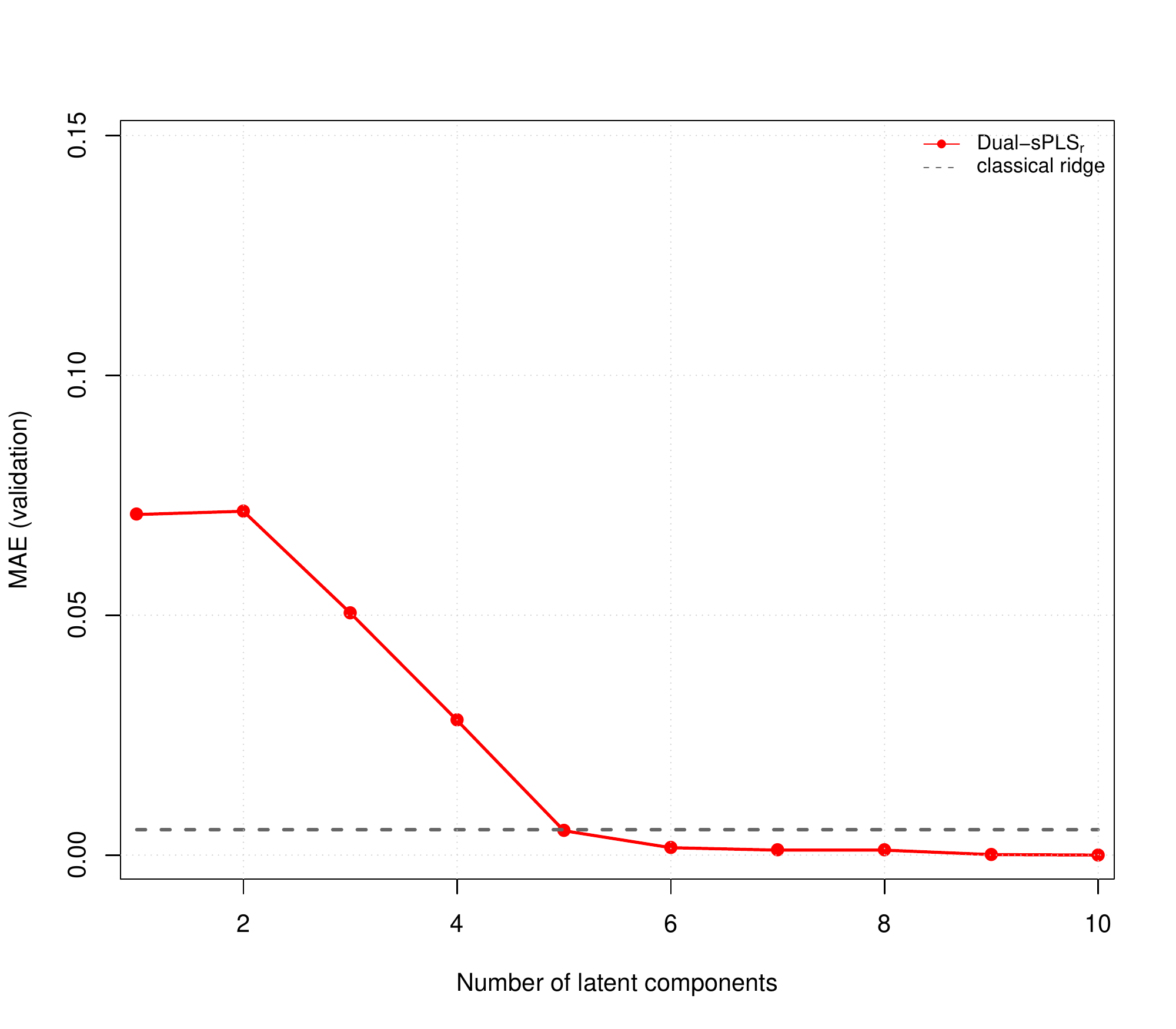}
    \end{subfigure}
    \\
     \begin{subfigure}{0.49\textwidth}
        \centering
        \includegraphics[width=\textwidth]{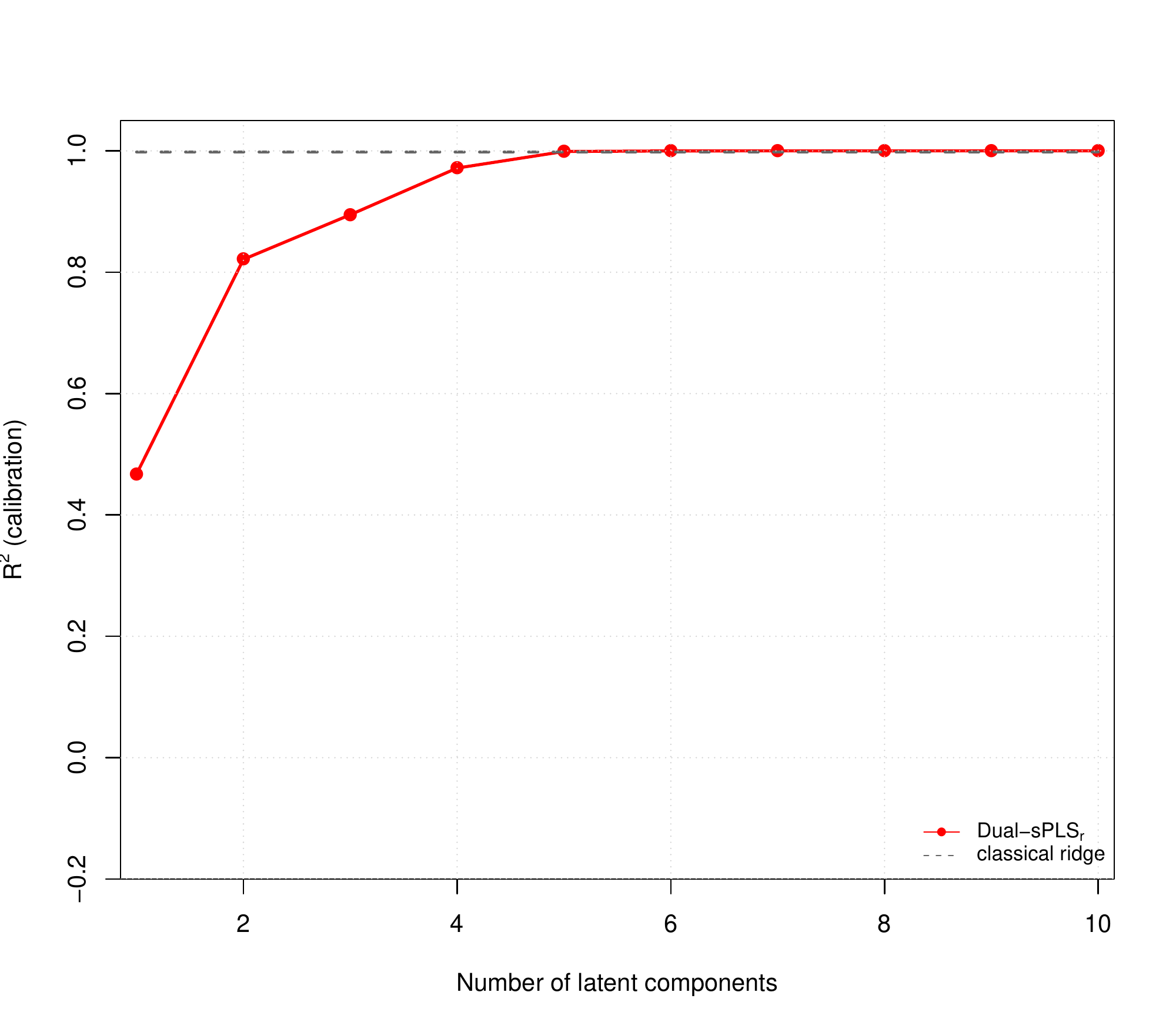}
    \end{subfigure}
    \hfill
        \begin{subfigure}{0.49\textwidth}
        \centering
       \includegraphics[width=\textwidth]{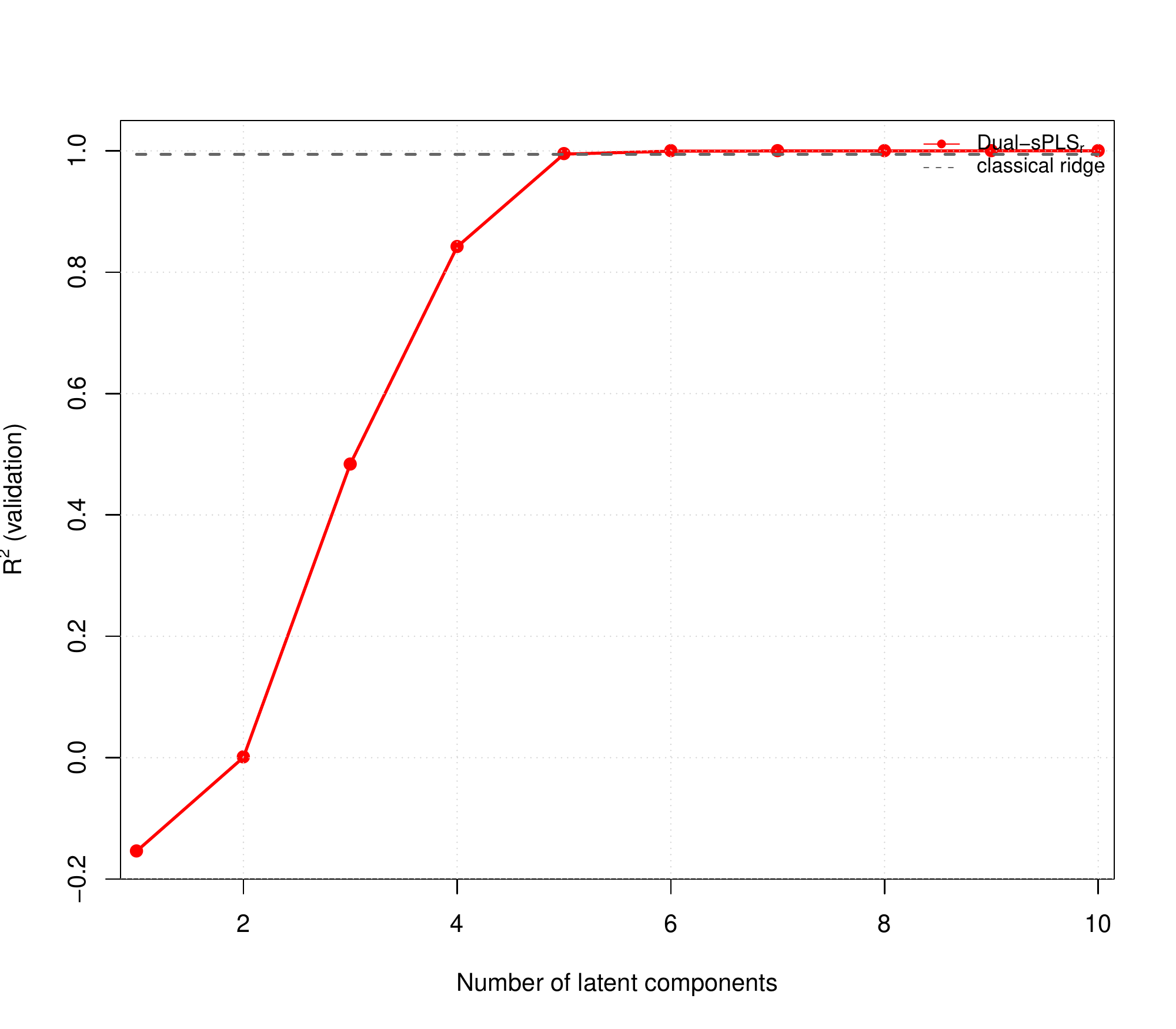}
    \end{subfigure}
    \end{tabular}
    \caption{\dspls{r} evaluation on simulated data \Ds. MAE (top) and R$^2$ (bottom) values for calibration (left) and validation (right) with respect to the number of latent components derived from \dspls{r} and ridge regressions.} 
    \label{fig_dspls_r_DSIM_MAE}
\end{figure}

\begin{figure}
    \centering
    \begin{tabular}{cc}
    \begin{subfigure}{0.49\textwidth}
        \centering
        \includegraphics[width=\textwidth]{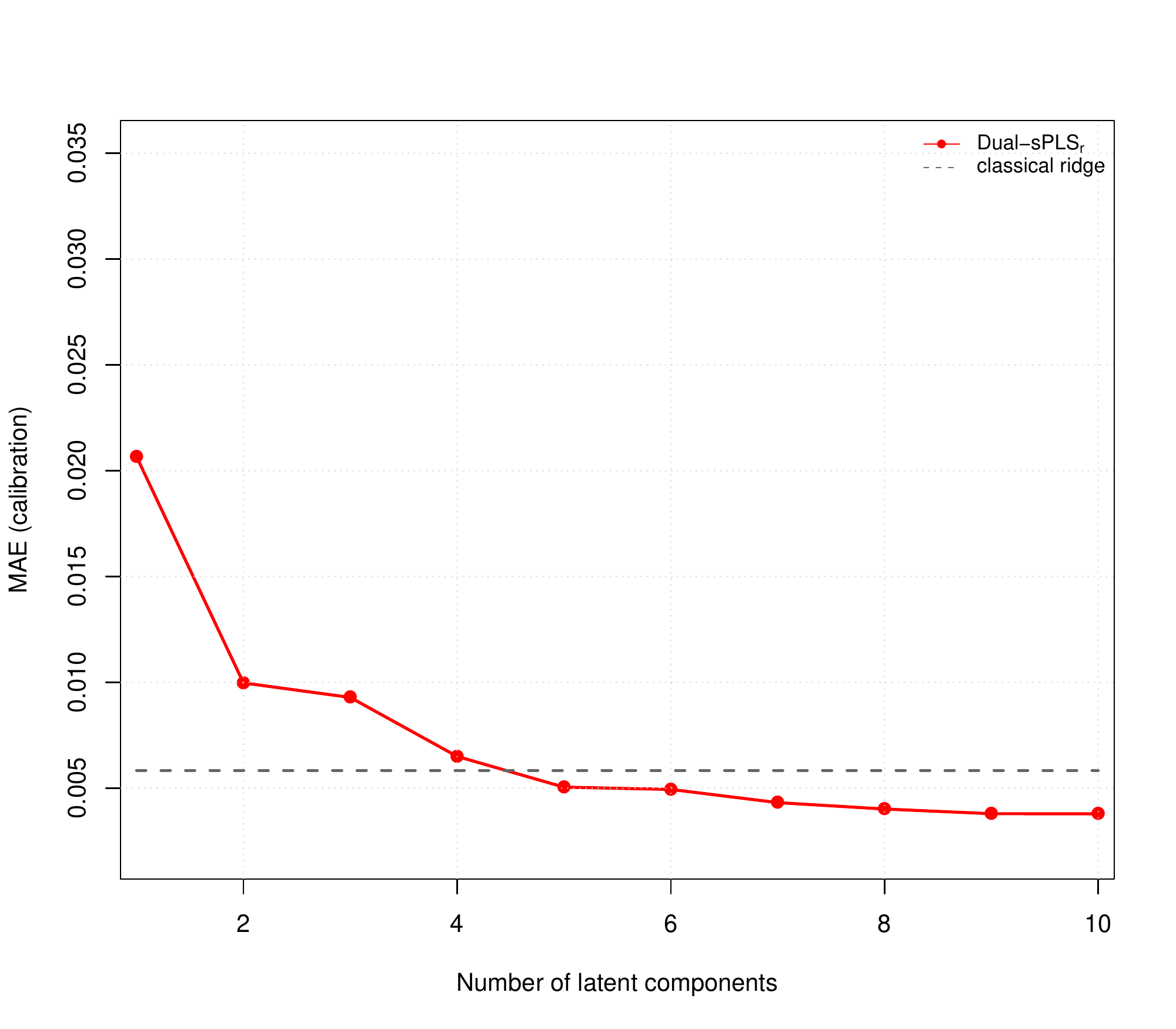}
    \end{subfigure}
    \hfill
        \begin{subfigure}{0.49\textwidth}
        \centering
       \includegraphics[width=\textwidth]{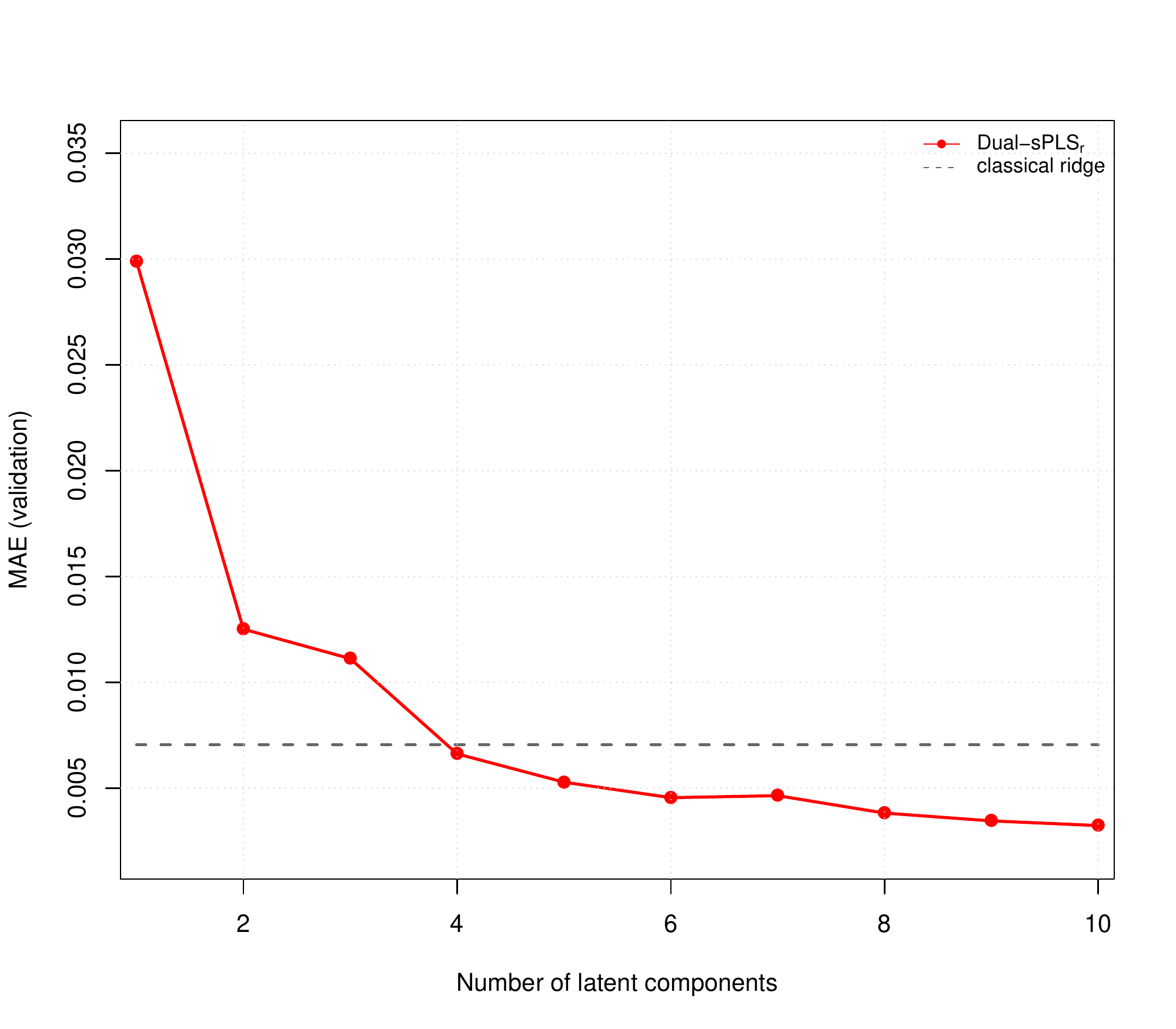}
    \end{subfigure}
    \\
     \begin{subfigure}{0.49\textwidth}
        \centering
        \includegraphics[width=\textwidth]{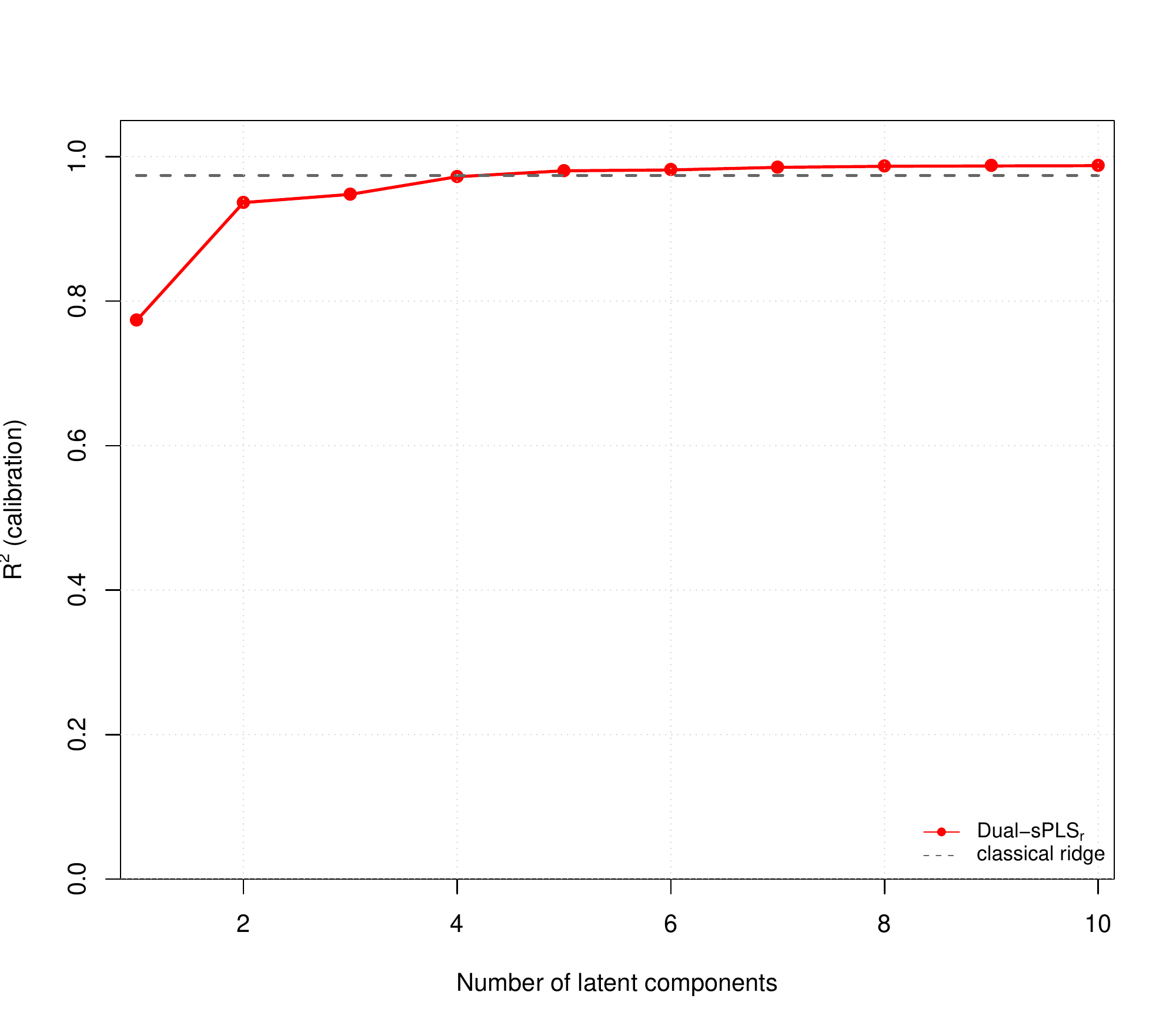}
    \end{subfigure}
    \hfill
        \begin{subfigure}{0.49\textwidth}
        \centering
       \includegraphics[width=\textwidth]{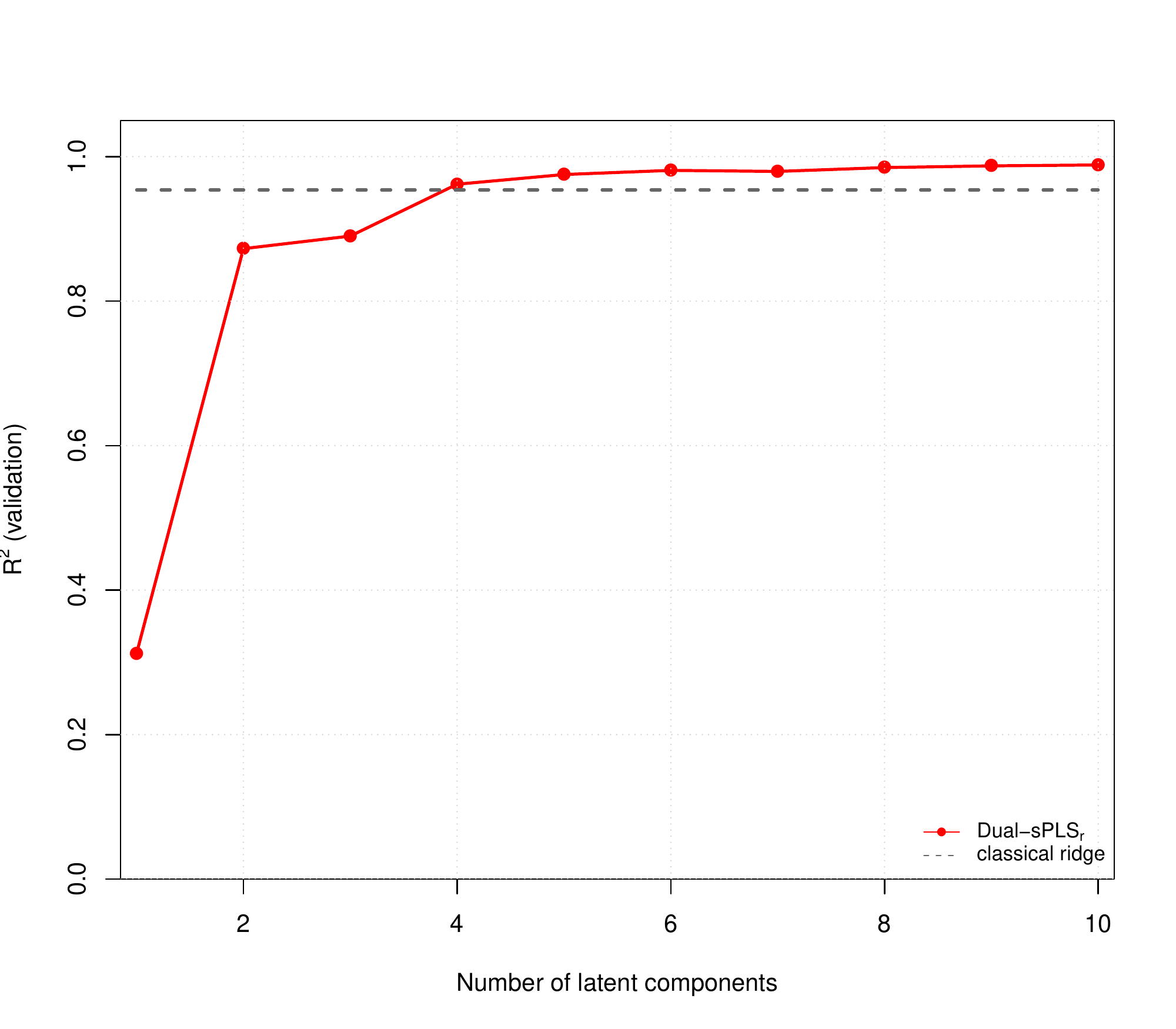}
    \end{subfigure}
    \end{tabular}
    \caption{\dspls{r} evaluation on simulated data \DNIR. MAE (top) and R$^2$ (bottom) values for calibration (left) and validation (right) with respect to the number of latent components derived from \dspls{r} and ridge regressions.} 
    \label{fig_dspls_r_DNIR_MAE}
\end{figure}

\end{document}